\documentclass[final]{ametsocV6.1_arxiv}

\usepackage{graphicx}
\usepackage{multicol,multirow}
\usepackage{amsmath,amssymb,amsfonts}
\usepackage{mathrsfs}

\usepackage{amsthm}
\usepackage{rotating}
\usepackage{appendix}
\usepackage{ifpdf}
\usepackage[T1]{fontenc}
\usepackage{newtxtext}
\usepackage{newtxmath}
\usepackage{textcomp}
\usepackage[table]{xcolor}
\usepackage{lipsum}
\usepackage[colorlinks,allcolors=blue]{hyperref}
\usepackage{float}
\usepackage{subcaption}
\usepackage{booktabs}
\usepackage{caption}
\usepackage{algorithm}
\usepackage{algpseudocode}
\usepackage{amsmath}
\usepackage{pdflscape}
\usepackage{longtable}

\newcommand{\R}{\ensuremath{\mathbb{R}}}    

\newcommand{\y}{\boldsymbol{y}}

\newcommand{\X}{\boldsymbol{X}}
\newcommand{\x}{\boldsymbol{x}}
\newcommand{\U}{\boldsymbol{U}}

\newcommand{\z}{\boldsymbol{z}}
\newcommand{\I}{\boldsymbol{I}}

\newcommand{\A}{\boldsymbol{A}}

\newcommand{\bL}{\boldsymbol{L}}

\newcommand{\methodname}{PullbackDMDc}
\newcommand{\add}{\color{black}}
\newcommand{\added}{\color{black}}

\theoremstyle{definition}

\numberwithin{equation}{section}

\let\internallinenumbers\relax

\makeatletter
\renewcommand{\p@subsection}{\thesection}
\renewcommand{\p@subsubsection}{\thesection\thesubsection}
\makeatother

\usepackage{etoolbox}
\makeatletter
\patchcmd{\maketitle}{%
\lhead{\color{gray}Generated using the official AMS \LaTeX\ template v6.1 two-column layout. This work has been submitted for publication. Copyright in this work may be transferred without further notice, and this version may no longer be accessible.}
\ifdraft\lhead{\hfill\color{gray} Generated using the official AMS \LaTeX\ template v6.1 \hfill} \fi
}{%
\lhead{\footnotesize\itshape This manuscript has been submitted for publication in the \textit{Journal of Climate}. It has not yet been peer reviewed and is not the version of record.}
}{\typeout{^^JAMS header patch: SUCCEEDED^^J}}{\typeout{^^JAMS header patch: FAILED^^J}}
\makeatother

\title{Disentangling Forced and Internal Climate Variability in Single Realizations using Dynamic Mode Decomposition with Control}

\authors{Nathan Mankovich$^+$\correspondingauthor{nathan.mankovich@uv.es}, Andrei Gavrilov$^+$\correspondingauthor{andrei.gavrilov@uv.es} and Gustau Camps-Valls}

\nolinenumbers

\affiliation{\nolinenumbers {Image Processing Laboratory (IPL), Universitat de Val\`encia, Spain}\\
\aff{+}{These authors contributed equally to the manuscript}
}

\abstract{ 
\added{Separating externally forced trends from internal climate variability is a fundamental challenge in climate science, particularly when analyzing observational records that exist only as a single realization.} 
We show that a single climate realization can be decomposed into forced and internal components by treating external forcing as a dynamical driver within a linear stochastic system, an idea grounded in pullback attractor theory. 
\added{This addresses a central challenge for climate projection and for detecting and attributing the forced response.}
\added{Existing single-realization methods rely on linear inverse models (LIMs), which ignore forcing, or linear regression, which omits dynamics.}
Here we introduce PullbackDMDc, a method grounded in non-autonomous dynamical systems theory and dynamic mode decomposition with control (DMDc), incorporating pullback attractor estimation to decompose a single climate realization into spatial modes and their associated forced and internal components, yielding a physically interpretable picture of the underlying dynamics. \added{After testing on a synthetic system with a known forced response, we} illustrate the utility of PullbackDMDc for Earth System Model (ESM) evaluation by applying it to near-surface air temperature and sea-level pressure from reanalysis and four ESM large ensembles. PullbackDMDc estimates the forced response with skill matching or exceeding established baselines and \added{helps select} forcing predictors against model-based ground truth. Its internal-variability components reveal that ESMs qualitatively capture interannual and decadal modes while exhibiting systematic differences relative to each other and to observations. Skillful forced-response estimation and a novel decomposition position PullbackDMDc as a practical tool for single-realization climate analysis and ESM evaluation.}

\begin{document}

\setcounter{secnumdepth}{3}
\maketitle

\section{Introduction}

Observed climate change reflects the superposition of two contributions: the response of the climate system to external forcing (e.g., greenhouse gases, aerosols, and volcanic and solar activity) and internal variability generated by its own coupled dynamics~\citep{Hasselmann1976,Deser2012}. The forced response drives the long-term trends behind the rising frequency and intensity of many climate extremes, whereas internal variability modulates these trends across time scales from seasons to decades and dominates the uncertainty of near-term, regional projections~\citep{Deser2012,Lehner2020}. Accurately disentangling the two is fundamental both to attributing observed changes to human influence and to producing credible, policy-relevant projections~\citep{hasselmann1993optimal,Sippel2021}.

However, we observe only a single realization of Earth's climate, in which forced and internal contributions are superposed, \added{and we cannot rerun the past climate from different initial conditions or under different forcings.} This limitation motivates the development of Earth system models (ESMs), which enable the simulation of multiple realizations of the climate system under varied forcing and initial conditions, now routinely run as large ensembles~\citep{Deser2020, Maher2019, maher2025updated}. 
\added{Under time-dependent forcing, the climate as a dynamical system has no fixed attractor, as it would under constant forcing, but a pullback (or snapshot) attractor, a time-dependent set of states defined in Section~\ref{sec:methods}. It is approximated by a large ensemble run from different initial conditions under identical forcing, once initial-condition dependence has decayed~\citep{langa2002stability, cheban2004global, Chekroun2011, Ghil_2020, tel2020theory, pierini2020statistical}. At each time step, the ensemble samples the instantaneous climate statistics. Its mean is} the standard ensemble-mean definition of forced response in large ensembles~\citep{frankcombe2018choice}, \added{and deviations around the mean represent the internal variability}.

Yet to disentangle forced response and internal variability in the observational record, we must develop methods to estimate these components from a single realization of the climate system. A wide range of approaches has been proposed for this task, spanning statistical detection and attribution methods~\citep{Sippel2021} to dynamical-systems-based techniques~\citep{Frankignoul17, xu2022increase}. Many of these methods are supervised by ESM ensembles, meaning they directly use the ensemble-mean ``ground truth'' forced response in ESMs and sometimes ESM simulations. Supervised methods include fingerprinting~\citep{Sippel2021, santer2023exceptional, pmlr-v258-durand25a, Kravtsov2025,Wills2026}, which can be extended to nonlinear settings using neural networks~\citep{barnes2019viewing, barnes2020indicator,Bone2024}, as well as dynamical adjustment~\citep{Wallace2012, Smoliak2015, Deser2016}, which removes the influence of atmospheric circulation variability by using ESM-derived analogs. Although these methods can be skillful, they often require substantial computational resources, can reduce physical interpretability, and depend on the fidelity of the underlying ESM ensemble in representing the real climate and its forced response.

Linear, unsupervised statistical machine learning (ML) methods have been highly studied and are simple, admit fast computation times for large-scale experimentation, and retain high physical interpretability. They include dynamical system models~\citep{Penland1995, Frankignoul17, xu2022increase} such as linear inverse models (LIM)~\citep{Penland1993}, regression-based approaches such as global mean temperature regression~\citep{dai2015decadal}, and decomposition methods such as low\added{-}frequency component analysis (LFCA)~\citep{Wills2018} or linear dynamical modes (LDMs)~\citep{Gavrilov2020}, \added{both of which favor temporally smooth components. Specifically, LFCA maximizes the ratio of low-frequency to total variance and LDMs place a smoothness prior with an optimized timescale on their latent time series}.

The recent Forced Component Estimation Statistical Method Intercomparison Project (ForceSMIP) delivered the first large-scale, common-protocol comparison of this zoo of statistical forced-response estimators~\citep{Wills2026}. Among the unsupervised single-realization methods in ForceSMIP, some separate forced response from internal variability using dynamical system models~\citep{Frankignoul17}, ignoring forcing information; others use regression with forcing information~\citep{dai2015decadal}, thereby missing climate system dynamics. Non-autonomous (forced) dynamical system models naturally bridge this gap, since they let the evolution operator depend on external forcing~\citep{Kloeden2013,Bezruchko2010,Proctor2018}. Despite active development and use in climate studies~\citep{Mukhin2015a,Mukhin2015b,Mukhin2019,Gavrilov2019,Gavrilov2022,Mankovich2025}, they have not been widely applied to forced-response and internal-variability identification. The simplest linear version is used in dynamic mode decomposition with control (DMDc)~\citep{proctor2016dynamic}, which augments a linear dynamical system with an external control term \added{(e.g., deterministic forcing)}. However, DMDc alone does not address forced-response and internal-variability disentanglement.

Here we introduce PullbackDMDc, a simple, interpretable, unsupervised linear method that turns this idea into a forced/internal decomposition. We view a single climate record as lying on the pullback attractor of a linear stochastic system driven by external forcing. Because the system is linear, its pullback attractor collapses to a single trajectory that splits exactly into a forced response, controlled by the forcing, and an internal component, driven by noise, both read directly off the fitted linear evolution matrices. This links pullback-attractor theory with data-driven modeling and, by projecting onto the DMDc eigenmodes, expresses the full spatio-temporal record as a sum of spatial patterns, each carrying its own forced and internal time series.

We evaluate the method across four large ensembles and reanalysis data using standard forced-response metrics. We show that~\methodname~achieves competitive or superior performance compared to \added{standard baselines} while retaining a physically interpretable decomposition into spatial modes and their associated forced and internal components. We further show that the implied decomposition enables analysis and evaluation of ESMs against observations in terms of the dominant modes. We present selected examples of this evaluation, highlighting differences and similarities in the representation of dominant interannual-to-multidecadal modes and \added{changes in the seasonal cycle} across ESMs and observations.

\added{Section~\ref{sec:methods} of the paper describes the baselines and \methodname, and Section~\ref{sec:results} presents the results for a synthetic system, ESMs and reanalysis. Details of methods, a synthetic example, and additional figures are in the Supplementary Material.}

\section{Methods}\label{sec:methods}

\added{Suppose we have a time series of a climate variable, $\{\x(1),\ldots,\x(T)\}\subset\R^{M}$, with $M$ the number of spatial locations. We aim to estimate from this single record a forced component $\x^{(f)}(t)$, the part of climate variability that responds to external forcings such as greenhouse gases and aerosols. For a given ESM, it is usually estimated as the mean over a large ensemble run under the same forcing~\citep{Wills2026}, or approximated from a smaller ensemble~\citep{Wills2020,Kravtsov2022,Gavrilov2024,Seleznev2026}, which assumes that this mean converges as the ensemble size and the evolution time before $t$ grow. The internal component is then the residual, $\x^{(i)}(t)=\x(t)-\x^{(f)}(t)$. This definition does not guarantee that the two components are independent. In fact, a rigorous definition of the forced response in an arbitrary nonlinear dynamical system is not readily available, and the above-defined $\x^{(f)}(t)$ can miss forced changes in the distribution, such as in extremes, which then appear in $\x^{(i)}(t)$. Nevertheless, the ensemble mean captures the forced change of the mean state, and we use it operationally as the ground truth, following~\citet{Wills2026}.} 

\subsection{Baselines for forced-response estimation}\label{sec:baselines}
Our baselines \added{below} either ignore external forcing \added{or use it without modeling the dynamics. \emph{RegGMST}~\citep{dai2015decadal} takes the global mean surface air temperature (GMST) as a proxy for the forced response. It fits $\x(t)\approx\boldsymbol{\beta}\mu_{\mathrm{SAT}}(t)$, with $\mu_{\mathrm{SAT}}(t)$ the mean-centered GMST, and estimates $\x^{(f)}(t)=\boldsymbol{\beta}\mu_{\mathrm{SAT}}(t)$.}

The Linear Inverse Model (LIM)~\citep{Penland1993,MCP96} \added{assumes that $\x(t)$ follows a linear stochastic dynamical system~\citep{arnold2006random}, whose discrete form over a unit time step is}
\begin{equation}\label{eq: lim_discrete}
    \x(t) = \A \x(t-1) + \boldsymbol{\xi}(t),
\end{equation}
\added{where $\A$ is the propagator over a unit time step and $\boldsymbol{\xi}(t)$ is temporally white noise with zero mean. The deterministic part of this model is used in Dynamic Mode Decomposition (DMD)~\citep{Schmid2010,tu2014dmd} (see the Supplementary Material for details on LIM and DMD).}

LIM estimates the forced response as the projection of the data onto the least-damped mode, the eigenvector of $\A$ with the longest decay time~\citep{compo2010removing,Frankignoul17}, \added{which often gives noisy estimates.} \added{LIM with optimal perturbation patterns (LIMopt)~\citep{solomon2012reconciling,Frankignoul17,Wills2020} uses the same $\A$ but smooths the estimates with the leading singular vectors of $\A^{\tau_e}$, the propagator over an optimal perturbation time $\tau_e$. We implement LIM and LIMopt following the ForceSMIP example notebook (\url{https://github.com/karenamckinnon/forcesmip}) and Tier~1 repository (\url{https://github.com/ForceSMIP/tier1-methods}).}

\added{Finally,} the Linear Regression (LR) baseline \added{uses the forcing $\y(t)$ directly,} fitting \added{$\x(t)\approx\boldsymbol{B}\y(t)$ and estimating $\x^{(f)}(t)=\boldsymbol{B}\y(t)$. It follows the MLR-Forcing method of ForceSMIP~\citep{Wills2026} and, like RegGMST, regresses the data on a proxy of the forced response, here the forcing rather than GMST.}

\subsection{Pullback DMDc}
\added{\methodname~combines the dynamical nature of LIM with the forcing input of LR. We fit a \textit{linear stochastic model with forcing}, separate its forced and internal components through the pullback attractor, and project both onto the DMDc modes~\citep{proctor2016dynamic}.}

\subsubsection{Linear stochastic model with forcing}

In a forced\footnote{The noise term of a stochastic model is often called ``forcing'' as well. However, here we separate it from external forcing and interpret it as an approximation of internal dynamical system components.} (driven) dynamical system, the evolution operator \added{depends} on the external forcing~\citep{Bezruchko2010,Kloeden2013,Proctor2018}. The simplest extension of discrete LIM~\eqref{eq: lim_discrete} \added{adds} a linear forcing term:
\begin{equation}\label{eq: linear_system}
    \x(t) = \A\x(t-1) + \boldsymbol{B} \y(t) + \boldsymbol{\xi}(t).
\end{equation}  
\added{We fit $\A$ and $\boldsymbol{B}$ by unregularized least squares} (see the Supplementary). 
\added{For $\boldsymbol{B}=\boldsymbol{0}$, the model reduces to LIM, while for $\A=\boldsymbol{0}$ it reduces to LR, so comparing LR with \methodname~isolates the contribution of the dynamics. Iterating Eq.~\eqref{eq: linear_system} from $\x_0$ at time $t_0$ gives a closed-form trajectory equation}

\begin{equation}\label{eq:solve-ds}
\x(t| \x_0,t_0) = \A^{t-t_0} \x_{0}  + \sum\limits_{j=0}^{t-t_0-1}  \left( \A^{j}\boldsymbol{B} \y(t-j) + \A^{j}\boldsymbol{\xi}(t- j) \right).
\end{equation}

\subsubsection{Pullback attractor of the model}

\added{An attractor of an autonomous system is the set of states the system approaches asymptotically, and it does not depend on time. A \emph{pullback attractor}~\citep{cheban2004global,Carvalho2013,Lucarini2017} extends this to non-autonomous (e.g., forced) systems. At each time $t$, it is the set of states reached once the initial conditions are forgotten, under \textit{the same non-autonomous changes} before $t$, so it retains the transient response to these changes. 
For our model with fixed $\y(t)$ and $\boldsymbol{\xi}(t)$, it is the set of limits of Eq.~\eqref{eq:solve-ds} over all initial states $\x_0$ as $t_0$ is pulled back to $-\infty$,}
\begin{align}
\begin{aligned}\label{eq:decompose_ds}
\x(t| \x_0,t_0) \underset{t_0 \to -\infty}{\xrightarrow{\hspace{1.2cm}}} &\underbrace{\sum\limits_{j=0}^{\infty}  \A^{j}\boldsymbol{B} \y(t-j)}_{\text{forced response}} + \underbrace{\sum\limits_{j=0}^{\infty}  \A^{j}\boldsymbol{\xi}(t-j)}_{\text{internal variability}}\\
&= \hspace{.4cm}\x^{(f)}(t) \hspace{.6cm}+\hspace{.55cm}\x^{(i)}(t).
\end{aligned}
\end{align}
Assuming that the stochastic model is stable, \added{(i.e., all eigenvalues of $\A$ lie in the unit circle)}, the first term in Eq.~\eqref{eq:solve-ds} disappears, \added{so the pullback attractor reduces to a single trajectory, independent of $\x_0$. It splits into} the forced response \added{and the} internal variability\added{, obtained by passing the forcing and noise histories through the same} decaying filter $\sum_{j\ge 0}\A^{j}$. \added{Each is thus} the system's fading memory of its drivers. Slowly decaying modes of $\A$ retain a longer memory and therefore dominate both terms.
\added{Since $\boldsymbol{\xi}$ has zero mean, the ensemble mean of the internal variability is zero. In a single realization, however, the weighted sum does not vanish and represents a process with finite variance and the timescales of $\A$, although the noise is white.}

\added{We assume that the data lie on the pullback attractor, i.e., that their initial conditions are forgotten. This holds for observations, since the climate system has always been externally forced~\citep{jansen2007palaeoclimate}, and for ESM simulations, which start after a spin-up~\citep{Eyring_2016}. We compute $\x^{(f)}(t)$ from the forcing history and take $\x^{(i)}(t)=\x(t)-\x^{(f)}(t)$ as the residual, which thus also holds any incomplete spin-up and any response to forcing missing from or misrepresented in $\y(t)$.}

\added{To compute $\x^{(f)}(t)$, the first term of Eq.~\eqref{eq:decompose_ds}, we run the deterministic part of the fitted model forward from a zero state and discard the first $s$ steps, the transition time. The initial error decays with the decay times of $\A$, so we choose $s$ much longer than them, and thus $s$ is set by the fitted model, whose decay times we know, not by the unknown timescales of the climate system. This requires a \emph{forcing history} that starts $s$ steps before the target interval. Algorithm~\ref{alg:algorithm1} summarizes the estimation of $\A$, $\boldsymbol{B}$ and the forced response.}

\added{Our approach relies on three key assumptions: linear dynamics ($\boldsymbol{A}$), a linear response to forcing ($\boldsymbol{B}$), and a collapsed pullback attractor. This third condition requires that initial conditions have decayed in the data and that the forcing history exceeds the fitted model's memory. How well they hold depends on the system and on the record length. The synthetic example (Section~\ref{sec:synthetic}) satisfies the first two exactly and thus isolates the errors of estimating the forced response from a single, finite and noisy record, while the ESM ensembles (Section~\ref{sec:forced_response}) test the first two jointly, as further discussed in Section~\ref{sec:synthetic}.}

\subsubsection{Data Decomposition}\label{sec:decomposition}

\added{Assuming $\boldsymbol{A}$ is diagonalizable (see Supplementary Material), we compute its eigendecomposition $\boldsymbol{A} = \boldsymbol{W}\boldsymbol{\Lambda}\boldsymbol{W}^{-1}$. This decomposition lies at the core of both DMD~\citep{kutz2016dmd} and DMDc~\citep{proctor2016dynamic}, which traditionally focus only on the deterministic dynamics.}

\paragraph{DMDc: Spatial Modes and Timescales} The eigenvectors \added{$\boldsymbol{w}_k$} of $\boldsymbol{A}$ are \added{the} dynamic modes.
The evolution of these dynamic modes has distinct decay times and oscillation frequencies encoded in the complex eigenvalues $\lambda_{k}$ on the diagonal of the matrix $\boldsymbol{\Lambda}$:
\begin{equation}\label{eq: eigenvalues}
    \lambda_k = r_k e^{i\theta_k} \text{ where }
r_k= |\lambda_k|,\quad 
\theta_k = {\arg\lambda_k} .
\end{equation}
Specifically, the decay time $T_k = -\frac{1}{\log{r_k}}$ and the oscillation frequency $f_k=\frac{\theta_k}{2\pi}$, both measured with respect to the evolution time step size $\tau$ of the operator $\A$. The dynamic modes with the largest decay times represent the longest-lived structures, which therefore dominate the model dynamics and are the objects of interest and interpretation. 

\paragraph{\methodname: Forced and Internal Components} 

\added{First, we project the data onto the basis of dynamic modes:}
\begin{equation}\label{eq-z}
\z(t) = \boldsymbol{W}^{-1} \x(t).
\end{equation}
\added{Inserting the pullback attractor~\eqref{eq:decompose_ds} splits it into forced and internal parts}
\begin{equation}
\label{eq:z-decompose-ds}
\z(t)  = \boldsymbol{W}^{-1} \x^{(f)}(t)+\boldsymbol{W}^{-1} \x^{(i)}(t)
=  \z^{(f)}(t) +\z^{(i)}(t).
\end{equation}
\added{Because $\boldsymbol{\Lambda}$ is diagonal, each mode $z_k(t)=z_k^{(f)}(t)+z_k^{(i)}(t)$ evolves independently in the model. Note that $z_k^{(f)}(t)$ sums the past forcing projected onto mode~$k$, weighted by $\lambda_k^{j}$ at lag~$j$, and $z_k^{(i)}(t)$ sums the projected noise in the same way (Supplementary Section~1e). In practice, we do not evaluate these sums but project $\x^{(f)}(t)$ and the residual $\x^{(i)}(t)$ onto the modes (Eq.~\eqref{eq:z-decompose-ds} and Algorithm~\ref{alg:algorithm1}).}

\begin{algorithm*}[t!]
\caption{Pullback DMDc with transition time $s$}\label{alg:algorithm1}
\begin{algorithmic}[1]
 
\Require  system states $\x(s+1), \x(\added{s+}2), \dots, \x(T) $, forcings $\y(1), \y(2), \dots, \y(T)$\added{, transition time $s$ (forcing history)}
 
\medskip
\State $\boldsymbol{A}, \boldsymbol{B} = \underset{\boldsymbol{A}, \boldsymbol{B}}{\mathrm{argmin}} \sum_{t=s+2}^{T} \left\|\,\boldsymbol{A}\x(t-1) + \boldsymbol{B} \y(t) - \x(t)\,\right\|_2^2$
 
\medskip
\State $\hat\x_0 = \boldsymbol{0}$
  \Comment{initial condition}

\medskip
\For{$j = 1, \ldots, T$}

\medskip
  \State $\hat \x_j = \boldsymbol{A}\,\hat\x_{j-1} + \boldsymbol{B}\,\y(j)$
    \Comment{estimate pullback attractor}

\medskip
\EndFor

\medskip
\State $\boldsymbol{\Lambda},\,\boldsymbol{W} = \text{eig}(\boldsymbol{A})$ 
  \Comment{eigenvalue decomposition}

\medskip
\State $\hat \X_f = [\, \hat{\x}_{s+1}, \,  \hat{\x}_{s+2}, \, \dots,\,  \hat{\x}_T \, ]$
\Comment{estimated forced response}

\medskip
\State $\X = [\, \x(s+1), \, \x(s+2), \, \dots, \, \x(T)\, ]$
\Comment{system states}
 
\medskip
\State \textbf{return}
  $\underbrace{\boldsymbol{W}}_{\text{spatial modes}}$,\quad
  $\underbrace{\hat{\boldsymbol{X}}_f}_{\text{forced response}}$, \quad
  $\underbrace{\boldsymbol{W}^{-1}\boldsymbol{X}}_{\text{full time series}}$,\quad
  $\underbrace{\boldsymbol{W}^{-1}\hat{\boldsymbol{X}}_f}_{\text{forced time series}}$,\quad
  $\underbrace{\boldsymbol{W}^{-1}\boldsymbol{X} - \boldsymbol{W}^{-1}\hat{\boldsymbol{X}}_f}_{\text{internal variability}}$

\end{algorithmic}
\end{algorithm*}

\added{Multiplying Eq.~\eqref{eq-z} by $\boldsymbol{W}$ then decomposes the data as}
\begin{equation}\label{eq-pullbackdmdc}
\x(t) = \boldsymbol{W} \z(t) = \sum_{k=1}^{M} \boldsymbol{w}_k z^{(f)}_k(t)+\sum_{k=1}^{M} \boldsymbol{w}_k z^{(i)}_k(t).
\end{equation}
Unlike a standard empirical orthogonal function (EOF) decomposition~\cite{lorenz1956empirical} or DMD, every spatial pattern comes with a \emph{pair} of physically distinct time series, one forced \added{($z_k^{(f)}(t)$)} and one internal \added{($z_k^{(i)}(t)$)}, so the same mode can be interrogated both for how it responds to forcing and for how it varies internally.

\paragraph{Rotation of complex-conjugate mode pairs}
\added{Pairs of modes with complex-conjugate eigenvalues describe damped oscillations in a real two-dimensional subspace, although their eigenvectors are complex. We represent each such pair by two real, orthogonal patterns $\widehat{\boldsymbol{w}}$ obtained by a 2D principal component analysis (PCA)~\citep{hotelling1933analysis} rotation within the subspace, with the corresponding internal and forced real time series $\widehat{z}^{(i)}(t)$ and $\widehat{z}^{(f)}(t)$ (see the Supplementary).}

\subsection{Parameters and Data Processing}\label{sec:data_processing}

For each ESM \added{member} and the \added{reanalysis}, we compute anomalies and \added{run area-weighted PCA, so that the PCs are the input to all methods. 
Hereafter, $\x(t)$ denotes the leading $M$ PCs, and the columns of $\U$ are the corresponding EOFs, which map the results back to the grid. Since the EOFs are orthonormal, the model fitted to the PCs is the same linear model in the original space, restricted to the retained EOF subspace, with $\U\A\U^{\top}$ and $\U\boldsymbol{B}$ in place of $\A$ and $\boldsymbol{B}$. Its pullback attractor, and hence the forced response, maps back by $\U$, and its eigenvectors $\U\boldsymbol{w}_k$, with the same eigenvalues $\lambda_k$, are the spatial patterns of the dynamic modes.
We keep} $20$ PCs for \added{temperature, following ForceSMIP submissions~\citep{Wills2026},} and $200$ for \added{SLP, whose forced signal is weak relative to its internal variability (Section~S1g). Our ground truth is the ESM ensemble mean, smoothed for SLP with a $3$-year running mean.}

\added{\methodname~and LR take as default input the global-mean annual effective radiative forcing (ERF) from IPCC AR6~\citep{IPCC_2021_WGI_Ch_7}, linearly interpolated to monthly resolution, with one value per year for each forcing agent and no spatial pattern. 
Two variants below use two or three inputs, each with its own column of $\boldsymbol{B}$, to test whether more detailed forcing information improves the estimate, namely} CO$_2$ and volcanic forcing\added{, which dominate the trend and the year-to-year variations of the forcing} (\methodname(2D)), and greenhouse gases, volcanic forcing, and aerosols (\methodname(3D)). \added{The 3D variant replaces CO$_2$ by all greenhouse gases and adds aerosols, two changes of similar magnitude and opposite sign in the forcing~\citep{IPCC_2021_WGI_Ch_7}.} All forcing variables are standardized over the analysis period\added{, so only their temporal patterns matter}.

For \added{the} dynamical methods (\methodname\ variants, LIM, and LIMopt), we fit $\boldsymbol{A}$ 
\added{at a lag of $\tau=3$ months, following some ForceSMIP submissions~\citep{Wills2026},}
while estimating the pullback attractor at a monthly sampling rate. \added{For \methodname, the transition time $s$ is a $100$-year forcing history before the analysis period, far longer than all decay times of the fitted $\A$ (Figs.~S11 and S28). The implementation details, sensitivity tests for the number of PCs, the lag and the forcing history, and other method settings are in the Supplementary.}

\section{Results}\label{sec:results}

\added{In this section, we first test \methodname~on a synthetic system whose forced response is known (Section~\ref{sec:synthetic}). We then } apply \methodname~versions and the baselines to monthly anomalies of global surface air temperature (SAT), SAT over the ocean between $40^\circ S$ and $65^\circ N$ (OSAT), and global sea level pressure (SLP) from the $20$th-century reanalysis v3~\citep{slivinski2019towards} and $4$ different ESMs from the MMLEA-v2 project~\cite[Table 2]{maher2025updated}. \added{We use the reanalysis as a proxy for the observed climate in this interval, since it assimilates observations~\citep{slivinski2019towards}.}
\added{The four ESMs,} CanESM5~\citep{swart2019canadian} \added{(65 members),} MIROC6~\citep{tatebe2019description}, CESM2\added{,} and MPI-ESM~\citep{muller2018higher, mauritsen2019developments} \added{(50 members each), were chosen for their large ensembles with 40+ members, which allow a reliable estimate of the} ``ground truth'' forced response.
The target time interval for our experiments is $1850$ to $2014$, 
\added{and the OSAT variable is chosen in addition to global fields because it excludes high latitudes and land, where a slow change of the seasonal cycle remains in the anomalies after subtracting a stationary climatology on this interval (see \textit{Seasonal-cycle change in SAT and SLP} in Section~\ref{sec:forced_response}). }

{\add

\subsection{Synthetic example}\label{sec:synthetic}

Before the climate application, we test \methodname~on a linear stochastic system with forcing of the form Eq.~\eqref{eq: linear_system} and a known forced response (see Fig.~S1). The system mimics several features of the climate dynamics in an idealized zonal-mean setting, following \citet{Kravtsov2022}. It combines a slow multidecadal mode, an \added{El Ni\~{n}o--Southern Oscillation} (ENSO)-like oscillating pair and fast residual variability, all driven by a forcing with a warming-like rise and a mid-twentieth-century pause, and observed with weak measurement noise. Notably, the internal variability of each mode arises from stochastic dynamics in Eq.~\eqref{eq: linear_system}, where the damped linear dynamics integrate white dynamical noise, as in \citet{Hasselmann1976}. Each mode therefore fluctuates on its own decay timescale, although the noise itself is white (e.g., the multidecadal internal variability time series in Fig.~S2a). The detailed equations and setting are in Supplementary Section~2.

Starting from a reference configuration, we systematically vary the signal-to-noise ratio (SNR), the slow-mode decay time $\tau_1$, the measurement noise level, the spatial overlap between modes, and the temporal profile of the forcing. For each configuration, we evaluate each method across $1{,}000$ single realizations. \methodname~recovers the forced response with a correlation close to $1$ across joint variations in SNR and $\tau_1$ and matches or exceeds LR in every configuration, whereas the LIM-based methods lose skill at low SNR (Fig.~\ref{fig:synthetic} and Fig.~S3). Since the skill is measured against the exact forced response on single realizations, it does not reflect overfitting. Measurement noise and spatial overlap degrade \methodname~at most to the LR level. We also assess how well the method identifies the characteristics of the slow mode. This mode carries most of the forced response but shares its spatial pattern with slow internal variability, so the two can only be separated in time, over a record that spans a few decay times. Its pattern is reliably extracted in almost all configurations, but its decay time is uncertain because of the finite record, and can be greatly shortened by the measurement noise and spatial overlap, so it should be read as a lower bound (Supplementary Section~2, Figs.~S2--S4).

\begin{figure}[t]
  \centering
  \includegraphics[width=\textwidth]{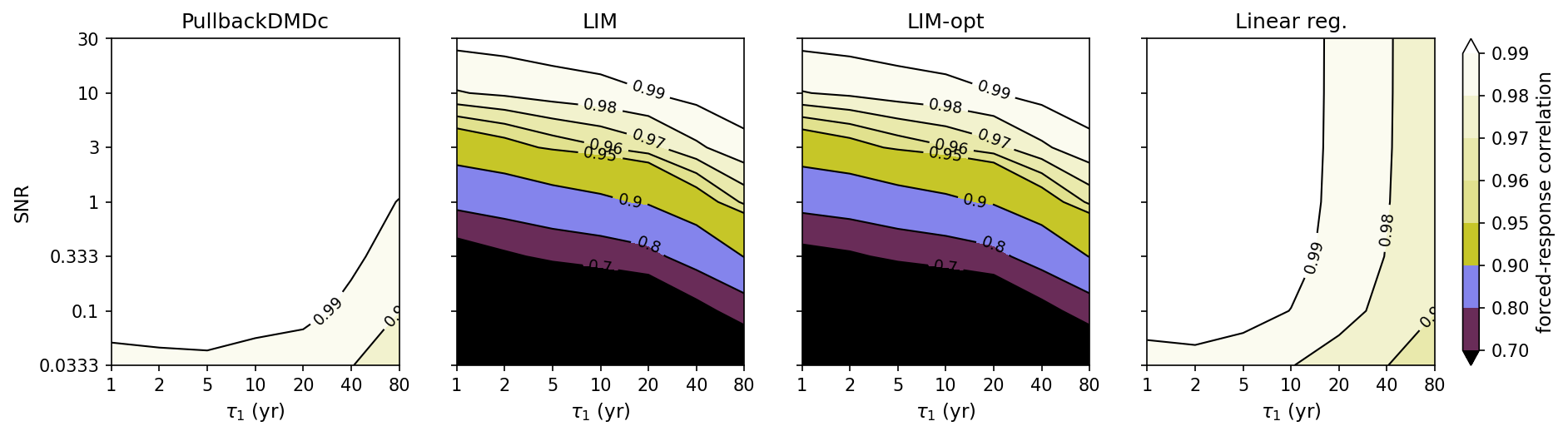}
  \caption{Correlation between the forced response estimated from single realizations and the exact forced response of the synthetic system (median value over $1000$ noise realizations is plotted), as a function of the SNR and the slow timescale $\tau_1$, for \methodname, LIM, LIMopt and LR (Linear reg.). Contours are labeled with their values. Details are given in Supplementary Section~2.}
  \label{fig:synthetic}
\end{figure}

The synthetic system is linear, like all the methods compared here, and it lets us assess the key error sources in estimating the forced response from a single, finite, noisy record. Pullback attractor of nonlinear systems, such as simple climate models \citep{Daron2013,Lucarini2024} or ESMs themselves, may be represented by  a set with a time-dependent distribution of states rather than a single trajectory. As noted in Section~\ref{sec:methods}, for ESMs, we define the forced response operationally as the mean over this distribution, estimated by the ensemble mean. We hypothesize that \methodname~applied to a single realization approximates this mean. The least-squares fit of Eq.~\eqref{eq: linear_system} gives the best linear one-step predictor of the state from its previous value and the forcing, so its pullback forced response can be read as a linear approximation of how the mean state follows the forcing history, in the spirit of linear response theory \citep{Lucarini2017}.
The nonlinear, fast dynamics that the linear model cannot represent then act as effective, possibly state-dependent noise, approximated by a stochastic term in \eqref{eq: linear_system}. Our experiments with measurement noise suggest that it mainly shortens the decay times estimated by \methodname, while the forced response and the spatial patterns of the modes stay robust.
We test these hypotheses for forced-response estimation with the ESM large ensembles in Section~\ref{sec:forced_response}, and in Section~\ref{sec:results_decomposition} we interpret the ESMs through this linear stochastic approximation.}

\subsection{Forced-response estimation}\label{sec:forced_response}

We apply each forced-response estimation method to each member of the MMLEAv2 ensemble, mimicking the estimation of forced response from a single observational climate realization. Using the large ensemble for each ESM, we compute the ground-truth forced response as defined in Section~\ref{sec:data_processing}.

\begin{figure}[p]
    \centering
    \begin{subfigure}{0.47\linewidth}
        \centering
        \includegraphics[width=\linewidth]{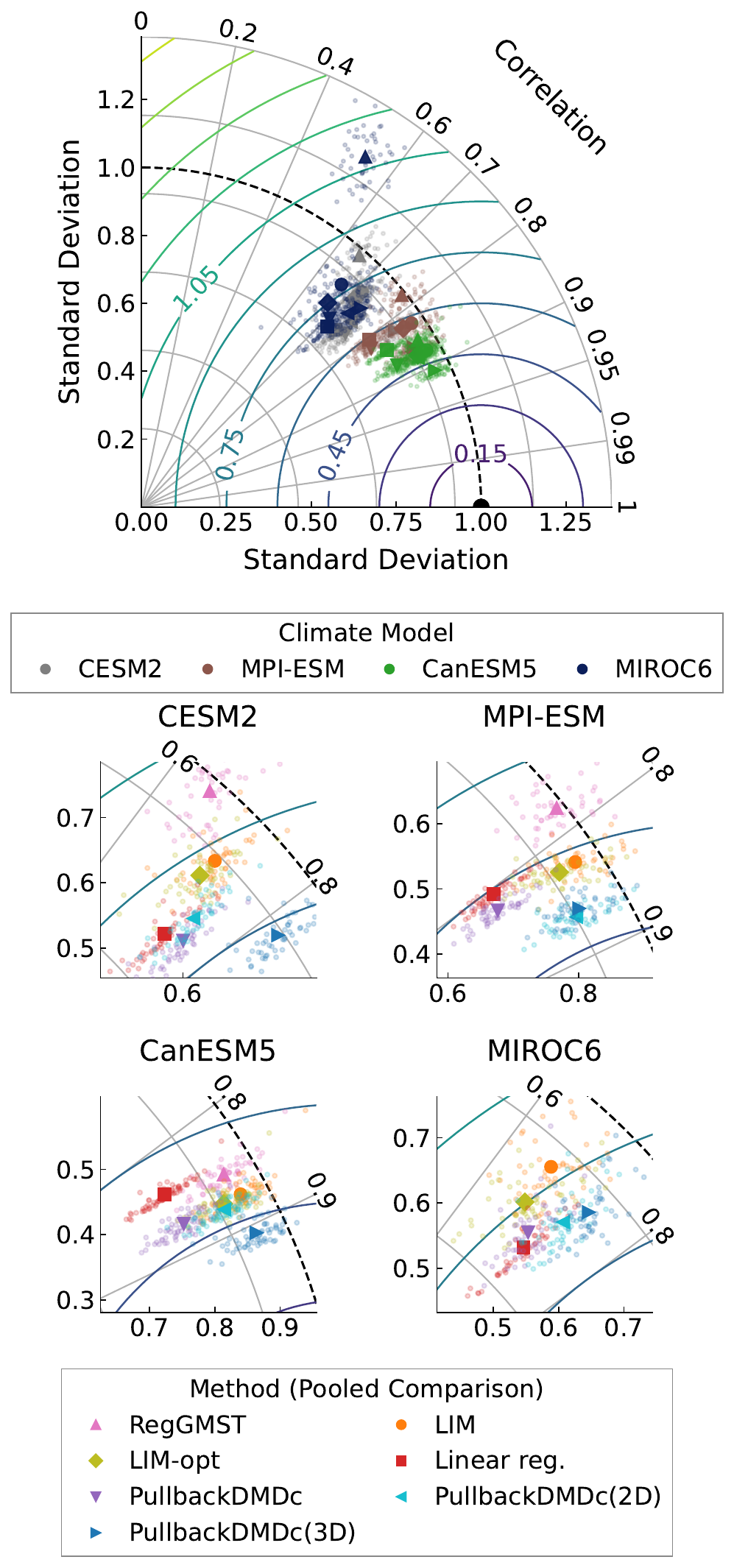} 
    \end{subfigure}
    \hfill
    \begin{subfigure}{0.47\linewidth}
        \centering
        \includegraphics[width=\linewidth]{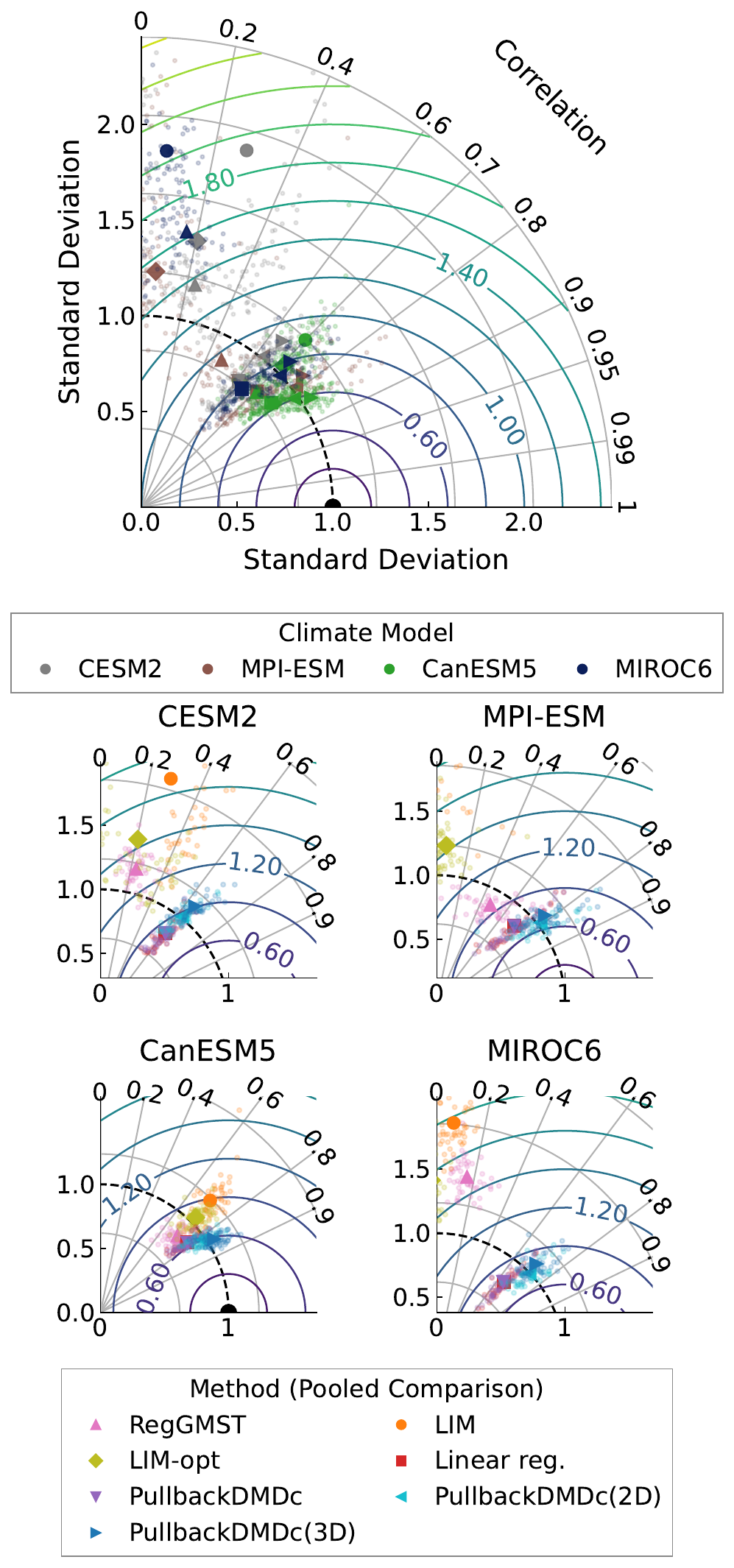} 
    \end{subfigure}
    \caption{Taylor diagrams comparing forced-response estimates from \methodname~variants and baseline methods against the large-ensemble mean \added{of the same ESM} for (left) ocean surface air temperature (OSAT) and (right) sea level pressure (SLP). \added{For each colored point, distance from the reference point (1,0) indicates the centered root-mean-square difference (RMSD)}, angle from the x-axis indicates correlation, and the \added{distance from the origin indicates standard deviation}. Each point compares a raw spatio-temporal forced-response estimate with the ground truth. All quantities are normalized by the standard deviation of the ground-truth forced response. Large points indicate forced-response estimation skill across all ensemble members. Small translucent points indicate the forced-response estimation skill of individual ensemble members. The first row of large Taylor diagrams compares estimations over all methods for different ESMs. The bottom $2 \times 2$ plots compare the performance of each method \added{per ESM.}}
    \label{fig:taylor_raw}
\end{figure}

\paragraph{Skill of \methodname~versus baselines}
We utilize Taylor diagrams to summarize forced-response model performance across ESMs and methods over the $1850$--$2014$ interval for raw OSAT/SLP (Fig.~\ref{fig:taylor_raw}) and SAT (Fig. S5) predictions. 
Over the same interval, we also assess the extracted linear trend coefficients for all three climate variables (Fig. S6)\added{, and their trends over the last $30$ years, $1985$--$2014$ (Figs.~S7--S9)}. 
\added{Then, w}e evaluate raw forced-response model performance for all variables over the $1950$--$2014$ interval, \added{similar} to the ``Tier1'' setting in ForceSMIP~\citep{Wills2026} (Fig. S10).

\added{A}ll diagrams indicate that both \methodname~and LR outperform all other baselines that do not use external forcing predictors, confirming that forcing predictors are an important ingredient for accurate forced-response estimation. Among methods that use forcing information, at least one~\methodname~variant matches or exceeds LR skill across all variables, suggesting that explicitly modeling the dynamical forced response as a pullback-attractor component corresponding to $\y(t)$ can add skill beyond a purely thermodynamic linear response. 
\added{\methodname~yields shorter leading decay times in $\A$ than LIM (Fig.~S11), as in the synthetic example. This is consistent with the least-damped LIM mode absorbing the forced response, which lengthens its decay time. In \methodname, the forced response passes through the dynamics instead, so the input matrix $\boldsymbol{B}$ of \methodname~captures the effect of the forcing over a single step, rather than the full forced-response fingerprint estimated by LR (Fig.~S12).}

\begin{figure}[t!]
    \centering
    \includegraphics[width=\linewidth]{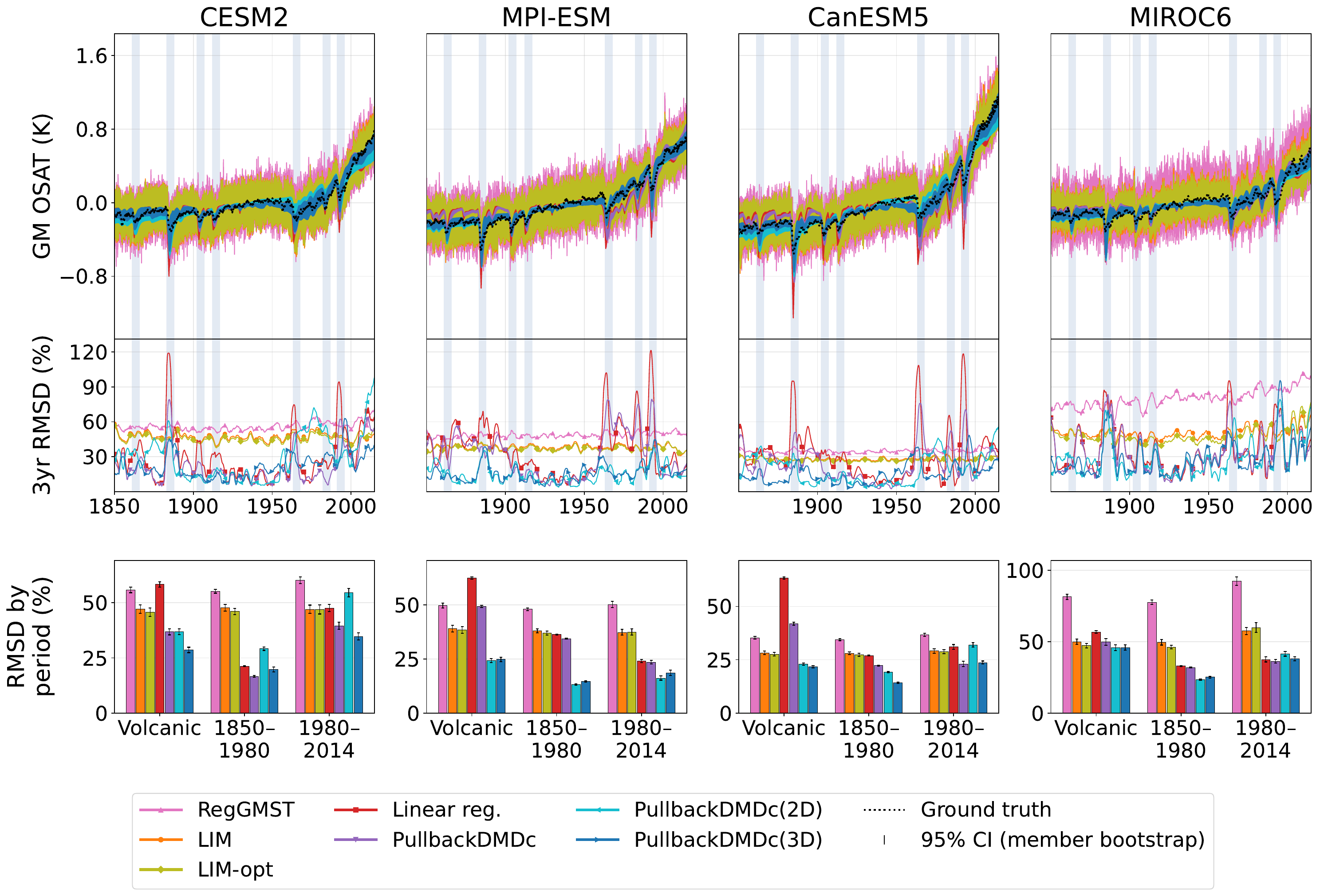}
    \caption{Global mean ocean surface air temperature (OSAT) forced response estimates from \methodname~variants alongside baseline methods. \textit{Top row}: global mean time series of forced-response predictions for each ensemble member. \textit{Middle row}: root-mean-square deviation (RMSD) from the ground truth, computed at each time step by averaging across all ensemble members within a centered 3-year window, expressed as a percentage of the ground-truth standard deviation over the full period. Light-blue vertical stripes mark $5$-year windows following $7$ major volcanic eruptions. \textit{Bottom row}: RMSD aggregated over three disjoint subsets of time steps. \added{Error bars are 95\% intervals from bootstrapping ensemble members (2000 resamples, the same resamples for every method).}}
    \label{fig:gms}
\end{figure}

\paragraph{Uncertainty of forced-response estimations}
The spread of the forced-response estimation across ensemble members is itself informative, since it measures how sensitive each method is to internal variability. We use each method's internal-variability uncertainty to confirm known facts about forced-response behavior across climate variables (Fig.~\ref{fig:taylor_raw}). For all methods, this distributional perspective highlights a difference between OSAT/SAT and SLP: forced-response estimates exhibit substantially higher variance across ensemble members for SLP than for OSAT/SAT. This reflects the weaker trend and larger internal variability in SLP than the strong global warming trend observed in the temperature time series~\citep{Lee2021AR6Chapter4}. A similar pattern is also apparent when comparing the global mean forced-response time series for OSAT (cf. Fig.~\ref{fig:gms}) with those for SAT \added{and for SLP averaged south of $65^\circ$S} (cf. Fig. S13--S14).\footnote{\added{For SLP, we use the mean south of $65^\circ$S instead of the global mean, because the forced SLP trend in the ESMs is strong there and does not average out within this region, unlike over the globe (Fig.~S8).}}

\added{Method performance also varies across ESMs. For example, the OSAT correlation reaches up to $0.9$ for CanESM5, which has the highest climate sensitivity and thus the strongest warming trend (cf. Fig.~\ref{fig:gms}). In general, a higher sensitivity produces a stronger forced signal, yielding higher reconstruction skill.}

The next important conclusion we draw from the distributions is that all baseline methods except LR exhibit substantially higher uncertainty than~\methodname, again reflecting their lack of strong external predictors of the forced response. As a result, they produce very noisy estimates with a much larger spread (cf. Fig.~\ref{fig:gms}). The performance of these methods is sometimes much worse for SAT and SLP variables (cf. Fig. S13--S14). 
\added{The forced SLP response is largest over the polar caps, particularly over Antarctica, as shown by both the forced trends (Fig.~S8) and the LR fingerprint (Fig.~S12). Over Antarctica, the forced SLP falls mainly after the 1960s, a trend toward the positive phase of the Southern Annular Mode (SAM)~\citep{thompson2002interpretation}. Only the methods with forcing information follow this decline (e.g., LR and \methodname~variants). In contrast, RegGMST, LIM, and LIMopt have errors exceeding the forced signal after 1980 in three of the four ESMs. Among these, \methodname(2D) has the smallest or near-smallest error in almost all ESMs and periods, with its clearest advantage in volcanic intervals, where LR and the default \methodname~have larger errors (Fig.~S14). In the Arctic, by contrast, the forced SLP signal north of $65^\circ$N is weak compared with internal variability, and all methods have errors comparable to or larger than the signal itself (Fig.~S15).}
\added{For SAT, the LIM baselines perform worst in MIROC6 (Fig.~S5). We hypothesize that this is because MIROC6 has the weakest forced SAT trend of the four ESMs (Fig.~S7) and hence the lowest SNR, to which LIM is particularly sensitive, as the synthetic example shows (Section~\ref{sec:synthetic}). In addition, the least-damped mode of LIM is oscillatory in some cases, including many MIROC6 SAT members (Fig.~S11), while the LIM estimate assumes a real slow mode and keeps only the real part of a complex one (Supplementary Section~1a).}
  
All the conclusions here also hold \added{when repeating the same analysis in $1950$--$2014$}, except that the errors of all methods increase (cf. Fig. S10, S16--S18). This is expected because the sample size (e.g., the number of time steps) is reduced by almost $2.5$ times, increasing uncertainty. \added{Furthermore, over this interval the forcing and forced response are closer to a linear trend, so the forcing history, which \methodname~uses and LR does not, matters less, and its advantage over LR decreases.}

\paragraph{Choice of forcing in~\methodname}

Not all~\methodname~variants perform equally. Specifically,~\methodname~is sensitive to input forcing information. First,~\methodname~\added{consistently} benefits from separating the volcanic contribution from other slow contributions (e.g., greenhouse gases). \added{Using 2D/3D forcing inputs} substantially reduces global mean root mean square deviation (RMSD) and raw predictive skill (cf. \added{Figs.~\ref{fig:taylor_raw},~\ref{fig:gms}}). This improvement reflects that volcanic and long-term forcing activate distinct response patterns on different timescales. By explicitly including volcanic forcing,~\methodname(2D)/(3D) better capture eruption-related cooling and reconstruct the temporal response structure under both weak and strong warming trends (cf. rows 2–3 of Fig.~\ref{fig:gms}).

Next, there is a difference in the optimal combinations of long-term forcing predictors depending on the target variable:~\methodname(2D) performs best for SLP and~\methodname(3D) performs best for OSAT and SAT (cf. Figs.~\ref{fig:taylor_raw},~\ref{fig:gms}, S5, S6, and S13--S14). This suggests that temperature fields (OSAT/SAT) benefit from a more complete representation of aerosol and greenhouse-gas forcings. Conversely, adding these forcing inputs slightly degrades SLP performance, likely because the lower SNR in circulation fields makes the model more susceptible to overfitting.  \added{This is consistent with detection and attribution results~\citep{bindoff2013detection}. Observed warming is attributed mainly to greenhouse gases, partly offset by other anthropogenic forcings dominated by aerosols, and the two responses can often be estimated separately. The positive SAM trend in the austral summer, in contrast, is attributed in part to stratospheric ozone depletion, with greenhouse gases also contributing. Neither variant includes ozone forcing, which may explain why the 3D variant does not improve on the 2D variant for SLP.}

The optimal \added{choice of the forcing set} varies little across ESMs \added{(see Figs.~\ref{fig:taylor_raw}, S5--S6, S10)}, suggesting robustness to the choice of climate generator, though the uncertainty regions overlap more for the less sensitive models, such as MIROC6 and MPI-ESM \added{(see OSAT in Fig.~\ref{fig:taylor_raw})}. Because the ESM ensemble mean provides a per-model ground truth, this comparison doubles as a principled way to select forcing predictors for~\methodname. We leave a systematic investigation of optimal forcing to future work, as this choice likely varies with the target variable and dataset parameters.

\begin{figure}[H]
    \centering
    \includegraphics[width=\linewidth]{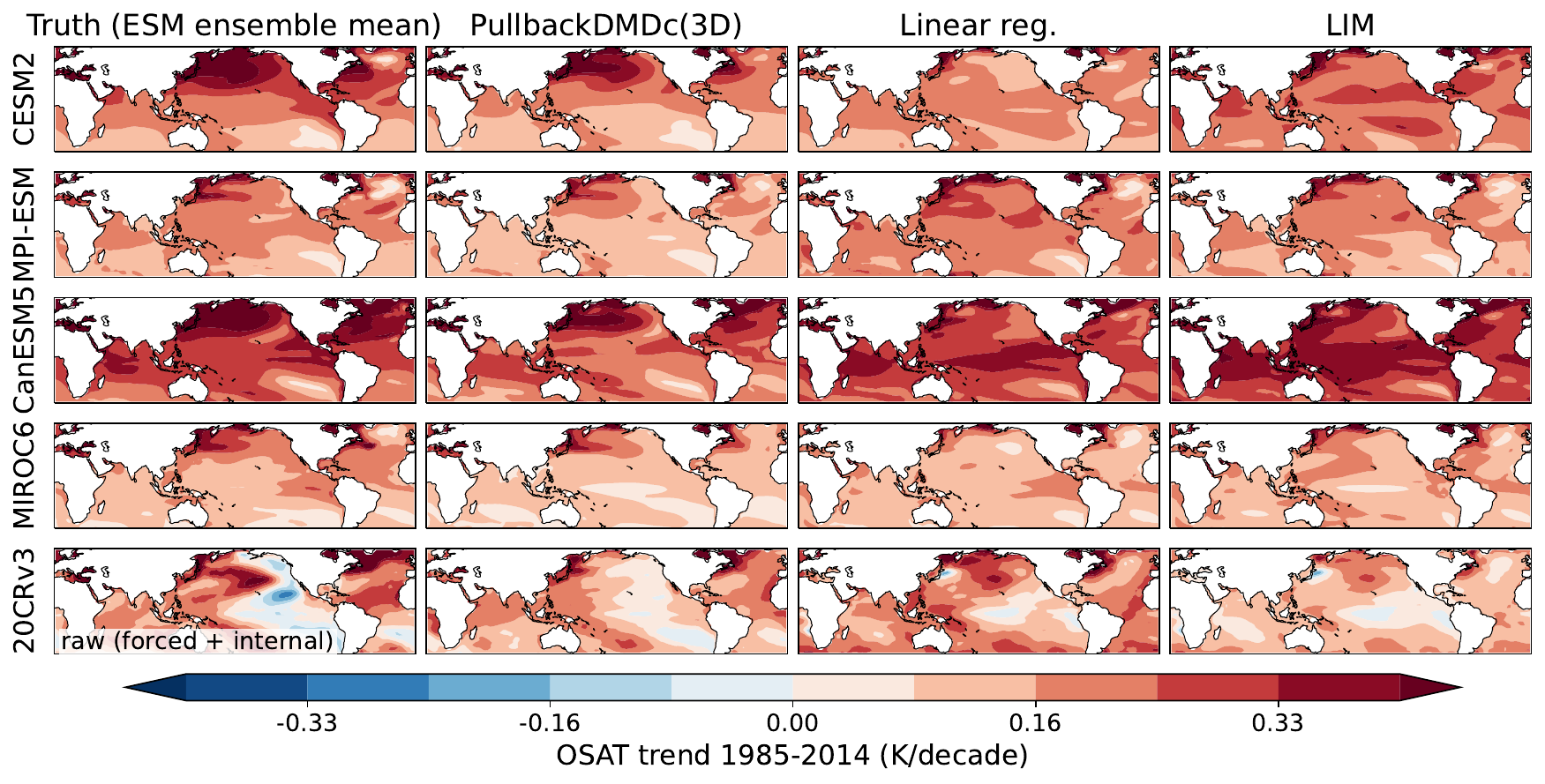}
    \caption{\added{Linear trends of the OSAT forced response over $1985$--$2014$ (K/decade). Left column: ESM ensemble mean (truth) and, for 20CRv3, the raw data (forced + internal). Other columns: forced-response trends estimated by \methodname(3D), LR and LIM, averaged over members for the ESMs. The color scale is saturated at the $95$th percentile of the absolute trends over all panels.}}
    \label{fig:trend-map-osat}
\end{figure}

\paragraph{Recent forced trends}
\added{Here we examine the spatial distribution of the forced trends over $1985$--$2014$, estimated by the best variant for each variable, \methodname(3D) for OSAT and SAT and \methodname(2D) for SLP, together with LR and LIM. For OSAT (Fig.~\ref{fig:trend-map-osat}), the true trends are positive almost everywhere, with maxima in the extratropical North Pacific and North Atlantic, and, consistent with climate sensitivity, they are largest in CanESM5 and CESM2 and weakest in MIROC6. \methodname(3D) and LR both recover this pattern, and \methodname(3D) better captures the North Pacific maximum in CESM2 and CanESM5. For SAT (Fig.~S7), the true trends show Arctic amplification and a land--ocean contrast, which \methodname(3D) and LR again reproduce, whereas LIM fails for CESM2 and produces a spurious cooling in the eastern tropical Pacific. For SLP (Fig.~S8), the true trends project onto the positive phase of the Southern Annular Mode (SAM). \methodname(2D) recovers this pattern, LR overestimates its amplitude (especially over Antarctica), and LIM nearly misses it in MPI-ESM and MIROC6.}

\added{The Taylor diagrams of the single-member trend estimates (Fig.~S9) summarize these results. For OSAT and SAT, the forced response is nearly linear over this short window, so LR performs about as well as \methodname(3D), which underestimates the trend amplitude in MPI-ESM and MIROC6, and the advantage of \methodname(3D) over the full record (Fig.~S6) largely disappears. For SLP, \methodname(2D) has the smallest or equal-smallest error in all ESMs.}

\added{In 20CRv3 (bottom row of Fig.~\ref{fig:trend-map-osat}), the raw OSAT trend differs from the ESM forced trends mainly by a pattern resembling the negative phase of the Interdecadal Pacific Oscillation (IPO)~\citep{Henley2015}, with cooling in the eastern tropical and subtropical Pacific and strong warming in the western Pacific. \methodname(3D) removes almost all of this pattern, leaving enhanced warming in the North Atlantic and northwestern Pacific, while LR retains part of it. In the global mean, \methodname(3D) likewise rises smoothly, close to the ensemble means of CESM2, MIROC6 and MPI-ESM, and assigns the cold early 1900s and the warm 1940s to internal variability (Fig.~S19), in line with mode~$1$ (Section~\ref{sec:results_decomposition}). The eastern tropical Pacific cooling in 20CRv3 is part of the strengthening of the tropical Pacific zonal SST gradient and Walker circulation observed since $1979$, which models generally fail to reproduce~\citep{kociuba2015inability,wills2022systematic}. Whether internal variability explains this discrepancy, as \methodname(3D) suggests, or the models miss part of the forced response is debated~\citep{watanabe2021enhanced,wills2022systematic}. Moreover, the forced response of \methodname~is a linear response to the global-mean forcing time series, so it misses forced changes that leave little imprint on them, such as the response to the shift of aerosol sources toward Asia while the global-mean aerosol forcing changed little. Such a response must appear in the internal component, so the method cannot determine whether this cooling is forced or internal. SAT shows the same behavior over the OSAT region (Fig.~S7), and the estimates differ mainly outside it, near $60^\circ$S, where \methodname(3D) retains a cooling and LR and LIM give a warming.}

\added{For SLP (Fig.~S8), \methodname(2D) and LR give nearly identical positive-SAM trends, much stronger than in the ESMs, and LIM gives almost none. The mean south of $65^\circ$S shows that \methodname(2D) attributes nearly all of the raw decline since about 1900 to forcing, far more than any ESM simulates (Fig.~S19). This difference may arise because the ESMs underestimate the forced SAM response or because the methods attribute reanalysis inhomogeneities to forcing, and two studies support the latter. \citet{fogt2021extratropical} show that before 1957, 20CRv3 and other centennial reanalyses have significant pressure declines over Antarctica that station-based reconstructions do not support. Regressing the SAM index on forcing over $1850$--$2014$,~\citet{morgenstern2021southern} likewise finds a strengthening in 20CRv3 beyond the CMIP6 range in winter and spring, despite agreement over $1957$--$2014$, and attributes it to spurious early trends that project onto the steadily rising greenhouse gas forcing. Since \methodname(2D) and LR also fit the full record, such a decline would enter their forced estimates and inflate the recent trend.}

\paragraph{Seasonal-cycle change in SAT and SLP}

\added{Finally, for SAT and SLP, the forced estimates miss one part of the ground truth: the change of the seasonal cycle. Recall that we compute anomalies by subtracting a fixed climatology. However, on the long record, the seasonal cycle itself changes over time~\citep{pezzulli2005variability}. Any departure of a given year's seasonal cycle from the long-term mean, such as a change in its amplitude or phase, therefore remains in the anomalies with a 12-month period. Such changes are well-documented for surface air temperature. \citet{stine2009changes} quantified changes in the amplitude and phase of its annual cycle and compared them with climate model simulations, \citet{stine2012changes} characterized them over the Northern Hemisphere in both temperature and atmospheric circulation, and \citet{nigam2017striking} examined their mechanisms. More recently, \citet{holtanova2026global} showed that CMIP6 models tend to underestimate these changes at middle and high latitudes of the Northern Hemisphere, while largely agreeing with reanalyses on the smaller changes in the tropics and the Southern Hemisphere.}

\begin{figure}[H]
    \centering
    \includegraphics[width=\linewidth]{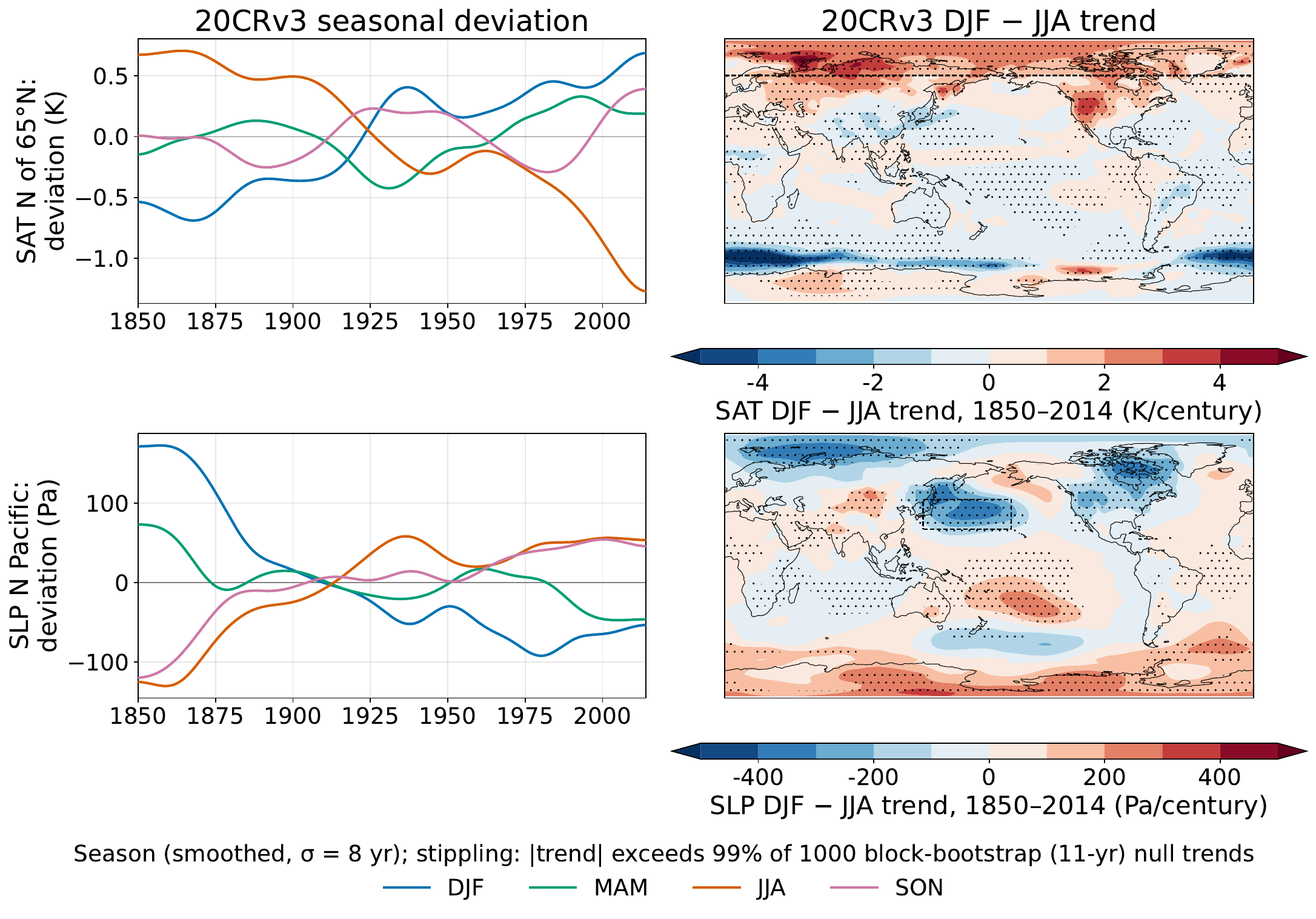}
    \caption{\added{Seasonal-cycle change in 20CRv3 for SAT (top) and SLP (bottom). Left: deviation of each season's mean anomaly from the annual-mean anomaly of the same year, averaged north of $65^\circ$N for SAT and over $25^\circ$--$45^\circ$N, $135^\circ$E--$165^\circ$W for SLP, \added{smoothed with a Gaussian filter ($\sigma = 8$~years).} Since the anomalies are taken with respect to a fixed $1850$--$2014$ climatology, these curves show how the seasonal cycle departs from its long-term mean. Right: linear trend of DJF$-$JJA over $1850$--$2014$. Stippling marks grid points where the absolute trend exceeds the 99th percentile of $1000$ trends of block-bootstrapped ($11$-yr blocks) detrended series.}}
    \label{fig:seasonal_drift}
\end{figure}

\added{In 20CRv3, the seasonal cycle changes most at high latitudes (Fig.~\ref{fig:seasonal_drift}, top). North of $65^\circ$N and over northern land, winters are warmer than summers, giving a positive DJF$-$JJA trend. In the north, most of this change occurs in the twentieth century and is fastest after about 1970. Near $60^\circ$S, the DJF$-$JJA trend is negative, so the austral winter (JJA) also warms relative to summer, although this may partly reflect the reanalysis sea ice, which is prescribed as a boundary condition rather than constrained by assimilated observations \citep{slivinski2019towards}. For SLP, the trend is largest over the North Pacific and high northern latitudes (Fig.~\ref{fig:seasonal_drift}, bottom), but much of the North Pacific change occurs before 1900, when few observations constrain the reanalysis. The forcing inputs of \methodname~and LR have no seasonal variation, so their forced estimates have nearly the same trend in every calendar month (not shown), and the change in the seasonal cycle can only stay in the internal component.}

\added{The ESMs also show a seasonal-cycle change consistent across members, so it is part of the forced response (Fig.~S20). Their ensemble means, however, reproduce mainly the Arctic part of the 20CRv3 pattern in SAT, with a similar or weaker trend north of $65^\circ$N. In SLP, they reproduce none of it, and over the North Pacific their trend even has the opposite sign.}

\added{The seasonal-cycle change affects our analysis in two ways. It is part of the ESM ground truth, yet \methodname~and LR cannot capture it, since their forced response follows the smooth forcing inputs, which carry no 12-month signal. This may slightly lower their skill, but it should hardly affect the estimated trends, since the change oscillates within each year and largely averages out over full years. The change also remains in the internal component, where it mixes with the dynamic modes differently in each dataset and hampers their comparison. The OSAT domain excludes the regions where the change is largest, and we therefore interpret the dynamic modes only for OSAT (Section~\ref{sec:results_decomposition}).}

\subsection{~\methodname~modes}\label{sec:results_decomposition}

\added{In the decomposition of Eq.~\eqref{eq-pullbackdmdc}, the leading modes (those with the slowest decay times) represent the data's dominant dynamics. Comparing them across datasets shows which patterns a model represents and how it distributes variability and forced response across time scales, relative to other models and to the reanalysis. We now illustrate such an evaluation using OSAT, as noted above.}

\paragraph{Reanalysis data decomposition}

\begin{figure}[t!]
    \centering
    \includegraphics[width=\linewidth]{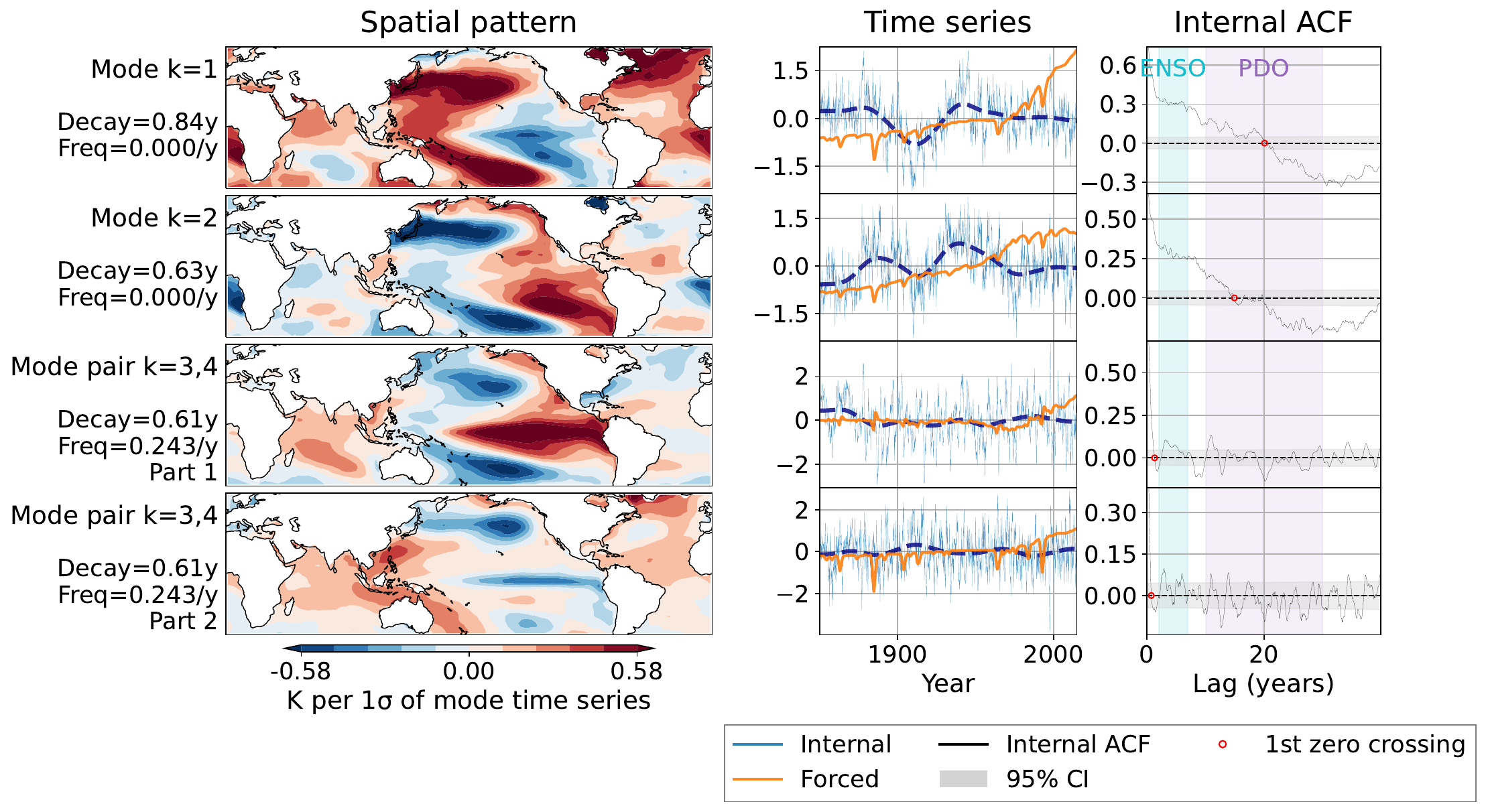}
    \caption{Leading \methodname~modes \added{(longest decay times)} from~\methodname(3D)~for ocean surface air temperature (OSAT) in  20CRv3 (reanalysis). Each mode is shown as a spatial pattern $\widehat{\boldsymbol{w}}$ (left), forced and internal component time series $\widehat{z}^{(f)}(t)$ and $\widehat{z}^{(i)}(t)$ (center), and the \added{ACF} of the internal component (right).  \added{The grey shading shows the pointwise $95\%$ white-noise confidence interval~\citep{bartlett1946theoretical}. The blue and purple windows mark nominal periods of ENSO ($2$--$7$ years)~\citep{Timmermann2018} and of the Pacific Decadal Oscillation (PDO) ($10$--$30$ years) where an oscillation of period $T$ would produce an ACF peak at lag $T$. Note that $10$--$30$ years is near the $15$--$25$-year band of~\citet{mantua2002pacific}.In general, these bands are visual guides only, since the PDO has no clear spectral peak~\citep{ipcc2021annex4}.} For each complex-conjugate pair of modes, we plot $2$ real patterns $\widehat{\boldsymbol{w}}$ and time series $\widehat{z}(t)$ obtained by 2D PCA rotation, see the details in Sec.~\ref{sec:decomposition}. The dashed blue line (middle column) \added{shows the internal component smoothed with a Gaussian filter ($\sigma = 8$~years, i.e. cutoff period near 40 years)}. The time series are normalized so that the variance of the sum of the internal and forced time series equals $1$, and the patterns are scaled so that, \added{when summed across all modes,} their product with the time series reconstructs the \added{O}SAT anomalies \added{up to EOF truncation}. Therefore, patterns show a signed contribution of the full mode to the data in \added{kelvin (K)}.}
    \label{fig:dmd_modes_side_by_side_20crv3}
\end{figure}

In the previous section, \added{the best forced-response estimates for OSAT were achieved by~\methodname(3D)}. We now consider its top four dynamic modes \added{in the reanalysis dataset}~(Fig.~\ref{fig:dmd_modes_side_by_side_20crv3}). 
The leading mode exhibits a global spatial pattern \added{that combines a warm phase of Atlantic Multidecadal Variability (AMV)~\citep{Knight2006a, Wyatt_2012} with a horseshoe in the Pacific resembling the negative phase of the IPO~\citep{Henley2015}, with larger Pacific loadings in amplitude and extent. This combination is consistent with the observed AMV pattern, which includes IPO-like cooling of the Pacific cold tongue, and with the negative correlation between AMV and the IPO~\citep{ipcc2021annex4}. We therefore interpret it as capturing multidecadal variability of both the Atlantic and the Pacific.}

Its internal time series component is dominated by multidecadal variability, with a negative phase spanning roughly $1900$-$1925$ followed by a positive phase from $1930$-$1960$ (see smoothed dashed curve, Fig.~\ref{fig:dmd_modes_side_by_side_20crv3}). \added{These phases match the cool and warm periods of observed AMV indices~\citep{ipcc2021annex4}. Accordingly, its autocorrelation function (ACF) remains positive up to a lag of about $20$~years, and its power spectrum is enhanced at periods longer than the ENSO band (Fig.~S21). 
This persistence far exceeds the mode's decay time ($0.84$~years). If the fitted model were exact, the ACF of the internal component would decay with this decay time. The internal component $\hat{z}^{(i)}(t)$ is, however, computed from the data through Eq.~\eqref{eq:z-decompose-ds} (Supplementary Material, Section~1e), so its ACF shows the persistence actually present, and the discrepancy indicates that the fitted decay time is biased low. The synthetic example shows such a bias when noise enters the fit like measurement error, which shortens the fitted decay time while the pattern of the mode is preserved (Section~\ref{sec:synthetic} and Supplementary Section~2). The same bias likely distorts the fitted frequencies, and mode~$1$ even has a real eigenvalue, although its ACF suggests a multidecadal timescale. Slow nonlinear dynamics, which the linear model cannot represent, may also contribute to this bias. Throughout this work, we therefore read decay times as lower bounds and summarize the behavior of each mode by a dominant timescale, defined as four times the first zero crossing of the ACF.\footnote{For a damped oscillation with ACF $e^{-t/\tau}\cos(2\pi t/T)$, the first zero crossing lies at $T/4$ regardless of the decay time $\tau$, so this timescale equals the period. In most of our experiments, the ACFs of the leading modes cross zero significantly, i.e., they reach negative values outside the white-noise confidence interval.} Since this crossing lies at a quarter of the period, the ACF of a short record estimates it much more reliably than the full period or a spectral peak at low frequencies. The spectra and the e-folding and decay times of the modes support this choice (Figs.~S21 and S27).}

\added{For mode~$1$, the timescale thus is about $80$~years (and about $60$~years for mode~$2$), comparable to the $65$--$70$-year oscillation in global-mean temperature identified by \citet{schlesinger1994oscillation}, which is strongest in the North Atlantic. Mode~$1$, however, has loadings in both the North Pacific and the North Atlantic, so the agreement concerns only the timescale. Because the $165$-year record spans only about two such timescales, this estimate is necessarily coarse. Moreover, \citet{mann2020absence} argue that the multidecadal spectral peak in observed temperatures primarily reflects combined natural and anthropogenic forcing rather than an internal oscillation. Under this interpretation, the internal component of mode~$1$ may absorb forced variance that the method fails to isolate, potentially due to limitations of the linear model.}
The second mode shows a complementary structure: a tripole pattern concentrated in the Pacific with comparatively weak Atlantic loadings, paired with an internal time series of similar multidecadal characteristics and autocorrelation shape to the first mode.
\added{The forced global warming response projects mainly onto these two modes. The forced component of their reconstruction accounts for $54\%$ of the variance of the full forced response, compared with $9\%$ and $19\%$ for modes~$3$ and~$4$. Among the dynamic modes of the \methodname~model inferred from data, these large-scale, slowest-decaying modes are thus the most responsive to the forcing.}

The complex-conjugate pair of modes with the third-longest decay time (modes $3$ and $4$ in Fig.~\ref{fig:dmd_modes_side_by_side_20crv3}) exhibits a two-dimensional ENSO/PDO-like spatial structure, with pronounced loadings over the tropical central and eastern Pacific. 
\added{The internal-component time series support this. Part~$1$ of the modes~$3,4$ pair has its maximum spectral power at ENSO frequencies (Fig.~S21), consistent with the DMD period of $1/0.243 \approx 4.1$~years for this pair, and the ACFs of both modes decay comparatively fast, first crossing zero near $1$~year, so the dominant timescale recovers this period. PDO variability, by contrast, is broadband, without a clear spectral peak, and is thought to arise partly as a reddened response to ENSO~\citep{ipcc2021annex4}, so we identify its signature in these modes mainly through their spatial structure. The second pattern of the pair (mode~$3,4$ part $2$) has large loadings over the western Pacific, Indian Ocean and North Atlantic, consistent with the remote warming of the South China Sea, Indian Ocean and tropical North Atlantic that lags ENSO through the atmospheric bridge~\citep{klein1999remote}.}
The dominant two-dimensional ENSO structure is well known and documented in observations~\citep{Tippett2020, Takahashi2011, Gavrilov2020}, and the complex-conjugate nature of this mode pair is consistent with the conceptual recharge oscillator ENSO model~\citep{Jin97a, Timmermann2018, Seleznev2023}.
The next modes (not shown) have less interpretable structure and contain more mixed time scales, ultimately resulting in noisy signals (cf. Fig. S22).

\paragraph{ESM data decomposition}

\begin{figure}[t!]
    \centering
    \includegraphics[width=\linewidth]{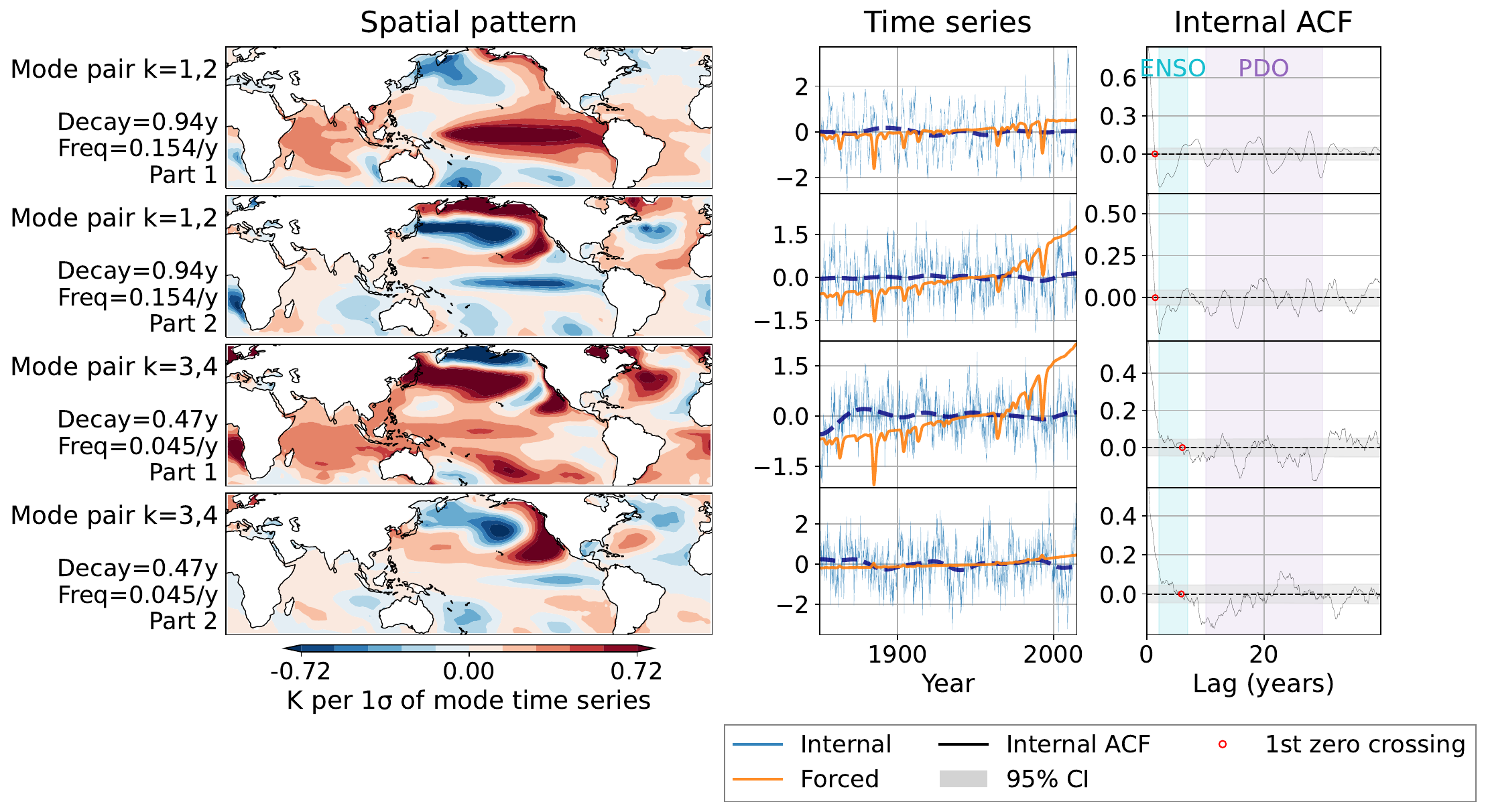}
    \caption{Same as in Fig.~\ref{fig:dmd_modes_side_by_side_20crv3} but computed for an example member of MPI-ESM \added{(r47i1p1f1)}.}
    \label{fig:dmd_modes_side_by_side_mpiesm}
\end{figure}

Applying the same decomposition to the ESMs exposes systematic structural differences from the reanalysis. 
\added{As an example, we show the decomposition for a randomly selected MPI-ESM member} (Fig.~\ref{fig:dmd_modes_side_by_side_mpiesm}). In doing so, and comparing additional members from other ESMs (Fig. S23--S25), we highlight several differences in OSAT variability between ESMs and reanalysis (20CRv3). 

First,~\methodname(3D) does not place multidecadal variability in the leading mode. Instead, the leading mode is complex and forms a complex conjugate pair with the $2$nd mode. This conjugate pair of modes corresponds to ENSO-like variability in \added{this} MPI-ESM ensemble member. The strong multidecadal timescale observed in the reanalysis does not appear in either the $3$rd or $4$th modes. We find neither an \added{AMV}-like spatial pattern nor pronounced multidecadal oscillations in the smoothed internal-variability time series (cf. Fig.~\ref{fig:dmd_modes_side_by_side_mpiesm}). \added{This agrees with the known difficulty of ESMs in generating AMV-like internal variability~\citep[defined in][]{ipcc2021annex4} of observed amplitude, which we examine across members and ESMs in \textit{Robustness across ESMs}.}

Second, the \added{ACF of the ENSO-like modes} looks more periodic than in reanalysis, suggesting more regular dynamics. \added{In addition,} modes $3$--$4$ exhibit mostly decadal 
\added{timescales} and are closer to the PDO \added{in spatial structure}, so the PDO appears more separated from ENSO in this realization than in observations. Even more regular ENSO-mode behavior is observed in CESM2 realizations, \added{visible as smooth, regular oscillations of the ACF within the ENSO band} (\added{see} Fig. S23).

Third, in MPI-ESM, the forced response is not isolated as a distinct slow mode but is instead embedded within the faster ENSO- and PDO-related dynamics. Consequently, the forced component is reconstructed as a response to these faster modes, in contrast to the reanalysis case, where independent slow multidecadal modes carry most of the forced response. \added{This difference is consistent with the biases in the tropical Pacific forced response discussed in Section~\ref{sec:forced_response}, since a forced warming concentrated in the eastern equatorial Pacific projects onto ENSO-like modes.}

\paragraph{Robustness across ESMs}

\begin{figure}[t!]
    \centering
    \includegraphics[width=\linewidth]{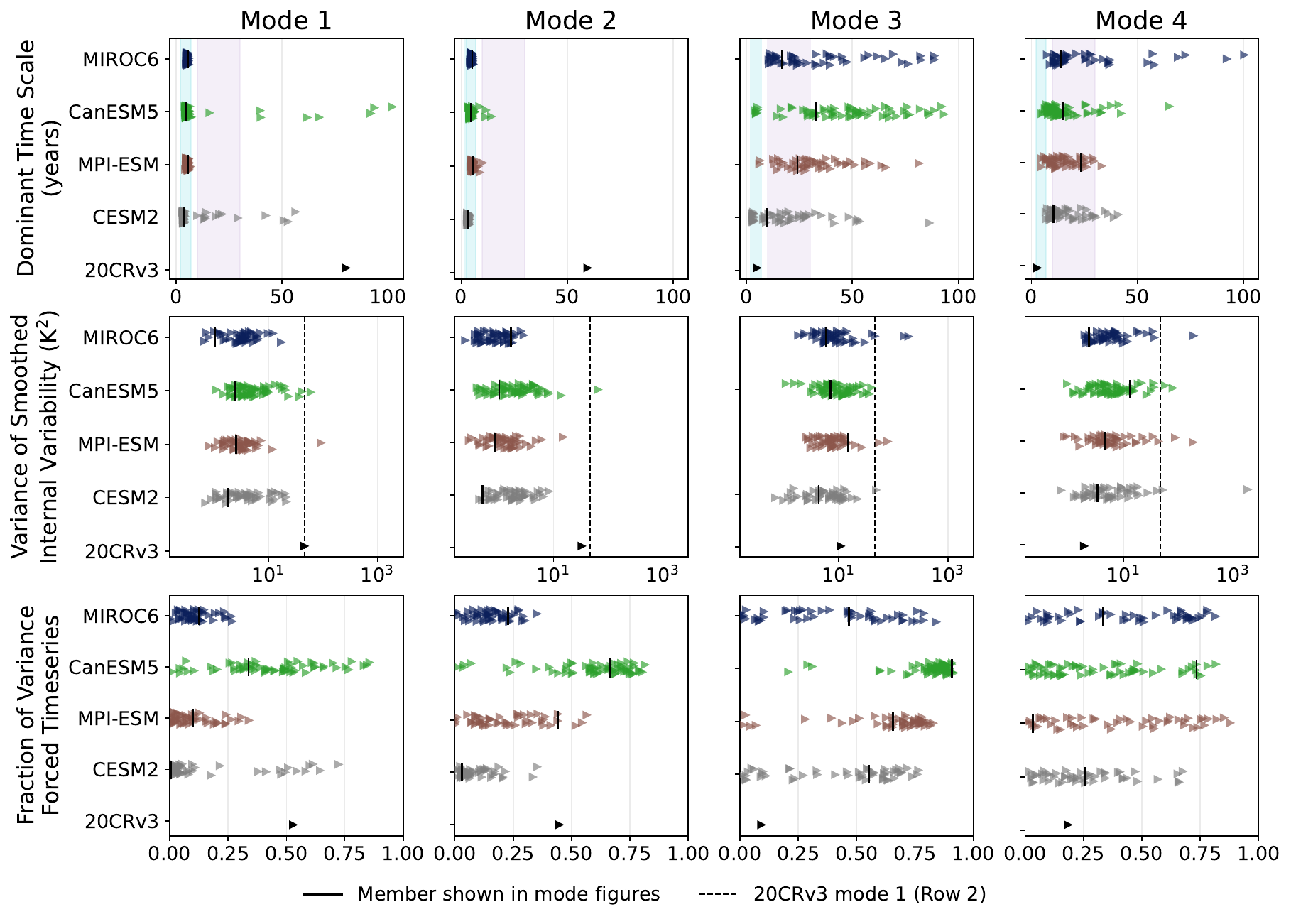}
    \caption{
    \added{Distributions of timescale, low-frequency variance, and forced fraction for the four leading modes to compare 20CRv3 with ESMs using the~\methodname(3D) model.} Row $1$ plots the dominant time scale of the internal-variability time series in years \added{as four times the lag of the first zero crossing of the ACF.
    The blue and purple bands mark the nominal ENSO and PDO ranges, as in Fig.~\ref{fig:dmd_modes_side_by_side_20crv3}. Row $2$ is the variance of the smoothed internal variability of each \added{rotated} mode. First, the internal time series are smoothed with a Gaussian filter ($\sigma = 8$~years, i.e. cutoff period near 40 years), which removes interannual variability such as ENSO.} 
    \added{Then we compute the variance of the smoothed reconstruction summed (area-weighted) over grid points.
    Values exceeding the horizontal axis range are not shown.
    The dashed vertical line in Row $2$ marks the variance of 20CRv3 mode~$1$.}
    Row $3$ is the fraction of variance of the forced time series \added{without smoothing}. \added{Black ticks mark the members shown in Figs.~\ref{fig:dmd_modes_side_by_side_mpiesm} and S23–S25.}}
    \label{fig:robustness_summary}
\end{figure}

To test whether the single-member contrasts above hold \added{more generally, we repeat the decomposition for every ensemble member of every ESM, focusing on the ACF-based} dominant timescale of each internal component time series across all ensemble members and ESMs (Row 1 of Fig.~\ref{fig:robustness_summary}). Note that this only represents the dominant time scale, meaning that 
\added{secondary or broadband timescales, such as decadal variability in modes $3$ and $4$ of 20CRv3, are missed.}
Modes $1$ and $2$ of all ESMs predominantly capture \added{variability on ENSO timescales} ($2$-$7$ years), with distributions tightly clustered around this range. 
\added{Since other modes also vary on these timescales, we identify modes $1$ and $2$ with ENSO from the spatial patterns of the single members of each ESM (Figs.~\ref{fig:dmd_modes_side_by_side_mpiesm}, S23--S25). Specifically, these patterns are concentrated in the equatorial central and eastern Pacific, closely matching the ENSO pair in 20CRv3 (modes $3$ and $4$), with no Indian Ocean dipole or equatorial Atlantic maximum.} Only a small number of outliers, primarily in CanESM5 and CESM2, exhibit longer timescales, likely reflecting a swap with the other modes in these members. Within the ENSO band, CESM2 shows slightly shorter periods ($4$--$5$ years) than other models ($5$--$7$ years).

Modes $3$ and $4$ of ESMs exhibit substantially broader distributions, including the \added{nominal} PDO \added{band}, spanning approximately $10$-$100$ years across models and ensemble members. While several members in each ESM reproduce multidecadal timescales comparable to the leading reanalysis modes, these occurrences are intermittent and co-exist with many realizations dominated by shorter variability. Specifically, reanalysis data have multidecadal timescales of $80$ and $60$ years in the first two modes. ESMs have dramatically fewer modes exceeding these timescale thresholds. This is clear in modes $1$, $2$, and $4$. Looking more closely at mode $3$, we find that the proportion of ensemble members with a mode $3$ timescale greater than the reanalysis multidecadal timescale of $60$ years ranges from $\sim2-28\%$ across ESMs, and only $\sim2-9.2\%$ for timescales greater than $80$ years. 
\added{This suggests that the ESMs mostly miss the observed multidecadal timescale, consistent with the widely reported underestimation of multidecadal variability in ESMs, mostly in magnitude. In the Atlantic, few CMIP5 models simulate substantial internally generated AMV~\citep{peings2016multidecadal}, CMIP models underestimate AMOC variability and its imprint on AMV~\citep{Yan_2018}, and in CMIP6 historical simulations much of the AMV is externally forced~\citep{robson2023contrasting}, so the IPCC AR6 has limited confidence in its model representation~\citep{Lee2021AR6Chapter4}. On the global scale, ESMs underestimate an observed multidecadal variability~\citep{Wyatt_2012,Kravtsov_2017,Kravtsov_2018,Seleznev2026} whose most distinct component in ESMs is AMV-like~\citep{Kravtsov2024}, strikingly similar to mode 1 of~\methodname(3D) in the reanalysis.}

\added{To compare multidecadal variability, we smooth the internal components with a $\sim$40 year Gaussian low-pass filter, which removes faster components like ENSO, and compute their variance (row~$2$ of Fig.~\ref{fig:robustness_summary}).}
We clearly extract the high variance of modes $1$ and $2$ \added{from the reanalysis}.
\added{Few ESM members reach comparable variance in any of their four leading modes.}
This indicates that the majority of the dominant timescales of the internal-variability time series that exceed the reanalysis timescale have weak amplitudes and/or low-variance spatial patterns, \added{consistent with the magnitude deficit noted above.}

We now examine the relative contribution of the forced response extracted by~\methodname(3D), quantified as the ratio of the variance of the forced component to that of the full time series, and assess whether the finding from the single MPI-ESM member generalizes across ESMs (Row $3$ of Fig.~\ref{fig:robustness_summary}). In reanalysis, the forced variance fraction is large in modes $1$ and $2$, consistent with the forced global warming response projecting predominantly onto the leading slow multidecadal modes. In contrast, the forced response in ESMs projects more strongly onto modes $3$ and $4$, which carry slower decadal-to-multidecadal time scales than the leading ENSO-like modes. \added{Across members, modes with slower internal variability tend to carry a larger forced fraction, as in the reanalysis, although ENSO-like modes in the ESMs can still carry a non-negligible forced response (Fig.~S26).} Among modes $1$ and $2$, CanESM5 and MPI-ESM show comparatively larger forced contributions, while CESM2 (excluding ensemble members with apparent mode swaps) and MIROC6 show smaller ones. Overall, the forced variance fraction is more distributed across all four modes, reflecting the absence of a dominant slow multidecadal mode that would otherwise capture most of the forced signal. Across all modes, the forced variance fractions are largest for CanESM5. \added{CanESM5 has the highest equilibrium climate sensitivity ($5.6$ K) and one of the highest transient climate responses among the CMIP6 models, compared with $2.6-5.2$ K for the other ESMs in our study~\citep[Table 2]{meehl2020context}. Thus, CanESM5's high forced variance fractions are} consistent with its strong sensitivity to external forcing relative to the other tested ESMs, also documented in Sec.~\ref{sec:forced_response}.

\added{Together, these diagnostics show that ESMs and reanalysis disagree on which variability is most persistent. In the reanalysis, slow multidecadal modes lead and carry most of the forced response. In the ESMs, ENSO/PDO-scale modes lead in almost all members, and the forced response spreads over more modes. The eigenvalues of $\A$ show the same reversal (Fig.~S28). Multidecadal internal variability occurs in some ESM members, but with low variance, although \citet{mann2020absence} argue that the observed multidecadal peak is largely forced. The reversal may also partly reflect estimating $\A$ from a short, noisy record, which can favor faster oscillatory modes when slower ones are weak (Section~\ref{sec:conclusion}).}

\section{Discussion and Conclusion}\label{sec:conclusion}

In this work, we proposed \methodname, which combines LIM-based dynamical structure with explicit external forcing as a non-stationary linear stochastic dynamical system. Our method uses pullback attractor estimation to formally separate forced-response and internal-variability contributions and to decompose a \textit{single} spatiotemporal climate signal into a collection of spatial patterns, each accompanied by forced and internal time series.

To assess the method's performance, we evaluated forced-response estimation skill against several baselines using ground truth from large ESM ensembles. \methodname~consistently matches or exceeds LR skill across temperature and pressure variables, while outperforming methods that do not incorporate forcing information. A key practical finding is that performance depends on the choice of forcing predictors: richer forcing representations improve temperature skill but can degrade pressure skill, and ESM ground truth provides a natural framework for identifying optimal predictor combinations.

The spatiotemporal decomposition reveals a physically interpretable picture of how ESMs differ from reanalysis. In reanalysis, the leading modes capture slow multidecadal variability \added{of both the Atlantic and the Pacific}, and the forced response projects predominantly onto these slow structures. ESMs systematically prioritize faster ENSO-scale dynamics in the leading modes, underrepresent multidecadal variability, and distribute the forced response across a broader set of modes. This structural difference likely reflects the well-reported underestimation of multidecadal variability amplitude in current ESMs \citep{Wyatt_2012,Kravtsov_2018,Kravtsov2024,Seleznev2026}. 
\added{The ESMs also only partly reproduce the change in the seasonal cycle seen in the reanalysis.}

The ensemble-mean forced response as a ground truth estimator is known to be inaccurate for small ensembles or under large internal variability~\citep{Kravtsov2022}; more advanced estimators such as signal-to-noise maximizing patterns, extended EOFs, or ensemble-based linear dynamical modes~\citep{Wills2020,Kravtsov2022,Gavrilov2024,Buyanova2025} could provide more reliable benchmarks and would be worth exploring in future evaluations. Here, we mitigated this issue by using large ensembles of $40+$ members and applying a simple smoothed ensemble mean to SLP, which substantially reduces ground-truth uncertainty~\citep{Gavrilov2024}. \added{The forcing inputs are also uncertain, especially for aerosols~\citep{IPCC_2021_WGI_Ch_7}, so model-based validation may overstate skill on observations. However, we use the same AR6 ERF for all ESMs rather than each model's own forcing, so our skill estimates already include some forcing mismatch.} Similarly, our baseline comparison is intentionally focused on a small set of simple methods and four ESMs; a broader comparison with state-of-the-art methods across more models would be valuable. \added{Within ForceSMIP, \methodname~belongs to a small group of methods that take the forcing history as input, alongside DMDc, GPCA, GPCA-DA, and MLR-Forcing~\citep{Wills2026}. Unlike DMDc, which combines the least-damped mode of $\boldsymbol{A}$ with the linear forcing term, \methodname~defines the forced response as the fitted system's pullback attractor. Since it requires no training on large ensembles and has few parameters, we hypothesize that its skill transfers to unseen models and observations.} To this end, \methodname~has been submitted to Tiers $2$ and $3$ of ForceSMIP~\citep{Wills2026}, where results will be compared systematically with the broader community of methods when released.

\added{As a decomposition method, \methodname~shares conceptual ground with approaches that target time scale separation. Average predictability time decomposition~\citep{DelSole2009,DelSole2009a} identifies patterns that maximize predictability time, and low-frequency component analysis~\citep{Wills2018} emphasize slower time scales. The LDM method~\citep{Gavrilov2016,Gavrilov2019} trains latent time series of the modes, enabling more effective noise filtering. A similar idea is used in optimized DMD~\citep{Askham2018,Sashidhar2022}, which suggests a natural extension of~\methodname~to improve the projected forced-response time series.}

Finally, we highlight three methodological directions we consider most promising for overcoming inherent limitations of the \methodname~framework. First, the framework is linear, meaning that the pullback attractor is represented as a single trajectory and the decomposition into forced and internal components is exact only under linear dynamics. In nonlinear systems, this structure would generally become set-valued, and a clean forced–internal separation is not guaranteed, requiring new formal definitions of forced response beyond the current framework. 
\added{The skill against the ensemble-mean forced response of four nonlinear ESMs (Section~\ref{sec:forced_response}) nevertheless indicates that the assumptions of linear dynamics and a linear response are adequate for estimating the forced response of the large-scale variables considered here, consistent with the hypothesis in Section~\ref{sec:synthetic} that \methodname~approximates the mean of the pullback attractor.}
The nonlinear parameterizations, such as polynomials~\citep{Chekroun2017,KKG05,kkg09_rev}, low-dimensional neural networks~\citep{Mukhin2015a, Seleznev2019} and Koopman operators in latent state space \citep{Nathaniel2025,turri2026selfsupervised,Lucarini2026}, given the forcing predictors at the input, fall naturally into the same non-autonomous dynamical system framework, representing the most straightforward generalization.
Neural operators~\citep{kovachki2023neural} or manifold learning approaches~\citep{constante2024data} could further capture state-dependent variability and regime behavior that lie outside the linear framework. \added{Many of these amount to a change of observables, since the model can act on delay coordinates~\citep{brunton2017chaos,Takens} or on nonlinear lifts of the state such as kernel features~\citep{williams2014kernel} instead of leading PCs. The sensitivity of the forced-response estimate to this choice remains unexplored, and it likely depends on which observables are used and how, both for modeling the dynamics and for estimating the response in the physical variables.}

Second, estimating the dynamical operator $\boldsymbol{A}$ from finite, noisy data is a primary source of uncertainty. In practice, decay times should be interpreted as lower bounds on physically relevant timescales under the current estimator. Noise and finite-sample-length bias estimates toward faster decorrelation and may obscure weak but persistent multidecadal structure. 
\added{The synthetic example confirms this bias and shows that it grows as the decay time approaches the record length (Section~\ref{sec:synthetic}), consistent with the larger errors for the shorter record from 1950 to 2014 (Section~\ref{sec:forced_response}).}
This effect is particularly relevant given that a $\approx100$-year forcing history may be insufficient for full convergence of the pullback dynamics when intrinsic timescales are longer than those recovered here. \added{On the other hand, for nearly white-noise variables, $\boldsymbol{A}$ approaches zero and the pullback attractor reduces to a regression on the forcing, which may explain why \methodname~matches rather than exceeds LR for some variables.} Thus, improving estimation of $\boldsymbol{A}$ is a priority for future work. Noise-aware filtering strategies, latent dynamical system approaches mentioned above, regularization, and physically informed constraints on the operator spectrum may shift estimated decay times toward longer values and improve recovery of weak multidecadal signals that are currently suppressed by sampling noise.

Third, methodological choices such as time lag, truncation level, \added{choice of observables,} and forcing specification \added{may} influence results. \added{The forced-response skill degrades with fewer PCs and changes little with more PCs, longer lags or a longer forcing history (Supplementary Section~1g). Observables were not tested, and forcing sets were only tested through the three variants in Section~\ref{sec:forced_response}. These choices may} contribute to mode mixing, particularly in separating oscillatory and non-oscillatory structures. Furthermore, such parameters jointly determine the effective spectral content of the learned operator and may explain part of the observed reversal of mode ordering between reanalysis and ESMs. In particular, 
lag-dependent spectral mixing combined with noise and finite-sample effects may bias leading modes toward faster oscillatory variability in ESMs, even when slower modes are present but weak. 
\added{For example, our $3$-month lag retains the ENSO-like modes analyzed in Section~\ref{sec:results_decomposition}, whereas longer lags tend to treat ENSO as a stochastic residual~\citep{Penland1993}.}
Therefore, a systematic exploration of \added{these methodological choices} is warranted to reduce estimation bias and improve the recovery of long-timescale variability. 
\added{The forcing specification also enters through the fixed pattern of each input, its column of $\boldsymbol{B}$. This suits well-mixed greenhouse gases but not aerosols, whose sources have shifted from Europe and North America toward Asia~\citep{persad2018divergent}, so in \methodname~the forced pattern can change only through the dynamics in $\A$, not through where the forcing acts.} 
\added{A natural refinement is spatially resolved forcing, as in ClimateBench~\citep{watson2022climatebench}, whose emulators use spatial maps of aerosol emissions and thus follow the changing geography of aerosol sources. Such inputs could enter $\boldsymbol{B}$ to capture time-varying aerosol response patterns, and could help decide whether the recent tropical Pacific cooling in 20CRv3, which global forcing time series cannot attribute (Section~\ref{sec:forced_response}), is forced or internal.}

Implementing any of these extensions will require careful attention to interpretability, which remains central to the climatological utility of the~\methodname~framework. More broadly,~\methodname~illustrates how classical non-autonomous dynamical-systems theory and modern data-driven decomposition can be brought together to read a single climate record as the fading imprint of its own forcing history. We expect our approach to extend naturally to other variables, to coupled multivariate fields, and, eventually, to the nonlinear regimes.

\acknowledgments
This work was supported by the European Union's Horizon Europe research and innovation programme under projects ``European Lighthouse of AI for Sustainability'' (ELIAS; 101120237) and ``Artificial Intelligence and Machine Learning for Enhanced Representation of Processes and Extremes in Earth System Models'' (AI4PEX; 101137682).

\datastatement
The main codebase for this project can be found here~\url{https://github.com/IPL-UV/PullbackDMDc}.

Our reanalysis data from observations can be accessed from~\url{https://www.psl.noaa.gov/thredds/catalog/Datasets/20thC_ReanV3/Monthlies/2mSI-MO/catalog.html} on \url{https://www.psl.noaa.gov/data/gridded/data.20thC_ReanV3.html}.

The ESM data are available at \url{https://gdex.ucar.edu/datasets/d651039/}.

The radiative forcing information is from the IPCC AR6 Chapter 7 (\citep{IPCC_2021_WGI_Ch_7}; accessible \url{https://raw.githubusercontent.com/IPCC-WG1/Chapter-7/main/data\_output/AR6\_ERF\_1750-2019.csv}) and interpolated to a monthly resolution.

The code for running the baseline methods: RegGMST, LIM, and LIM-opt was adapted from~\url{https://github.com/karenamckinnon/forcesmip} associated with the ForceSMIP project.

\bibliographystyle{ametsocV6}
\bibliography{references}

\clearpage
\newpage

\appendix[S] 
\appendixtitle{Supplementary Methodological Details}
Here $M$ denotes the dimension of the state vector $\x(t)$. When spatial dimensionality reduction is applied (as in Section 2c), $\x(t)$ represents the vector of the leading $M$ principal components (PCs).

\subsection{Linear inverse models (LIM)}

In LIM, we assume \added{that $\x(t)$ follows} the low-dimensional stochastic dynamical system
\begin{equation*}
    \frac{d  \x(t)}{dt} = \bL \x(t) + \boldsymbol{\eta}(t),
\end{equation*}
\added{where $\bL$ is a constant matrix whose eigenvalues have negative real parts, so that the deterministic dynamics are stable, and $\boldsymbol{\eta}(t)$ is Gaussian white noise with zero mean and constant covariance.} The solution to this system for the discrete time lag $\tau=1$ is:
\begin{equation*}
\x(t)
=
e^{\boldsymbol{L} } \x(t-1)
+
\int_{t-1}^{t}
e^{\boldsymbol{L} (t - s)} \boldsymbol{\eta}(s)\, ds.
\end{equation*}
Using $\boldsymbol{A}=e^{\boldsymbol{L} }$ and $\boldsymbol{\xi}(t)=\int_{t-1}^{t}
e^{\boldsymbol{L} (t - s)} \boldsymbol{\eta}(s)\, ds$, we write discrete LIM \textit{with a random component} as
\begin{equation*}
    \x(t) = \boldsymbol{A} \x(t-1) + \boldsymbol{\xi}(t).
\end{equation*}
In LIM, we estimate $\boldsymbol{A}$ via the least squares problem
\begin{equation}\label{eq: lim_discrete_opt}
    \min_{\boldsymbol{A}} \sum_t\| \x(t) - \boldsymbol{A}\x(t-1)\|_2^2.
\end{equation}

\added{Following~\cite{Frankignoul17}, as implemented in the ForceSMIP example notebook,} we use the eigendecomposition of $\boldsymbol{A}$ to \emph{estimate the forced response with LIM}. Specifically, we use the least damped mode, $\boldsymbol{w}_{1}$, the eigenvector of $\boldsymbol{A}$ associated with the eigenvalue with the largest absolute value. This mode is assumed to be real-valued\added{, although in practice it can be complex (see below)}. It is the pattern that decays slowest in time. We denote the time series associated with this mode as $z_1(t)$, namely the first entry of $\boldsymbol{z}(t)$ that solves $\boldsymbol{W}\boldsymbol{z}(t) = \x(t)$. Using these quantities, we construct the forced response as
\begin{equation}\label{eq: lim_forced_response}
    \x^{(f)}(t) \approx \boldsymbol{w}_1 z_{1}(t).
\end{equation}
\added{When $\boldsymbol{w}_1$ is complex, we take the real part of this contribution, $\operatorname{Re}\big( \boldsymbol{w}_1 z_1(t)\big)$, and discard its complex-conjugate partner. This is half of the joint contribution of the conjugate pair, $2\operatorname{Re}\big( \boldsymbol{w}_1 z_1(t)\big)$.}

\subsection{LIM with Optimal Perturbation Patterns (LIMopt)} \label{app: limopt}
In LIMopt, 
$\boldsymbol{A}$ is estimated using the least-squares solution to~\eqref{eq: lim_discrete_opt}~\citep{solomon2012reconciling,Frankignoul17,Wills2020}.

The smoothing of LIMopt comes in the \emph{optimal perturbation filter}. Specifically, we compute the \added{propagator of $\x(t)$ over $\tau_e$ time steps, $\boldsymbol{A}^{\tau_e}$}.

LIMopt uses the patterns in the left and right singular vectors \added{(denoted $\boldsymbol{\psi}$ and $\boldsymbol{\phi}$ respectively)} associated with the largest singular value of $\boldsymbol{A}^{\tau_e}$ to estimate the forced response as
\begin{equation}\label{eq: limopt_forced_response}
    \x^{(f)}(t) \approx   \boldsymbol{\psi} \boldsymbol{\phi}^\top \x(t).
\end{equation}
\added{We use $\tau_e = 20$ steps at the $3$-month lag for the climate data and $\tau_e = 60$ steps at the $1$-step lag for the synthetic example, i.e., an optimal perturbation time of $60$ months in both cases.}

\subsection{Dynamic Mode Decomposition with Control}
Dynamic Mode Decomposition with Control (DMDc)~\citep{proctor2016dynamic} assumes the discrete-time linear dynamical system with control
\begin{equation}\label{eq: discrete time forced}
    \x(t) = \boldsymbol{A} \x(t-1) + \boldsymbol{B} \y(t-1).
\end{equation}

In our work, we replace $\y(t-1)$ with $\y(t)$ to align with the convention commonly adopted in the climate literature. Mathematically, this change reflects only a difference in indexing and does not alter the underlying dynamics. More importantly, if we go from a continuous formulation, the forcing term at time $t$ is dynamically integrated over the interval between the previous observed state (at $t-1$) and the moment $t$. Given \added{the forcing varies slowly relative to $\tau$}, $\y(t)$ is close to \added{$\y(t-\tau)$}, one can show that the first-order approximation of the integral would be a linear function of $\y(t)$. Therefore, in the main text, we treat $\y(t)$ as the forcing \textit{actually observed} at the month $t$.

\added{DMDc was originally formulated on the raw data. We instead fit a full-rank model to the truncated state of Section~2c of the main text, for computational efficiency and a simpler comparison with LIM. Finding $\boldsymbol{A}$ and $\boldsymbol{B}$ then reduces to} a simple least-squares linear regression problem
\begin{align*}
    &\min_{\boldsymbol{A},\boldsymbol{B}} \sum_t \|\x(t) - \boldsymbol{A} \x(t-1) - \boldsymbol{B} \y(t) \|_2^2.
\end{align*}

{\add
\subsection{Data matrix construction and lag notation}
For simplicity of exposition in the main text, we presented~\methodname, LIM and LIMopt assuming the sampling step of the data coincides with the lag step of the trained model, i.e., a model of the form
\begin{equation}\label{eq: supp_linear_system}
    \x(t) = \A\x(t-1) + \boldsymbol{B}\y(t) + \boldsymbol{\xi}(t),
\end{equation}
(with $\boldsymbol{B}=0$ for LIM and LIMopt) where ``$t-1$'' denotes one native sampling step (in our case, months). In practice, however, the model is trained at a chosen lag $\tau$ (3 months for climate experiments and 1 for synthetic data), which need not equal the native sampling interval of the underlying anomaly fields. The model fit is therefore
\begin{equation}\label{eq: supp_linear_system_tau}
    \x(t) = \A\x(t-\tau) + \boldsymbol{B}\y(t) + \boldsymbol{\xi}(t),
\end{equation}
with $\A$ a $\tau$-step propagator rather than a $1$-native-step propagator.

To fit Eq.~\eqref{eq: supp_linear_system_tau}, we retain all snapshots at native (monthly) resolution, but construct the paired snapshot matrices by pairing each snapshot with the one $\tau$ steps ahead, rather than one step ahead:
\begin{align*}
    &\boldsymbol{X} = \big[\x(s+1), \x(s+2), \dots, \x(T-\tau)\big],
    \qquad
    \boldsymbol{X}' = \big[\x(s+1+\tau), \x(s+2+\tau), \dots, \x(T)\big],\\
    &\boldsymbol{Y} = \big[\y(s+1+\tau), \y(s+2+\tau), \dots, \y(T)\big].
\end{align*}
That is, column $j$ of $\boldsymbol{X}'$ is the snapshot occurring $\tau$ steps after column $j$ of $\boldsymbol{X}$, while consecutive columns within each matrix remain separated by a single native step, not by $\tau$. This design choice makes full use of all available native-resolution snapshots rather than subsampling every $\tau$-th one. We solve the DMDc regression $\boldsymbol{X}' \approx \A\boldsymbol{X} + \boldsymbol{B}\boldsymbol{Y}$ for the propagator $\A$ associated with lag $\tau$.

The pullback-attractor solution of Eq.~\eqref{eq: supp_linear_system_tau} retains the same functional form as in the unit-lag case, but with $\tau$ appearing explicitly in the spacing of past terms under both sums:
\begin{align}
\begin{aligned}\label{eq:decompose_ds_tau}
\x(t| \x_0,t_0) \underset{t_0 \to -\infty}{\xrightarrow{\hspace{1.2cm}}} &\underbrace{\sum\limits_{j=0}^{\infty}  \A^{j}\boldsymbol{B} \y(t-j\tau)}_{\text{forced response}} + \underbrace{\sum\limits_{j=0}^{\infty}  \A^{j}\boldsymbol{\xi}(t-j\tau)}_{\text{internal variability}}\\
&= \hspace{.4cm}\x^{(f)}(t) \hspace{.6cm}+\hspace{.55cm}\x^{(i)}(t).
\end{aligned}
\end{align}
In this case, each term in the forced response and internal variability sums is spaced $\tau$ steps apart in the past, rather than one native step. 
In practice, this means the algorithm computes the forced response as a deterministic (noise-free) prediction \added{in $\tau$ interleaved chains, each starting from a zero state at one of the first $\tau$ time steps and propagated forward recursively} via Eq.~\eqref{eq: supp_linear_system_tau}, then stitched back together. Since each chain starts from a zero state, its difference from $\x^{(f)}(t)$ is the initial error, which is multiplied by $\A$ at every step and therefore decays with the decay times of $\A$. Discarding the first $s$ steps, the transition time, with $s$ much longer than these decay times, makes the remaining error negligible. For example, for $\tau=3$, we have three matrices of forced-response predictions 
\begin{align}
    \begin{aligned}
        \X_1^{(f)} &= \begin{bmatrix} \x^{(f)}(1) & \x^{(f)}(1+3) & \cdots & \x^{(f)}(T-2)\end{bmatrix},\\ 
        \X_2^{(f)} &= \begin{bmatrix} \x^{(f)}(2) & \x^{(f)}(2+3) & \cdots & \x^{(f)}(T-1)\end{bmatrix},\\ 
        \X^{(f)}_3 &= \begin{bmatrix} \x^{(f)}(3) & \x^{(f)}(3+3) & \cdots & \x^{(f)}(T)\end{bmatrix}.
    \end{aligned}
\end{align} 
When stitched together, one column at a time, they contain the entire forced response time series. This makes $\tau$ an explicit hyperparameter of \methodname, rather than a quantity fixed by the notation in Eq.~\eqref{eq: supp_linear_system}.}

\subsection{Mode-wise decomposition for~\methodname}
\added{We assume that $\A$ is diagonalizable. In general, it need not be. One can then approximate $\boldsymbol{W}^{-1}$ with numerical solvers for linear systems or reduce the dimension $M$ of $\x(t)$ to ensure diagonalizability. Using the eigendecomposition $\A = \boldsymbol{W}\boldsymbol{\Lambda}\boldsymbol{W}^{-1}$, we project onto the dynamic modes via $\z(t) = \boldsymbol{W}^{-1}\x(t)$, with $\z^{(f)}(t) = \boldsymbol{W}^{-1}\x^{(f)}(t)$ and $\z^{(i)}(t) = \boldsymbol{W}^{-1}\x^{(i)}(t)$.  For simplicity, we write the unit-lag case; for lag $\tau$, $t-1$ and $t-j$ become $t-\tau$ and $t-j\tau$, as in Eq.~\eqref{eq:decompose_ds_tau}. Applying this change of variables to Eq.~\eqref{eq: supp_linear_system} gives}
\begin{equation}\label{eq:z-model}
\z(t)=\boldsymbol{\Lambda}\z(t-1)+\hat{\y}(t)+\hat{\boldsymbol{\xi}}(t),
\end{equation}
with $\hat{\y}(t)=\boldsymbol{W}^{-1} \boldsymbol{B} \y(t)$ and $\hat{\boldsymbol{\xi}}(t)=\boldsymbol{W}^{-1}\boldsymbol{\xi}(t)$. The terms $\z^{(f)}(t)$ and $\z^{(i)}(t)$ represent the pullback attractor projected to the space of dynamic modes, and also correspond to the pullback attractor of the model~\eqref{eq:z-model}. Similarly to classic deterministic DMD modes, the evolution of each component of the pullback attractor is determined by its eigenvalue $\lambda_k$ and the forcing/noise signals:
\begin{equation}\label{eq:zk-decompose-ds}
z_k(t) =  \sum\limits_{j=0}^{\infty}  \lambda_k^{j}\hat{y}_k(t-j) +  \sum\limits_{j=0}^{\infty}  \lambda_k^{j}\hat{\xi}_k(t-j)\;= \; z^{(f)}_k(t) +z^{(i)}_k(t).
\end{equation}
\added{In DMDc, each mode evolves from its initial amplitude under the deterministic control input. Here, the initial amplitude is forgotten, and the noise history enters as a second driver, so every mode carries both a forced and an internal time series, each filtered by $\lambda_k$.}

\subsection{Rotation of complex-conjugate mode pairs}\label{sec:pca_rotation}

\added{A conjugate pair $(\boldsymbol{w}_\ell, \overline{\boldsymbol{w}_\ell})$ spans a real two-dimensional invariant subspace of $\A$, spanned by $2\mathrm{Re}(\boldsymbol{w}_\ell)$ and $-2\mathrm{Im}(\boldsymbol{w}_\ell)$, within which the dynamics oscillate with the pair's shared decay time and frequency. Splitting this subspace into two real patterns is not unique: any real basis of the subspace is equally valid, and $\boldsymbol{w}_\ell$ itself is defined only up to multiplication by a complex scalar, which changes its real and imaginary parts. Both reduce the identifiability of the modes and limit their physical interpretation. We remove this ambiguity with a 2D PCA rotation within each pair, computed as follows.}

We first verify that the projected internal time series $z_\ell^{(i)}(t)$ form complex-conjugate pairs. Since $\x^{(i)}(t)$ is real and $\boldsymbol{z}^{(i)}(t) = \boldsymbol{W}^{-1}\x^{(i)}(t)$, and the rows of $\boldsymbol{W}^{-1}$ corresponding to a conjugate eigenvector pair are themselves conjugate, we have $z_{\ell+1}^{(i)}(t) = \overline{z_\ell^{(i)}(t)}$, and analogously $z_{\ell+1}^{(f)}(t) = \overline{z_\ell^{(f)}(t)}$.

The internal contribution of the conjugate pair $(\boldsymbol{w}_\ell, \overline{\boldsymbol{w}_\ell} = \boldsymbol{w}_{\ell+1})$ to $\x^{(i)}(t)$ is the sum of two conjugate terms, which is purely real:
\begin{align}
    \begin{aligned}
    \boldsymbol{w}_\ell z_\ell^{(i)}(t) + \overline{\boldsymbol{w}}_\ell 
    \overline{z_\ell^{(i)}(t)} 
    = &2\mathrm{Re}\left(\boldsymbol{w}_\ell z_\ell^{(i)}(t)\right) \\
    = &2\mathrm{Re}(\boldsymbol{w}_\ell)\mathrm{Re}\left(z_\ell^{(i)}(t)\right) \\
    &- 2\mathrm{Im}(\boldsymbol{w}_\ell)\mathrm{Im}\left(z_\ell^{(i)}(t)\right).
   \end{aligned}
\end{align}
This can be written as a product of two $2$D real matrices by stacking the real and imaginary parts of the mode and its internal time series:
\begin{align}
    \widetilde{\boldsymbol{W}}_\ell &= [2\mathrm{Re}(\boldsymbol{w}_\ell),\, 
    -2\mathrm{Im}(\boldsymbol{w}_\ell)]\in \mathbb{R}^{M\times 2},\\
    \widetilde{\boldsymbol{Z}}^{(i)}_\ell &= \left[\mathrm{Re}\left(\boldsymbol{z}_\ell^{(i)}\right),\, 
    \mathrm{Im}\left(\boldsymbol{z}_\ell^{(i)}\right)\right] \in \mathbb{R}^{T\times 2},
\end{align}
so that the contribution equals $\widetilde{\boldsymbol{W}}_\ell \left(\widetilde{\boldsymbol{Z}}^{(i)}_\ell\right)^\top$.

The $2$D PCA rotation is computed via QR decompositions\footnote{\added{The QR decomposition factors a matrix $\boldsymbol{X}$ into a matrix with orthonormal columns $\boldsymbol{Q}$ and an upper triangular matrix $\boldsymbol{R}$, so that $\boldsymbol{X}=\boldsymbol{Q}\boldsymbol{R}$.}}  $\widetilde{\boldsymbol{W}}_\ell = \boldsymbol{Q}_W \boldsymbol{R}_W$ and $\widetilde{\boldsymbol{Z}}^{(i)}_\ell = \boldsymbol{Q}_Z \boldsymbol{R}_Z$, followed by the SVD $\boldsymbol{R}_W \boldsymbol{R}_Z^\top = \boldsymbol{G} \boldsymbol{\Sigma} \boldsymbol{H}^\top$. \added{This avoids forming the $M \times T$ matrix $\widetilde{\boldsymbol{W}}_\ell(\widetilde{\boldsymbol{Z}}^{(i)}_\ell)^\top$ and reduces the computation to the SVD of a $2\times 2$ matrix.}
Note that $\boldsymbol{R}_W \in \mathbb{R}^{2\times 2}$ is invertible because $\widetilde{\boldsymbol{W}}_\ell$ has full column rank: its two columns $2\mathrm{Re}(\boldsymbol{w}_\ell)$ and $-2\mathrm{Im}(\boldsymbol{w}_\ell)$ are linearly independent whenever $\lambda_\ell$ is not purely real, which holds for any genuinely oscillatory complex-conjugate pair. 
The left singular vectors of  $\widetilde{\boldsymbol{W}}_\ell(\widetilde{\boldsymbol{Z}}^{(i)}_\ell)^\top$ are exactly $\boldsymbol{Q}_W \boldsymbol{G}$, which \added{has orthonormal columns,} since \added{$\boldsymbol{Q}_W$ has orthonormal columns and $\boldsymbol{G}$ is orthogonal}. 
This defines a $2\times 2$ invertible linear map $\boldsymbol{P} = \boldsymbol{R}_W^{-1}\boldsymbol{G} \in \mathbb{R}^{2\times 2}$ acting on $\widetilde{\boldsymbol{W}}_\ell$ from the right, yielding the rotated patterns and time series:
\begin{align}
    [\widehat{\boldsymbol{w}}_\ell,\, \widehat{\boldsymbol{w}}_{\ell+1}] &= 
    \boldsymbol{Q}_W \boldsymbol{G} = \widetilde{\boldsymbol{W}}_\ell \boldsymbol{P}
    \in \mathbb{R}^{M\times 2},\\
    [\widehat{z}^{(i)}_\ell,\, \widehat{z}^{(i)}_{\ell+1}] &= 
    \widetilde{\boldsymbol{Z}}^{(i)}_\ell \boldsymbol{P}^{-\top}
    \in \mathbb{R}^{T\times 2},
\end{align}
where $\boldsymbol{P}^{-\top} = \boldsymbol{R}_W^\top \boldsymbol{G}$ follows from $\boldsymbol{P}^{-1} = \boldsymbol{G}^\top \boldsymbol{R}_W$.
Of course, this rotation preserves the reconstruction of the internal component:
\begin{align}
    (\widetilde{\boldsymbol{W}}_\ell \boldsymbol{P})
    (\widetilde{\boldsymbol{Z}}^{(i)}_\ell \boldsymbol{P}^{-\top})^\top
    &= \widetilde{\boldsymbol{W}}_\ell \boldsymbol{P} \boldsymbol{P}^{-1}
    (\widetilde{\boldsymbol{Z}}^{(i)}_\ell)^\top
    = \widetilde{\boldsymbol{W}}_\ell(\widetilde{\boldsymbol{Z}}^{(i)}_\ell)^\top.
\end{align}

\added{The rotated internal time series are $\widetilde{\boldsymbol{Z}}^{(i)}_\ell \boldsymbol{P}^{-\top} = \boldsymbol{Q}_Z \boldsymbol{R}_Z \boldsymbol{R}_W^\top \boldsymbol{G} = \boldsymbol{Q}_Z \boldsymbol{H}\boldsymbol{\Sigma}$, so they are orthogonal, with norms given by the singular values $\sigma_1 \ge \sigma_2$; this orders the two components by their contribution to internal variability. The rotation is unique up to the sign of each component, provided $\sigma_1 > \sigma_2$.}

\added{We apply the same map to the forced time series. With $\widetilde{\boldsymbol{Z}}^{(f)}_\ell = [\mathrm{Re}(\boldsymbol{z}^{(f)}_\ell),\, \mathrm{Im}(\boldsymbol{z}^{(f)}_\ell)]$, the rotated forced time series are $[\widehat{z}^{(f)}_\ell,\, \widehat{z}^{(f)}_{\ell+1}] = \widetilde{\boldsymbol{Z}}^{(f)}_\ell \boldsymbol{P}^{-\top}$, which reproduce the pair's forced contribution by the same argument as above. Because $\boldsymbol{P}$ is determined by the internal component, the rotated forced time series are not orthogonal in general.}

\subsection{Parameters and Data Preprocessing}
\added{Preprocessing and the truncation level of PCs ($20$ for SAT and OSAT, $200$ for SLP) are described in Section~2c of the main text. For SLP, the forced signal is much weaker than the fast internal variability and noise, so a small number of PCs would not capture it.}

\paragraph{Ground truth ensemble mean}
The same factors that lead to selecting a larger \added{number of PCs} for SLP introduce additional error into the raw ensemble-mean forced-response estimates. Therefore, we smooth the ensemble-mean SLP anomalies using a $3$-year sliding window to improve the computed ground-truth forced-response signal for comparison.

\paragraph{Forcing} 
\added{The forcing inputs and \methodname\ variants are described in the main text. The default total effective radiative forcing matches the input used in several ForceSMIP submissions~\citep{Wills2026}.} Regardless of dimension, the forcing data are mean-centered and scaled by their standard deviation over the target interval,\footnote{This interval is $1850$--$2014$ for the results in the main manuscript, and a shorter interval of $1950$--$2014$ are available in this Supplementary.} and the past forcing used in \methodname~variants is scaled with the same coefficients.

\paragraph{Time lags}
\added{For methods with a dynamical component (\methodname~variants, LIM, LIMopt), we use a lag of $3$ months throughout, following ForceSMIP submissions~\citep{Wills2026}. Since longer lags treat faster modes such as ENSO as stochastic residuals~\citep{Penland1993}, this lag retains ENSO in the dynamics while still capturing slower variability and the forced response. In general, the lag should be chosen according to the phenomena to be retained in the dynamics or filtered out as noise.}

\added{We use a $100$-year forcing history before the analysis period for all \methodname\ variants. This window needs to exceed the decay times of the fitted propagator $\A$, not the equilibration time of the coupled climate system, which can extend to centennial timescales through deep-ocean heat uptake~\citep{gregory2000predictions} and is handled in the ESMs by their own spin-up. However, the more fitted the propagator, the more it is expected to inherit the timescales of the system.}

{
\add
\paragraph{Sensitivity to parameters} We tested the sensitivity of the \methodname\ forced-response estimate to the truncation level of PCs, the lag and the forcing history, using the RMSD shown in the Taylor diagrams (not shown). A smaller number of PCs degrades all three variables, most strongly SAT and OSAT. A larger number of PCs and longer lags slightly improve SAT and OSAT and slightly increase the RMSD for SLP, so the skill reported in the main text is slightly conservative for SAT and OSAT. Without any forcing history, the OSAT error rises slightly, and one year of history is enough because the decay times of $\boldsymbol{A}$ are short.
}

\clearpage
\newpage

\section{Supplementary Synthetic Example}
\providecommand{\outref}[1]{\textcolor{red}{\ref{#1}}}
\providecommand{\outeqref}[1]{\textcolor{red}{\eqref{#1}}}


To test~\methodname, we construct a synthetic system of the form Eq.~\outref{eq: supp_linear_system} with prescribed modes $\boldsymbol{W}$, eigenvalues, forcing pattern $\boldsymbol{B}$, a one-dimensional forcing time series $\y(t) = y(t)$, and noise statistics. The system contains a slow mode, an ENSO-like oscillatory pair and a set of residual fast modes, and we observe it with white measurement noise. Since Eq.~\ref{eq: supp_linear_system} is a linear system, its forced response is known exactly and equals to the pullback attractor forced part (see Eq.~\ref{eq:decompose_ds_tau} with $\tau=1$). So, by construction, we can attribute estimation errors to the method, the finite record and the measurement noise rather than to the definition of the ground truth. In Section~\ref{app:syn_patterns} we define the spatial patterns of $\A$, then describe the data generation (Section~\ref{app:syn_generation}), the ablation experiments (Section~\ref{app:syn_start}), and report their results (Section~\ref{app:syn_results}).

\subsubsection{Grid and spatial patterns of the modes}\label{app:syn_patterns}

\begin{figure}[ht]
    \centering
    \includegraphics[width=0.6\linewidth]{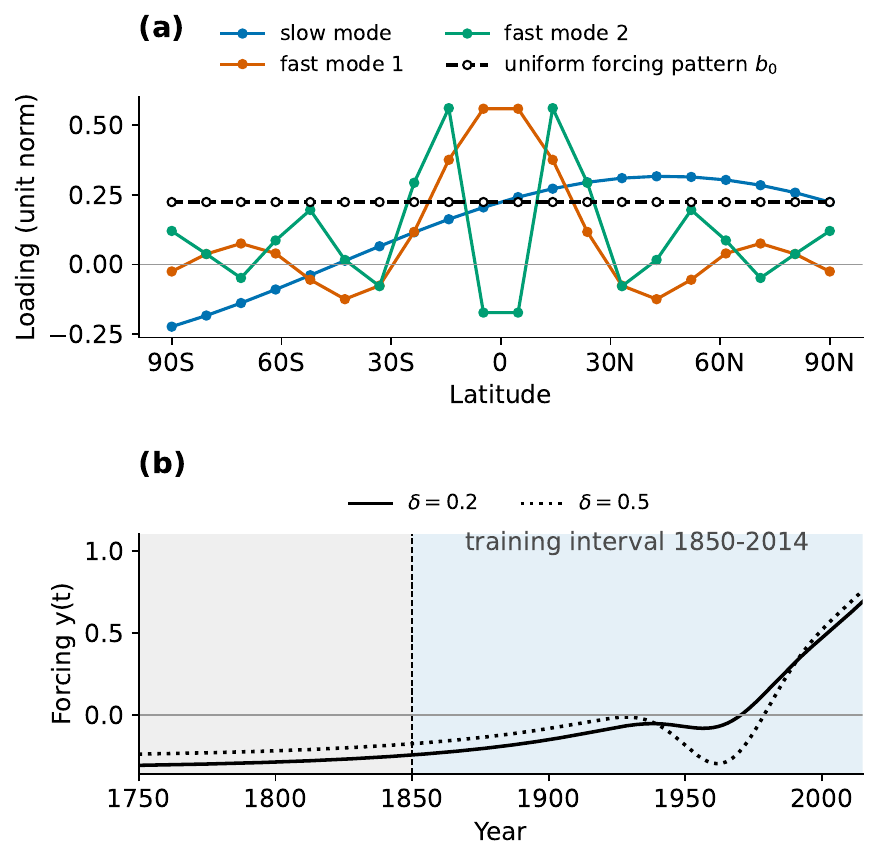}
\caption{Setup of the synthetic example. (a) Unit-norm slow mode $\boldsymbol{w}_1$~\eqref{eq:syn_slow}, fast modes $\boldsymbol{w}_2$~\eqref{eq:syn_fast1} and $\boldsymbol{w}_3$~\eqref{eq:syn_fast2}, and the uniform forcing pattern on the $M=20$ latitude grid~\eqref{eq:syn_grid}. (b) Mean-centered forcing $y(t)$~\eqref{eq:syn_forcing} from 1750 with dip depth $\delta=0.2$ (solid) and $\delta=0.5$ (dotted). The gray region is the end of the spin-up and the blue region is the record from 1850 to 2014.}
    \label{fig:syn_patterns}
\end{figure}

Following the idealized synthetic setting used in ~\citet{Kravtsov2022} and~\citet{Gavrilov2024}, we define the patterns mimicking climate dynamics observed on zonal grid with slow and fast components. 
The state vector $\x(t)\in\R^{M}$ represents a zonally averaged field at $M=20$ equally spaced latitudes from the South Pole to the North Pole,
\begin{equation}\label{eq:syn_grid}
    \phi_i = -\frac{\pi}{2} + \frac{i\,\pi}{M-1}, \qquad i = 0,1,\dots,M-1,
\end{equation}
with spacing $180^\circ/19 \approx 9.47^\circ$ and both poles included. All formulas below measure $\phi$ in radians. 
We use three structured patterns, one slow and two fast, where the two fast patterns together mimic the two-dimensional oscillatory ENSO-like mode recovered from reanalysis (modes 3 and 4 in Fig.~6):
\begin{align}
    a_{1}(\phi) &= \cos\!\left(\phi - \frac{\pi}{4}\right), \label{eq:syn_slow}\\
    a_{2}(\phi) &= \operatorname{sinc}(2\phi) = \frac{\sin(2\pi\phi)}{2\pi\phi}, \label{eq:syn_fast1}\\
    a_{3}(\phi) &= \frac{0.5 - \cos(10\phi)}{0.5 + |10\phi|}. \label{eq:syn_fast2}
\end{align}
The slow pattern~\eqref{eq:syn_slow} and the first fast pattern~\eqref{eq:syn_fast1} follow the idealized model of~\citet{Gavrilov2024}. The former is a global, multidecadal (AMO-like) pattern with its maximum at $45^\circ$N, and the latter is an ENSO-like pattern with enhanced tropical loadings and negative subtropical side lobes. The second fast pattern~\eqref{eq:syn_fast2} mimics the second real pattern of the ENSO-like pair formed by modes $3$ and $4$, $\widehat{\boldsymbol{w}}_3$ and $\widehat{\boldsymbol{w}}_4$ (Fig.~6). The ocean zonal mean of $\widehat{\boldsymbol{w}}_3$ closely follows $\operatorname{sinc}(2\phi)$, which supports~\eqref{eq:syn_fast1}. The ocean zonal mean of $\widehat{\boldsymbol{w}}_4$ has a narrow negative equatorial minimum, associated with the equatorial Pacific, and two positive off-equatorial maxima near $\pm 15^\circ$, which are dominated by positive loadings outside the central and eastern equatorial Pacific, in particular over the Indo-Pacific warm pool, the Indian Ocean and the tropical Atlantic. The form~\eqref{eq:syn_fast2} computed on the selected grid reproduces this structure with a cosine whose amplitude is modulated by a rational envelope (Fig.~\ref{fig:syn_patterns}a).

We evaluate all patterns on the grid~\eqref{eq:syn_grid} and normalize them to unit Euclidean norm, $\boldsymbol{w}_k = \boldsymbol{a}_k/\|\boldsymbol{a}_k\|_2$ ($k=1,2,3$), where $\boldsymbol{a}_k = [a_k(\phi_0),\dots,a_k(\phi_{M-1})]^\top$. Notice that the patterns are not orthogonal: $\boldsymbol{w}_1^\top\boldsymbol{w}_2 = 0.396$, $\boldsymbol{w}_1^\top\boldsymbol{w}_3 = 0.329$ and $\boldsymbol{w}_2^\top\boldsymbol{w}_3 = 0.278$.

For the remaining $M-3=17$ modes, we use a set of orthonormal basis vectors $\boldsymbol{Q}_\perp = [\boldsymbol{q}_4,\dots,\boldsymbol{q}_M] \in \R^{M\times(M-3)}$ of the orthogonal complement of the three structured patterns. We draw this basis once at random and keep it fixed in all experiments, so that the modal matrix is
\begin{equation}\label{eq:syn_W}
    \boldsymbol{W} = \big[\boldsymbol{w}_1,\ \boldsymbol{w}_2,\ \boldsymbol{w}_3,\ \boldsymbol{Q}_\perp\big].
\end{equation}
Thus, all columns of $\boldsymbol{W}$ have unit norm. Notice that the complement modes are not physical. They stand in for unresolved fast variability, and only their aggregate properties are meaningful.

In one of the experiments below (Section~\ref{app:syn_start}), we modify the spatial overlap between the complement and the slow mode and introduce
\begin{equation}\label{eq:syn_overlap}
    \boldsymbol{W}(\alpha) = \big[\boldsymbol{w}_1,\ \boldsymbol{w}_2,\ \boldsymbol{w}_3,\ \boldsymbol{q}_4(\alpha),\dots,\boldsymbol{q}_M(\alpha)\big], \qquad
    \boldsymbol{q}_k(\alpha) = \alpha\,\boldsymbol{w}_1 + \sqrt{1-\alpha^2}\,\boldsymbol{q}_k,
\end{equation}
with the overlap $\alpha\in[0,1)$. Since every $\boldsymbol{q}_k$ is orthogonal to $\boldsymbol{w}_1$, each mode $\boldsymbol{q}_k(\alpha)$ is rotated to make the angle $\arccos\alpha$ with $\boldsymbol{w}_1$, so that its (uncentered) pattern correlation with the slow mode is $\alpha$. Thus $\boldsymbol{W}(0) = \boldsymbol{W}$, which all other experiments use, and $\boldsymbol{W}(\alpha)$ becomes singular as $\alpha \to 1$, when all rotated modes approach $\boldsymbol{w}_1$. 

\subsubsection{Generation of the synthetic data}\label{app:syn_generation}
\label{app:syn_dynamics}\label{app:syn_noise}\label{app:syn_forcing}\label{app:syn_snr}

We generate the data at a monthly step, the native sampling of the data in the main analysis, by
\begin{align}
    \boldsymbol{x}_\mathrm{obs}(t) &= \x(t) + \boldsymbol{\varepsilon}(t), \qquad
    \boldsymbol{\varepsilon}(t)\overset{\text{iid}}{\sim}\mathcal{N}\big(\boldsymbol{0},\,\sigma_{\varepsilon}^2\I\big), \label{eq:syn_obs}\\
    \x(t) &= \A\,\x(t-1) + \boldsymbol{B}\,y(t) + \boldsymbol{\xi}(t), \qquad
    \boldsymbol{\xi}(t)\overset{\text{iid}}{\sim}\mathcal{N}\big(\boldsymbol{0},\,\boldsymbol{W}\boldsymbol{D}\boldsymbol{W}^\top\big), \label{eq:syn_model}\\    
    \A &= \boldsymbol{W}\boldsymbol{\Lambda}_{\mathrm{R}}\boldsymbol{W}^{-1}, \qquad
    \boldsymbol{D} = \mathrm{diag}\big(\sigma_1^2,\ \sigma_p^2,\ \sigma_p^2,\ \sigma_4^2,\dots,\sigma_M^2\big), \label{eq:syn_A}
\end{align}
with $\boldsymbol{W}$ from~\eqref{eq:syn_W} and $t$ in months. The methods only see the measured data $\boldsymbol{x}_\mathrm{obs}(t)$. The measurement noise $\boldsymbol{\varepsilon}$ has the same variance at every latitude, set by the scalar parameter $\sigma_\varepsilon$. We choose the ingredients of Eqs.~\eqref{eq:syn_model} and~\eqref{eq:syn_A} as follows.

\paragraph{Time scales.} We prescribe the eigenvalues of $\A$, and hence the internal time scales, through the real block-diagonal matrix
\begin{equation}\label{eq:syn_LambdaR}
    \boldsymbol{\Lambda}_{\mathrm{R}} = \begin{bmatrix}
        \lambda_1 & & & \\
        & \rho\cos\theta & \rho\sin\theta & \\
        & -\rho\sin\theta & \rho\cos\theta & \\
        & & & \mathrm{diag}(\lambda_4,\dots,\lambda_M)
    \end{bmatrix},
\end{equation}
where the subscript distinguishes it from the complex diagonal $\boldsymbol{\Lambda}$ of Section~2 of the main text. The eigenvalues of $\A$ are $\lambda_1$ for the slow mode, $\rho e^{\pm i\theta}$ for the ENSO-like pair and $\lambda_4,\dots,\lambda_M$ for the complement. Because the $2\times2$ block of $\boldsymbol{\Lambda}_{\mathrm{R}}$ acts only on the coordinates of $\boldsymbol{w}_2$ and $\boldsymbol{w}_3$, the deterministic evolution of the pair is a damped rotation in the plane spanned by these two patterns. We convert decay times and periods to eigenvalues in units of months:
\begin{equation}\label{eq:syn_eigs}
    \lambda_1 = e^{-1/(12\,\tau_1)}, \qquad \rho = e^{-1/(12\,\tau_p)}, \qquad \theta = \frac{2\pi}{12\,P_p},
\end{equation}
with $\tau_1$, $\tau_p$ and $P_p$ in years. The slow decay time $\tau_1$ is the only timescale we vary in the experiments (Section~\ref{app:syn_start}), and unless stated otherwise $\tau_1 = 20$\,yr. The pair has a fixed decay time $\tau_p = 1.5$\,yr and period $P_p = 4$\,yr. We choose these values so that the pair resembles the observed ENSO variability. A damped oscillating pair has a single broad spectral peak centered at its period, with a width set by the decay time. With $P_p = 4$\,yr and $\tau_p = 1.5$\,yr, the spectrum has its half maximum between about $2.8$ and $7.5$\,yr, close to the conventional $2$ to $7$\,yr ENSO band. We draw the complement eigenvalues once, uniformly from $(0, 0.1)$, and keep them fixed in all experiments, so these modes decorrelate within a month and represent a nearly white residual.

\paragraph{Dynamic noise covariance.} In modal coordinates $\z(t) = \boldsymbol{W}^{-1}\x(t)$ (cf. Eq.~2.7), the model~\eqref{eq:syn_model} reads
\begin{equation}\label{eq:syn_zmodel}
    \z(t) = \boldsymbol{\Lambda}_{\mathrm{R}}\z(t-1) + \boldsymbol{\gamma}\,y(t) + \hat{\boldsymbol{\xi}}(t),
    \qquad \boldsymbol{\gamma} = \boldsymbol{W}^{-1}\boldsymbol{B}, \qquad
    \hat{\boldsymbol{\xi}}(t) = \boldsymbol{W}^{-1}\boldsymbol{\xi}(t)\sim\mathcal{N}(\boldsymbol{0},\boldsymbol{D}).
\end{equation}
The covariance $\boldsymbol{W}\boldsymbol{D}\boldsymbol{W}^\top$ therefore gives each mode its own independent noise source, and the equal $\sigma_p^2$ of the two pair components keeps the oscillation isotropic in its plane. As a result, the stationary modal processes are uncorrelated, with theoretical variances
\begin{equation}\label{eq:syn_modal_var}
    s_1^2 = \frac{\sigma_1^2}{1-\lambda_1^2}, \qquad
    s_p^2 = \frac{\sigma_p^2}{1-\rho^2}, \qquad
    s_k^2 = \frac{\sigma_k^2}{1-\lambda_k^2}\ \ (k=4,\dots,M).
\end{equation}
We specify $\boldsymbol{D}=\mathrm{diag}(\sigma_1^2, \sigma_p^2,\sigma_p^2, \sigma_4^2, \ldots, \sigma_M^2)$ through three target modal variances, $s_1^2$ (slow mode), $s_p^2$ (each pair component) and a common $s_c^2$ for the complement modes:
\begin{equation}\label{eq:syn_sigma}
    \sigma_1^2 = s_1^2(1-\lambda_1^2), \qquad
    \sigma_p^2 = s_p^2(1-\rho^2), \qquad
    \sigma_k^2 = s_c^2(1-\lambda_k^2)\ \ (k=4,\dots,M).
\end{equation}
This ties the noise of each mode to its decay time and $s_1, s_p, s_c$ which are set below. 

\paragraph{Forcing signal.} We choose the forcing as a smooth exponential rise with a Gaussian dip centered on 1965,
\begin{equation}\label{eq:syn_forcing}
    F(t) = \,\exp\!\left(\frac{t - 2014}{65}\right) - \delta \,\exp\!\left(-\frac{1}{2}\left(\frac{t - 1965}{15}\right)^{2}\right),
\end{equation}
with $t$ in calendar years, evaluated at monthly steps. It loosely mimics the growing CO2 signal with the mid-twentieth-century pause in warming, and its depth $\delta$ controls how far the forcing departs from a monotone trend. We use two dip depths (Fig.~\ref{fig:syn_patterns}b): $\delta = 0.2$ gives a weak dip, in which the rise pauses from the late 1930s to the mid-1950s with a slight decrease, and $\delta = 0.5$ gives a pronounced dip, which highlights the dynamical response in our example, since the forced response lags and smooths the forcing rather than following it instantaneously. The analyzed record covers January 1850 to December 2014 ($T = 1980$ months). We center the forcing on the analyzed record, $y(t) = F(t) - \overline{F}$, where $\overline{F}$ is the mean of $F$ over the record. Before the rise, $y(t)$ approaches the constant level $y_0 = -\overline{F}$.

\paragraph{Forcing pattern.} We use a spatially uniform input pattern, which does not single out any latitude, and scale it as
\begin{equation}\label{eq:syn_B}
    \boldsymbol{B} = \frac{\boldsymbol{b}_0}{\big\|(\I - \A)^{-1}\boldsymbol{b}_0\,y_0\big\|_2}, \qquad \boldsymbol{b}_0 = [1,\dots,1]^\top .
\end{equation}
Notice that $\boldsymbol{B}$ is the pattern with which the forcing enters the dynamics, whereas the spatial pattern of the forced response follows from $\A$. With this scaling, the pre-industrial forced state, that is, the quasi-equilibrium $(\I - \A)^{-1}\boldsymbol{B}\,y_0$ reached under the constant pre-industrial forcing $y_0$ before the rise, has the norm equal to 1 for any propagator $\A$. The scaling changes only the total amplitude of the forced response, not its spatial pattern or its temporal shape, and its purpose is to keep the same pre-industrial state in experiments with different $\A$.

\paragraph{Variance budget.} Since the forced response is deterministic, the modal processes are uncorrelated, and the measurement noise is independent of both, the theoretical total variance of $\boldsymbol{x}_\mathrm{obs}(t)$ on the observed interval decomposes into five components, $V_{\mathrm{tot}} = V_f + V_1 + V_p + V_c + V_\varepsilon$,
namely the forced-response variance $V_f$, the internal variances of the slow mode, the pair and the complement, $V_1$, $V_p$ and $V_c$, and the measurement noise variance $V_\varepsilon$, each summed over the grid points. The forced variance $V_f$ is fixed by the forcing and the dynamics (see the next paragraph), and we specify the other four components through the ratio $V_f : V_1 : V_p : V_c : V_\varepsilon$. This determines the previously introduced parameters,
\begin{equation}\label{eq:syn_svar}
    s_1^2 = V_1, \qquad s_p^2 = \frac{V_p}{2}, \qquad s_c^2 = \frac{V_c}{M-3}, \qquad \sigma_\varepsilon^2 = \frac{V_\varepsilon}{M},
\end{equation}
and the theoretical signal-to-noise ratio:
\begin{equation}\label{eq:syn_snr}
    \mathrm{SNR} = \frac{V_f}{V_1 + V_p + V_c + V_\varepsilon}.
\end{equation}
Unless stated otherwise, we use $V_f : V_1 : V_p : V_c : V_\varepsilon = 1 : 1 : 1 : 0.99 : 0.01$, so that the forced response, the slow mode, the pair, and the complement together with the tiny measurement noise carry equal variances, and $\mathrm{SNR} = 1/3$. Because the components are set relative to $V_f$, which depends on the eigenvalues, the SNR stays fixed when we change $\tau_1$.

\paragraph{Spin-up and ground truth.} The forced response at the start of the record depends on the forcing history, and the internal component must forget the initial state. To this end, we run~\eqref{eq:syn_model} from $\x = \boldsymbol{0}$ over a spin-up of $T_s = 100\max(\tau_1, \tau_p)$ years ($2000$\,yr for $\tau_1 = 20$\,yr) followed by the record, with $y(t)$ given by~\eqref{eq:syn_forcing} also during the spin-up, discard the spin-up, and add the measurement noise~\eqref{eq:syn_obs} on the record. The memory of the initial state decays as $e^{-T_s/\max(\tau_1,\tau_p)} = e^{-100}$, so the system is on its pullback attractor when the record starts. We obtain the exact forced response by running the deterministic part of~\eqref{eq:syn_model} ($\boldsymbol{\xi}=\boldsymbol{0}$) from the same initial state over the same spin-up,
\begin{equation}\label{eq:syn_forced}
    \x^{(f)}(t) = \A\,\x^{(f)}(t-1) + \boldsymbol{B}\,y(t),
\end{equation}
which is common to all realizations. The forced-response variance in the variance budget is
\begin{equation}\label{eq:syn_Vf}
    V_f = \sum_{i=1}^{M}\mathrm{Var}_t\big(x_i^{(f)}(t)\big),
\end{equation}
with the time variance taken over the record about its mean.

\paragraph{Realizations.} For each parameter configuration below, we use $1000$ independent realizations of the noises $\boldsymbol{\xi}$ and $\boldsymbol{\varepsilon}$ with a common forcing, drawn from the same random seeds, so that configurations with the same spin-up length share identical noise draws.

\subsubsection{Experiments}\label{app:syn_start}

Starting from the reference configuration ($\delta=0.2$, $\tau_1=20$\,yr, $\alpha = 0$, $V_f : V_1 : V_p : V_c : V_\varepsilon = 1 : 1 : 1 : 0.99 : 0.01$), we vary one property at a time (Table~\ref{tab:syn_studies}). Every level uses $1000$ realizations, and we repeat all experiments with the pronounced dip $\delta = 0.5$.

\begin{table}[ht]
    \centering
    \caption{Experiments on the synthetic system. Each varies one property of the reference configuration ($\delta=0.2$, $\tau_1=20$\,yr, $\alpha=0$, $V_f : V_1 : V_p : V_c : V_\varepsilon = 1 : 1 : 1 : 0.99 : 0.01$, $\mathrm{SNR}=1/3$).}
    \label{tab:syn_studies}
    \small
    \begin{tabular*}{\textwidth}{@{\extracolsep{\fill}}lll@{}}
        \hline
        Experiment & Levels & Change \\
        \hline
        Total SNR & $\mathrm{SNR} = 1/30,\ 1/10,\ 1/3,\ 1,\ 3,\ 10,\ 30$ & $V_1, V_p, V_c, V_\varepsilon$ scaled by $(1/3)/\mathrm{SNR}$ \\
        Slow timescale & $\tau_1 = 1,\ 2,\ 5,\ 10,\ 20,\ 40,\ 80$\,yr & $\tau_1$ at fixed ratios (fixed SNR) \\
        Measurement noise & $V_\varepsilon/V_f = 0.0001,\ 0.001,\ 0.01,\ 0.1,\ 1,\ 10$ & $V_\varepsilon$ alone (SNR decreases) \\
        Spatial overlap & $\alpha = 0,\ 0.3,\ 0.5,\ 0.7,\ 0.9,\ 0.95,\ 0.96,\ 0.97,\ 0.99$ & $\boldsymbol{W}(\alpha)$, Eq.~\eqref{eq:syn_overlap} \\
        Joint & SNR levels $\times$ $\tau_1$ levels & both, as above \\
        \hline
    \end{tabular*}
\end{table}

We vary the total SNR by scaling $V_1$, $V_p$, $V_c$ and $V_\varepsilon$ by a common factor, which tests how much internal variability a method tolerates before the forced response is lost in it. We vary the slow timescale $\tau_1$ while keeping the full ratio $V_f : V_1 : V_p : V_c : V_\varepsilon$, and hence the SNR, fixed, so the experiment isolates the effect of the timescale itself: as $\tau_1$ grows, the slow forced response lags the forcing more and the slow internal variability evolves on timescales comparable to the $165$-yr record, which makes the two harder to separate. We vary the measurement noise through $V_\varepsilon$ alone, so the SNR decreases as $V_\varepsilon$ grows. This noise is not propagated by the dynamics and enters the fit as an error in the regressors, which biases the fitted eigenvalues toward zero, so the experiment tests how a method responds to observational error as opposed to internal variability. We vary the overlap $\alpha$ between the complement modes and the slow mode~\eqref{eq:syn_overlap}, which makes the unresolved fast variability project increasingly onto the slow pattern, and tests whether a method can still separate the slow mode when the modal basis is poorly conditioned. Finally, we vary the SNR and $\tau_1$ jointly on a $7\times7$ grid, since these are the two properties that govern the separation of the forced response from the internal variability.

In each experiment, we compare~\methodname~with LIM, LIM-opt and LR used as in the main analysis. We treat each realization as a single observed record: we center $\boldsymbol{x}_{\mathrm{obs}}(t)$ by its record mean and compute its EOFs once, keeping all $M=20$ so that no variance is discarded. All methods share this basis, fit the projected data, and add the mean back to their estimates. All methods are applied as in the main text and use a lag of $\tau = 1$ month. \methodname~obtains its forced response by integrating the fitted model from zero over the full forcing history, including the spin-up, so it sees the same forcing history as the generator.

Our primary score is the forced-response correlation, the Pearson correlation between the estimated and the true forced response after removing each latitude's time mean and vectorizing over latitudes and time. In other words, it measures whether the shape of the forced response is recovered, regardless of a constant offset. For the methods with a propagator (\methodname~and LIM), we also score the eigendecomposition, specifically the decay time and the pattern of the extracted slowest mode. We select the largest real eigenvalue $\hat{\lambda}$ of the fitted propagator, which the method itself identifies as its slowest non-oscillating mode, and report the decay time $\hat{\tau}_1 = -\tau/(12\log\hat{\lambda})$ in years and the slow-pattern correlation $|\hat{\boldsymbol{v}}^\top\boldsymbol{w}_1|/\|\hat{\boldsymbol{v}}\|_2$, where $\hat{\boldsymbol{v}}$ is the corresponding eigenvector mapped to grid space. We exclude fits with $\hat{\lambda}\geq 1$, or without a positive real eigenvalue, which have no decay time, from the statistics of both $\hat{\tau}_1$ and the slow-pattern correlation.

\subsubsection{Results}\label{app:syn_results}

In our setting, a mode accumulates the forcing over its own memory, so for a forcing that varies slowly, on multidecadal timescales, the response of a mode grows with its decay time. The slow mode, with a decay time of decades, therefore responds much more strongly than the pair and the complement modes, whose memory is of a few years or less than a month, and it loses this dominance only when it becomes fast itself.
Hence, the forced response is carried almost entirely by the slow mode. This is consistent with the large forced variance fractions of the slow modes of 20CRv3 in the main analysis (row 3 of Fig.~8 of the main text). As a consequence, the forced response and the slow internal variability in our example share almost the same spatial pattern $\boldsymbol{w}_1$ and, in the default configuration, the same variance ($V_1 = V_f$), so they cannot be separated in space and have to be separated in time, through the dynamics and the forcing.

\begin{figure}[ht]
    \centering
    \includegraphics[width=\linewidth]{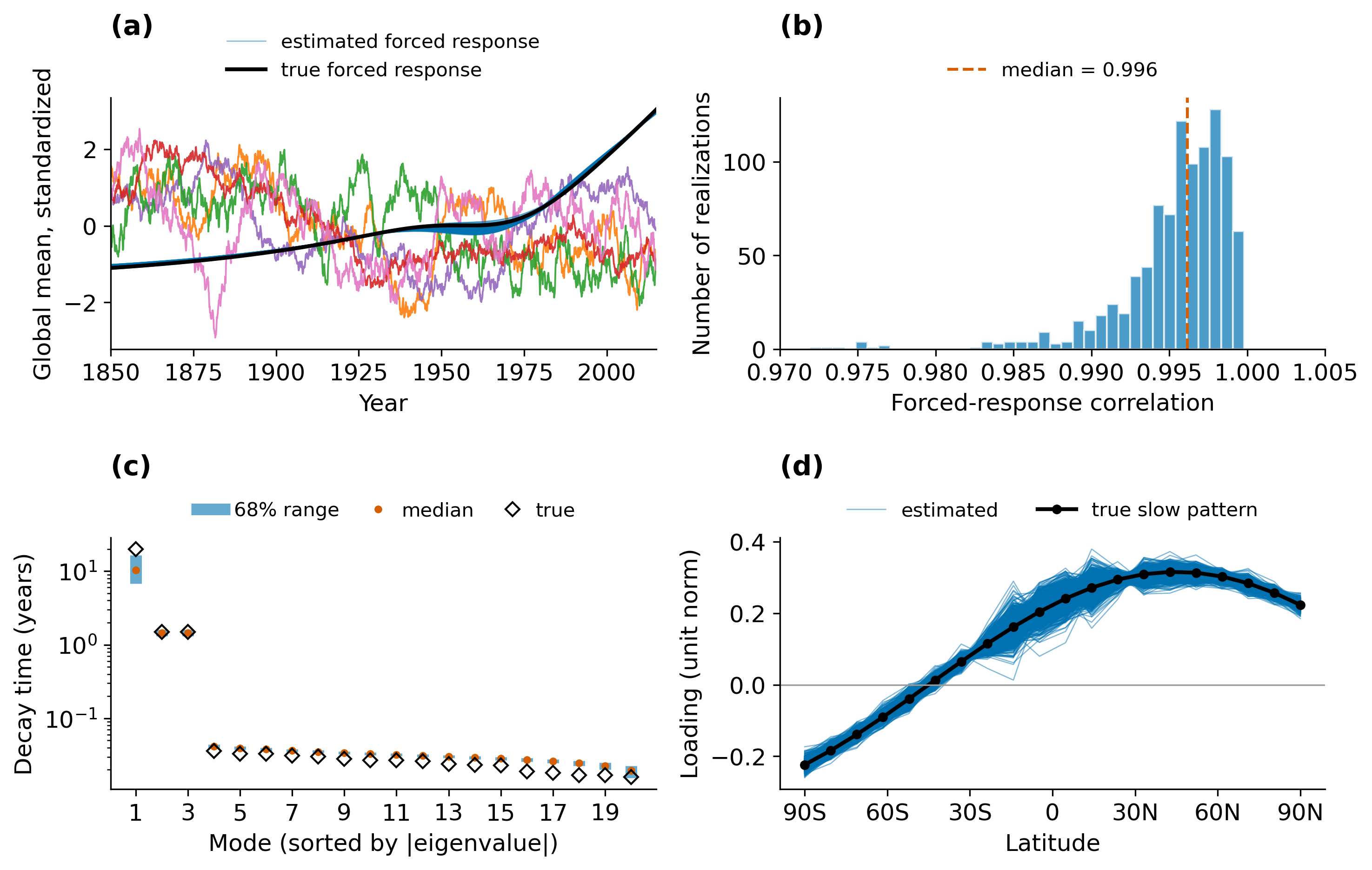}
    \caption{\methodname~fitted to each of the $1000$ realizations of the reference configuration ($\delta=0.2$, $\tau_1=20$\,yr, $\mathrm{SNR}=1/3$). (a) Global mean of the estimated forced responses (blue) and of the true forced response (black), and the slow internal variability $\boldsymbol{w}_1 z_1(t)$ of five realizations (colors), each standardized to zero mean and unit variance. (b) Histogram of the forced-response correlation over the realizations, with its median (dashed). (c) Median (orange) and 68\% range (16th to 84th percentiles, blue) of the estimated decay times, and the true decay times (diamonds), with modes sorted by $|\lambda|$. (d) Estimated slow-mode patterns (blue) and the true slow pattern $\boldsymbol{w}_1$ (black).}
    \label{fig:syn_default}
\end{figure}

\paragraph{Default configuration.} Fig.~\ref{fig:syn_default} shows \methodname~fitted to each of the $1000$ realizations of the default configuration. 
The estimated forced responses follow the true one closely, including the weak mid-century dip, and their spread is largest around the dip (Fig.~\ref{fig:syn_default}a). The slow internal variability shown for five realizations fluctuates on multidecadal timescales comparable to the dip, which makes the separation non-trivial. The median forced-response correlation metric is $0.996$, and nearly all realizations lie above $0.98$ (Fig.~\ref{fig:syn_default}b). \methodname~recovers the decay times of the pair and of the complement modes, and the slow pattern closely (Fig.~\ref{fig:syn_default}c,d). The slow decay time, however, is underestimated. Its median is about half of the true $20$\,yr, with a wide spread. This uncertainty is due to the fact that the record covers only a few slow decay times, leading to a poorer statistics for good estimation of a persistent autoregressive coefficient. Yet, this does not prevent quite accurate forced-response shape estimation.

\begin{figure}[ht]
    \centering
    \includegraphics[width=\linewidth]{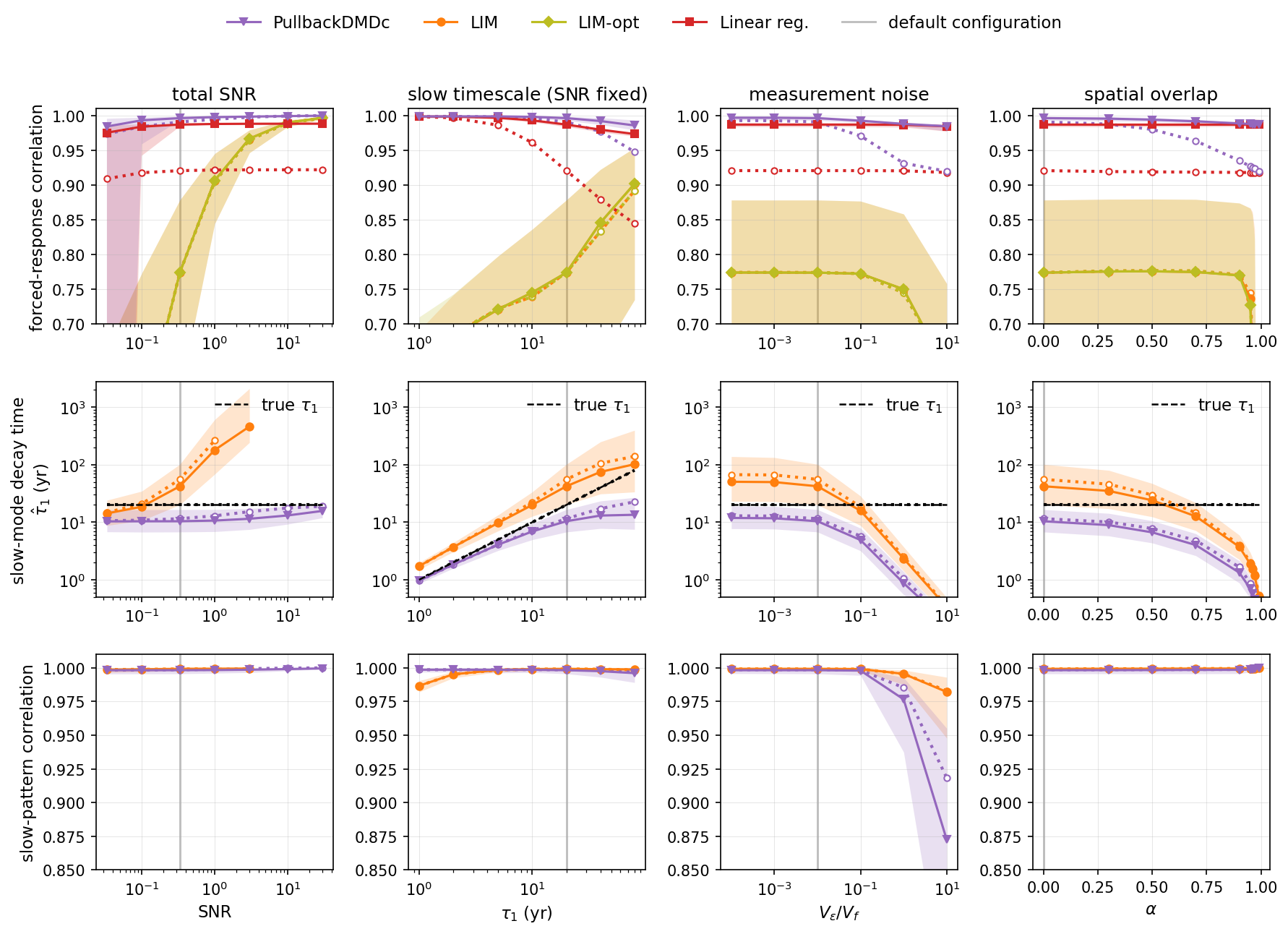}
    \caption{Experiments of Table~\ref{tab:syn_studies} for the dip depths $\delta=0.2$ (solid lines with shading) and $\delta=0.5$ (dotted lines with open circles). The columns show the total SNR, the slow timescale $\tau_1$ at a fixed ratio $V_f : V_1 : V_p : V_c : V_\varepsilon$, the measurement noise $V_\varepsilon/V_f$ and the spatial overlap $\alpha$ between the complement modes and the slow mode. The vertical gray line marks the reference configuration. The top row shows the forced-response correlation of the four methods. The middle row shows the fitted slow decay time $\hat{\tau}_1$ of~\methodname~and LIM, and the true $\tau_1$ (dashed), excluding fits whose slow eigenvalue does not decay. The bottom row shows the correlation between the fitted and the true slow pattern. Lines show the median over $1000$ realizations, and the shading shows the 68\% range (16th to 84th percentiles) for $\delta=0.2$.}
    \label{fig:syn_sweeps}
\end{figure}

\paragraph{Ablation sweeps.} In the results shown in Fig.~\ref{fig:syn_sweeps} we vary one property at a time (Table~\ref{tab:syn_studies}). At the default configuration, the forced-response correlation is close to $1$ for \methodname, slightly lower for LR, and below $0.8$ for LIM and LIM-opt, which nearly coincide in all experiments. With the pronounced dip $\delta = 0.5$ the forced response deviates stronger from the forcing shape, which is the only shape LR can reconstruct, and therefore LR drops to about $0.92$ correlation, whereas \methodname~changes little.

All methods estimate the forced response worse as the SNR decreases, which is natural, but they differ strongly in their sensitivity (first column of Fig.~\ref{fig:syn_sweeps}). \methodname~is the most accurate at every SNR and degrades only slightly. For the weaker dip ($\delta = 0.2$) it is close to LR, but the difference is much larger for the pronounced dip ($\delta = 0.5$), where \methodname~remains accurate, as discussed above. LIM is by far the most sensitive, and its correlation falls from nearly $1$ at the highest SNR to below the plotted range at the lowest. Its estimate is the projection of the data onto the least-damped mode, that is, the slow mode, so it retains the slow internal variability together with the forced response, which limits its correlation unless the SNR is high. At the lowest SNR, the forced response carries only a few percent of the total variance, and in a part of the realizations the response estimated by either LR or ~\methodname has the opposite sign, so the correlation is negative and the percentile band extends to negative values. This is a weak-signal effect: a single record does not constrain the sign of the estimated forced response when internal variability dominates, while the median correlation of \methodname~remains high. It mirrors the larger spread of the forced-response estimates across ensemble members for sea level pressure than for temperature in the main text (Fig.~2).

The fitted slow decay time of \methodname~and LIM (middle row of the first column of Fig.~\ref{fig:syn_sweeps}) also behaves differently. \methodname~underestimates it at low SNR and approaches the true value as the SNR grows. LIM is below the truth at low SNR, where the data are dominated by internal variability and the finite-record bias of the autoregressive fit discussed above applies to it as well. It then overestimates the decay time increasingly with the SNR, and at high SNR most of its fits no longer decay, so the curve ends. The reason is that LIM gets affected by the presence of the forced response, which is a smooth trend over the record, described by autoregression only as persistence, so the fitted slow eigenvalue is pushed towards one and beyond, and the larger the forced-response signal, the stronger this effect. \methodname~assigns the trend to the forcing term and estimates the eigenvalue from the internal variability. At the same time, SNR is not a problem for the slow mode pattern reconstruction (bottom row of the first column of Fig.~\ref{fig:syn_sweeps}).

As $\tau_1$ grows at a fixed SNR (second column of Fig.~\ref{fig:syn_sweeps}), the correlation of \methodname~decreases slowly, to about $0.99$ ($\delta = 0.2$) and $0.95$ ($\delta = 0.5$) at $\tau_1 = 80$\,yr. LR skill decreases faster, to about $0.97$ and $0.85$, because the forced response lags the forcing more, and its loss is largest for the deeper dip, which is smoothed out by the slow response. The slow internal variability then also evolves on timescales comparable to the $165$-yr record, which makes it harder to separate from the forced response. LIM improves with $\tau_1$ but stays below about $0.9$. The slow mode response to forcing dominates as $\tau_1$ increases, and the least-damped mode pattern becomes a better estimate, but the projection onto it also retains the slow internal variability. 
For short $\tau_1$, \methodname~extracts the slow timescale correctly, and its estimate saturates below the truth only as $\tau_1$ approaches the record length, consistent with the finite-record bias above. LIM overestimates $\tau_1$, as the forced trend drags its slow eigenvalue towards the timescale of the forced response. Both methods recover the slow pattern (bottom row of the second column of Fig.~\ref{fig:syn_sweeps}).

\begin{figure}[ht]
    \centering
    \includegraphics[width=\linewidth]{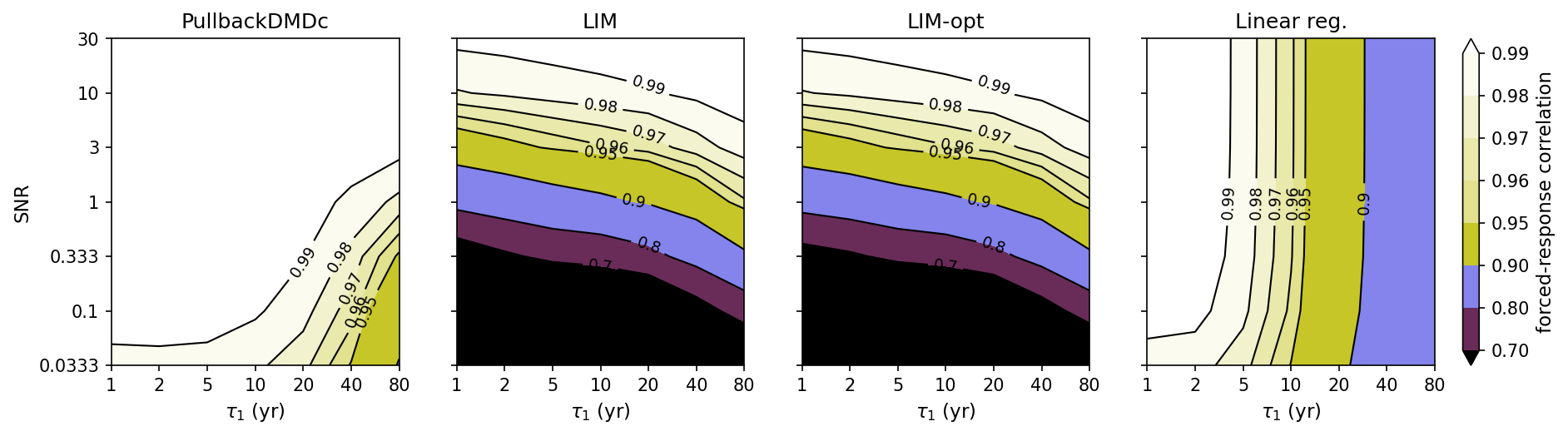}
    \caption{Same as Fig.~1 of the main text, but for the dip depth $\delta=0.5$.}
    \label{fig:syn_phase05}
\end{figure}

The 2D colormaps of the joint experiment confirm these conclusions (Fig.~1 of the main text for $\delta = 0.2$ and Fig.~\ref{fig:syn_phase05} for $\delta = 0.5$). The skill of LIM depends mainly on the SNR, so its contours are nearly horizontal. The skill of LR depends on $\tau_1$, so its contours are vertical. \methodname~departs from near-perfect skill only at long $\tau_1$ and low SNR, and it matches or exceeds LR everywhere.

Measurement noise up to the reference level $V_\varepsilon/V_f = 0.01$ has little effect (third column of Fig.~\ref{fig:syn_sweeps}). For the forced response (top row), LR is insensitive to the measurement noise, because white noise is uncorrelated with the forcing and averages out in the regression. \methodname~keeps its skill up to the reference noise level, and for larger noise its correlation decreases towards that of LR, mainly for the pronounced dip. LIM degrades at large noise, because its estimate is a projection of the data and therefore contains the projected measurement noise. Larger noise also enters the fit as an error in the regressors and biases the fitted eigenvalues towards zero. The fitted slow decay time of \methodname~and LIM therefore shrinks with the noise level, down to a few months at the largest noise, close to the timescale of the white noise itself (middle row). Once the propagator fitted by \methodname~has lost its memory, its forced response reduces to a regression on the instantaneous forcing, as in LR, which explains the convergence of the two methods in the top row. The slow pattern is much more robust (bottom row). \methodname~recovers it almost exactly at noise levels where the decay time is already strongly underestimated, and its pattern correlation decreases noticeably only when the noise variance becomes comparable to or larger than the forced variance. A mode can thus be detected correctly even when its fitted decay time is far too short. This is why in the main text we assess the timescales of the modes with the autocorrelation function of their internal time series rather than with the eigenvalues alone (Figs.~6 and~8 of the main text). White measurement noise only adds a spike at zero lag to this function and rescales it at all other lags, so its decay and its zero crossings are preserved.

The spatial overlap $\alpha$ shows the same behaviour (fourth column of Fig.~\ref{fig:syn_sweeps}). The overlap changes only the conditioning of $\boldsymbol{W}$, not the modal dynamics. Without measurement noise, the fitted eigenvalues do not depend on it, since the least-squares fit is equivariant to the choice of the modal basis. With the small reference measurement noise, however, the noise enters the slow modal coordinate through the first row of $\boldsymbol{W}^{-1}$, whose norm grows rapidly as $\alpha \to 1$. The overlap thus acts as an amplified measurement noise in the slow mode. LR barely depends on $\alpha$ (top row). The fitted decay time of \methodname~falls from about $10$\,yr to well below a year (middle row), and its forced-response correlation decreases towards that of LR, for the same reason as above. The slow pattern is still recovered by both \methodname~and LIM (bottom row). LIM keeps its correlation up to large overlaps and then collapses abruptly as $\alpha$ approaches $1$, presumably because its fitted slow decay time falls below the decay time of the pair, so that its least-damped mode becomes the oscillating pair.

In summary, \methodname~recovers the shape of the forced response in all experiments. Its median correlation is about $0.9$ or higher and close to $1$ in most configurations. Its advantage over LR comes from the dynamics and grows with the slow timescale and the depth of the dip. Its advantage over LIM comes from separating the forced response from the slow internal variability that shares its pattern. The fitted slow decay time is less robust than the forced response. \methodname~extracts it correctly for short timescales, but it is biased low when the timescale approaches the record length, and measurement noise, amplified by an ill-conditioned modal basis, shortens it further. LIM instead overestimates it, because the forced trend drags it towards the timescale of the forced response. The slow pattern is recovered almost everywhere, even when the fitted decay time is far too short.

\subsubsection{Limitations}\label{app:syn_discussion}

The synthetic system is linear, so its forced response is defined unambiguously. It is the forced part of the pullback attractor, and it also equals the mean over realizations of the noise, because the noise enters linearly and has zero mean. For nonlinear systems, whose pullback attractor is a set with a time-dependent distribution of states (Section~2 of the main text), the forced response is defined operationally as the mean of this distribution, and we hypothesize that \methodname~applied to a single realization approximates this mean (Section~3a of the main text). We expect this linear approximation to hold when the forcing shifts the mean state but changes the shape of the distribution around it only weakly.

As noted in Section~3a of the main text, the nonlinear and fast dynamics that the linear model cannot represent act as an effective noise. Part of it plays the role of the dynamical noise $\boldsymbol{\xi}$ in Eq.~\eqref{eq:syn_model}. Part of it enters the fit like the measurement noise $\boldsymbol{\varepsilon}$, as an error in the regressors that the fitted dynamics do not propagate. The measurement-noise experiments above show what to expect from this second part. It shortens the fitted decay times, while the slow patterns and the forced response stay close to the truth. This supports the interpretation of the decay times as lower bounds on the physical timescales (Section~4 of the main text) and the use of autocorrelation functions to assess the timescales of the modes. It also motivates fitting the climate data at a lag longer than the sampling step (Section~2c of the main text), since fast dynamics that decorrelate within the lag contribute less to the fitted propagator.

Finally, the synthetic forcing is known exactly and enters through a single time series. In the real application, the forcing time series are themselves uncertain and several forcings act with different response patterns. As explained in Section~2b of the main text, the part of the response that the forcing inputs do not represent appears in the internal component. The synthetic experiments therefore show the behaviour of \methodname~under ideal knowledge of the forcing, while the sensitivity to the choice of forcing predictors is assessed with the ESM ensembles in Section~3b of the main text.

\clearpage
\newpage

\appendixtitle{Supplementary Figures}
\begingroup
\renewcommand{\arraystretch}{0.88}
\rowcolors{2}{gray!15}{white}

\begin{longtable}{@{}>{\raggedright\arraybackslash}m{0.15\linewidth}
                   >{\raggedright\arraybackslash}m{0.70\linewidth}@{}}
\caption{Overview of the Supplementary Figures.}
\label{tab:supp_roadmap} \\

\rowcolor{gray!30}
\textbf{Figure(s)} & \textbf{Description} \\
\endfirsthead

\multicolumn{2}{@{}l}{\small\itshape Table~\ref{tab:supp_roadmap} (continued)} \\
\rowcolor{gray!30}
\textbf{Figure(s)} & \textbf{Description} \\
\endhead

\multicolumn{2}{@{}r}{\small\itshape Continued on next page} \\
\endfoot

\endlastfoot
\added{Figs.~\ref{fig:syn_patterns}--\ref{fig:syn_phase05}}
&
\added{Synthetic climate prototype experiments: system setup, verification of forced-response estimates against exact ground truth, and parameter sensitivity analyses.} \\

Fig.~\ref{fig:taylor_tas}
&
Evaluation of forced-response estimates over the full historical period (1850--2014) using Taylor diagrams for SAT. \\
Fig.~\ref{fig:trend_taylors}
&
Comparison of estimated forced-response trends against the large ensemble mean over 1850--2014. \\
\added{Fig.~\ref{fig:trend-map-tas}--\ref{fig:classic_taylor_trend_window}}
&
\added{Forced-response trends over 1985--2014: trend maps for SAT and SLP (OSAT in the main text), and Taylor diagrams for SAT, OSAT and SLP.} \\
Fig.~\ref{fig:classic_taylor_short}
&
Evaluation of forced-response estimates over the shorter period (1950--2014) using Taylor diagrams for SLP, SAT and OSAT. \\
Fig.~\ref{fig:eigs_supp}
&
Comparison of the decay times and oscillation frequencies of the leading dynamical modes between LIM and~\methodname~variants for SAT and OSAT across ESMs and reanalysis. \\
Fig.~\ref{fig:B}
&
\added{Spatial patterns of the input matrix $\boldsymbol{B}$ for LR and \methodname~(OSAT and SLP).} \\
Fig.~\ref{fig:gm_tas}--\ref{fig:gm_psl_n65}
&
Time series of forced-response estimates over the full historical period (1850--2014): \added{global-mean SAT and SLP averaged south and north of $65^\circ$}. \\
\added{Fig.~\ref{fig:gm_short_psl}--\ref{fig:gm_short_tas_ocean}}
&
Time series of forced-response estimates over the shorter period (1950--2014): \added{SLP averaged south of $65^\circ$S, and global-mean SAT and OSAT}. \\
\added{Fig.~\ref{fig:gm_20crv3}}
&
\added{Forced-response estimates from 20CRv3: global-mean OSAT and SAT, and SLP averaged south of $65^\circ$S, with the ESM ensemble means as a reference.} \\
\added{Fig.~\ref{fig:seasonal_drift_esm}}
&
\added{Change of the seasonal cycle (DJF$-$JJA trends of SAT and SLP) in the ESM ensemble means, compared with 20CRv3.} \\
Fig.~\ref{fig:psd_tas_ocean}
&
Power spectral density of the internal-variability time series of the leading OSAT modes across ensemble members and 20CRv3. \\
Fig.~\ref{fig:mode_selection}
&
Justification for retaining four dynamical modes in the spatial pattern and time series analyses. \\
Fig.~\ref{fig:modes_tas_cesm2}--\ref{fig:modes_tas_cesm5}
&
Example decompositions of \added{OSAT} into spatial patterns, internal variability, and forced response for CESM2, MIROC6, and CanESM5. \\
Fig.~\ref{fig:forced_fraction_vs_timescale}--\ref{fig:acf_efolding_tas_ocean}
&
Relationship between forced fraction and dominant internal-variability timescale\added{, and a comparison of ACF e-folding times with dominant timescales and DMD decay times}. \\
\added{Fig.~\ref{fig:supp_eigs}}
&
\added{Eigenvalues of $\boldsymbol{A}$ in the complex plane for OSAT across 20CRv3 and ESM members, grouped by decay rank.} \\
\end{longtable}
\endgroup

\begin{figure}[H]
    \centering
    \includegraphics[width=.4\linewidth]{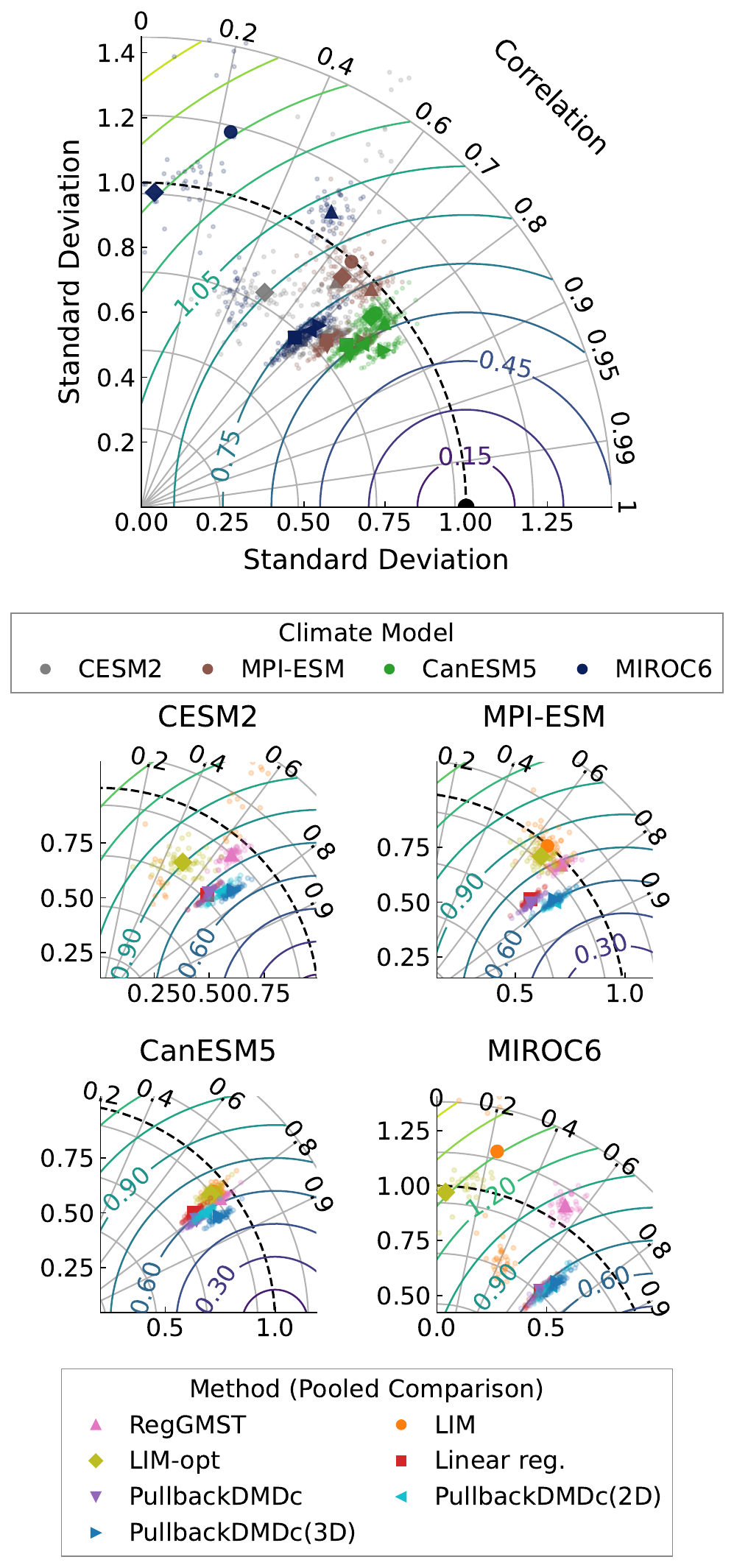}
    \caption{    
    Same as in Fig. 2 (main text) but for SAT.
    Across methods, the SAT forced response is most difficult to estimate in MIROC6 and easiest in CanESM5, with CESM2 and MPI-ESM falling between these extremes, as reflected by the spread in method performance. Within each ESM, at least one~\methodname~variant achieves the best overall performance, with~\methodname(3D) generally performing best, particularly for CESM2 and CanESM5. Across all ESMs,~\methodname~variants tend to slightly underestimate the standard deviation of the forced response while maintaining low RMSD relative to the large ensemble mean.}
    \label{fig:taylor_tas}
\end{figure}

\begin{figure}[H]
    \centering
    \begin{subfigure}[b]{0.3\linewidth}
        \centering
        \includegraphics[width=\textwidth]{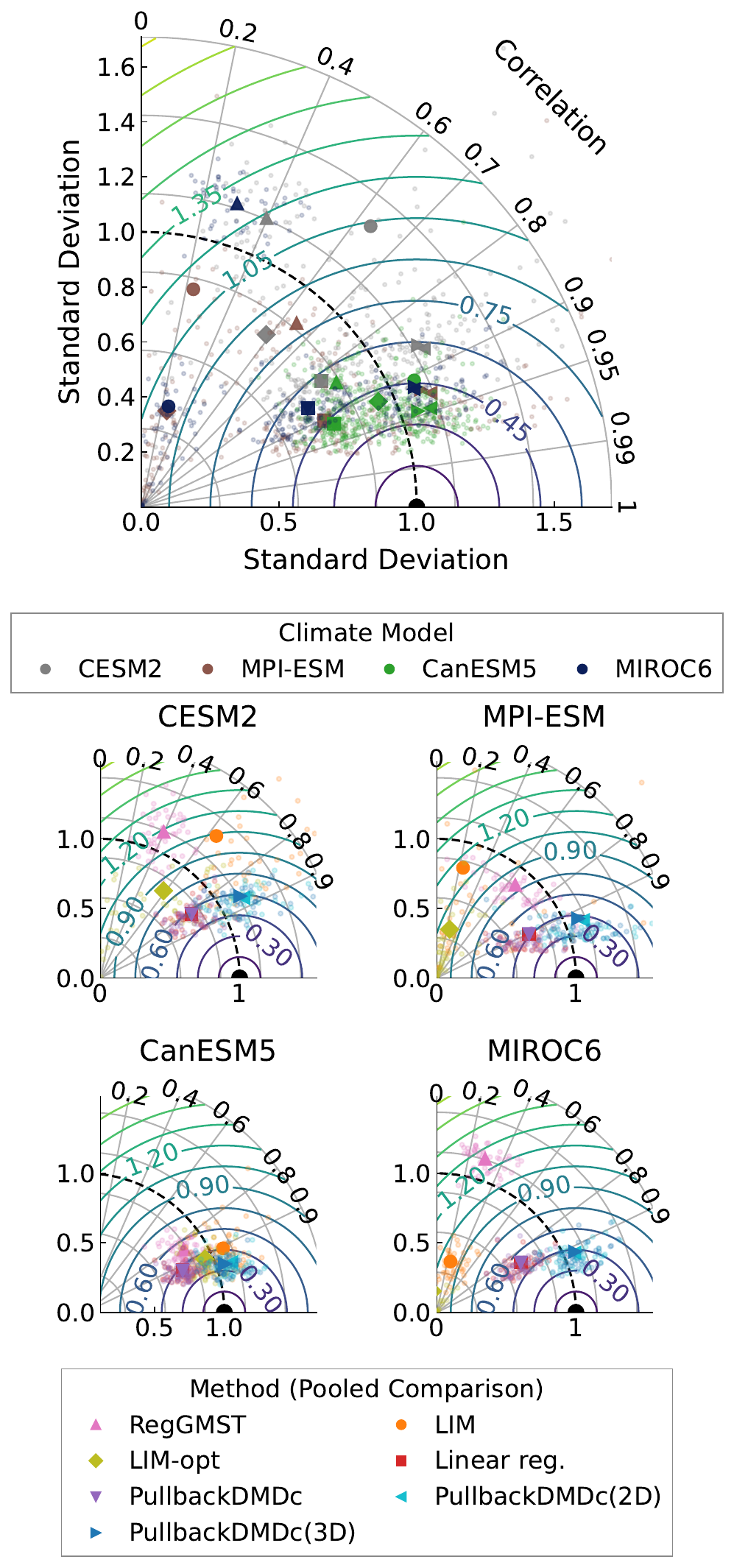}
    \end{subfigure}
    \hfill
    \begin{subfigure}[b]{0.3\linewidth}
        \centering
        \includegraphics[width=\textwidth]{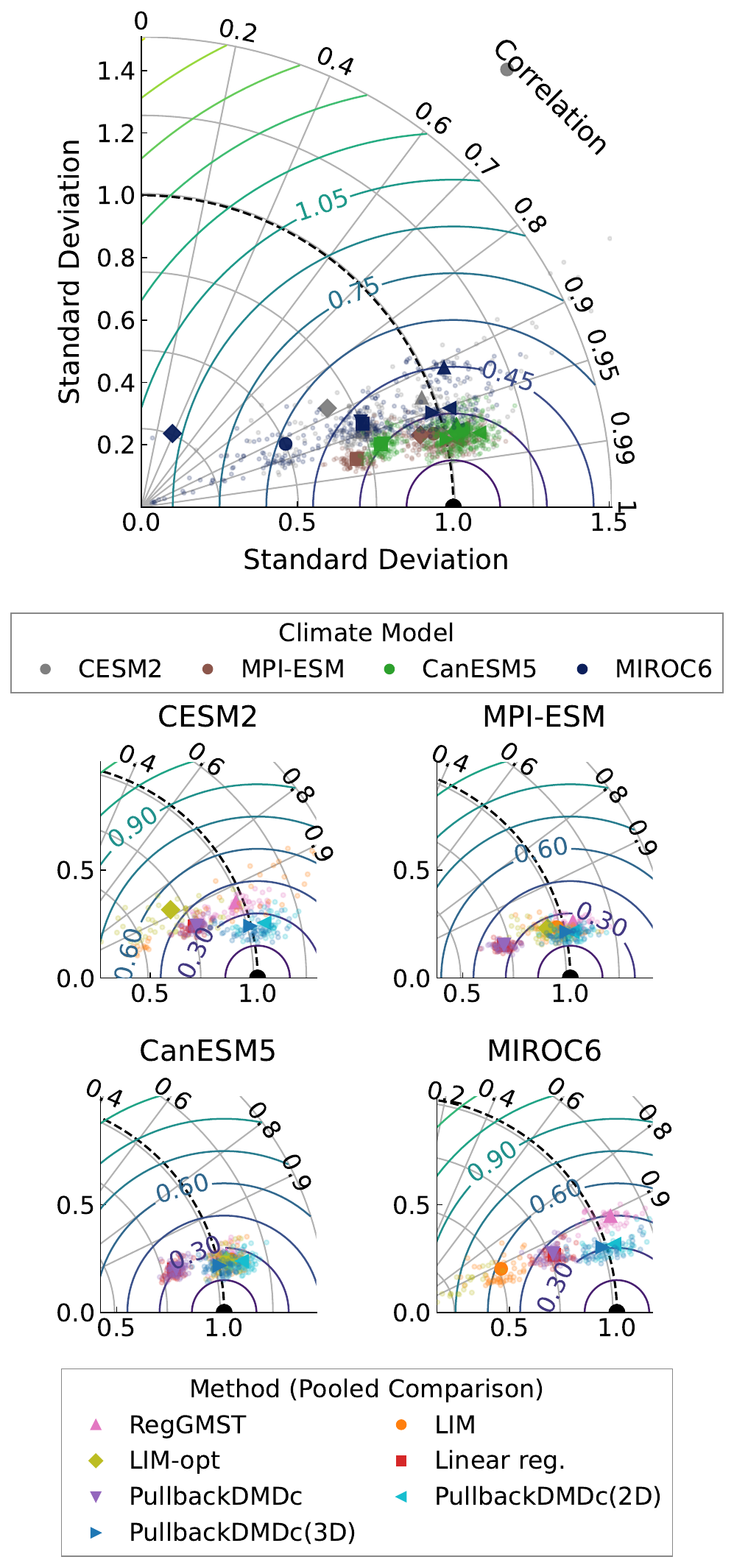}
    \end{subfigure}
    \hfill
    \begin{subfigure}[b]{0.3\linewidth}
        \centering
        \includegraphics[width=\textwidth]{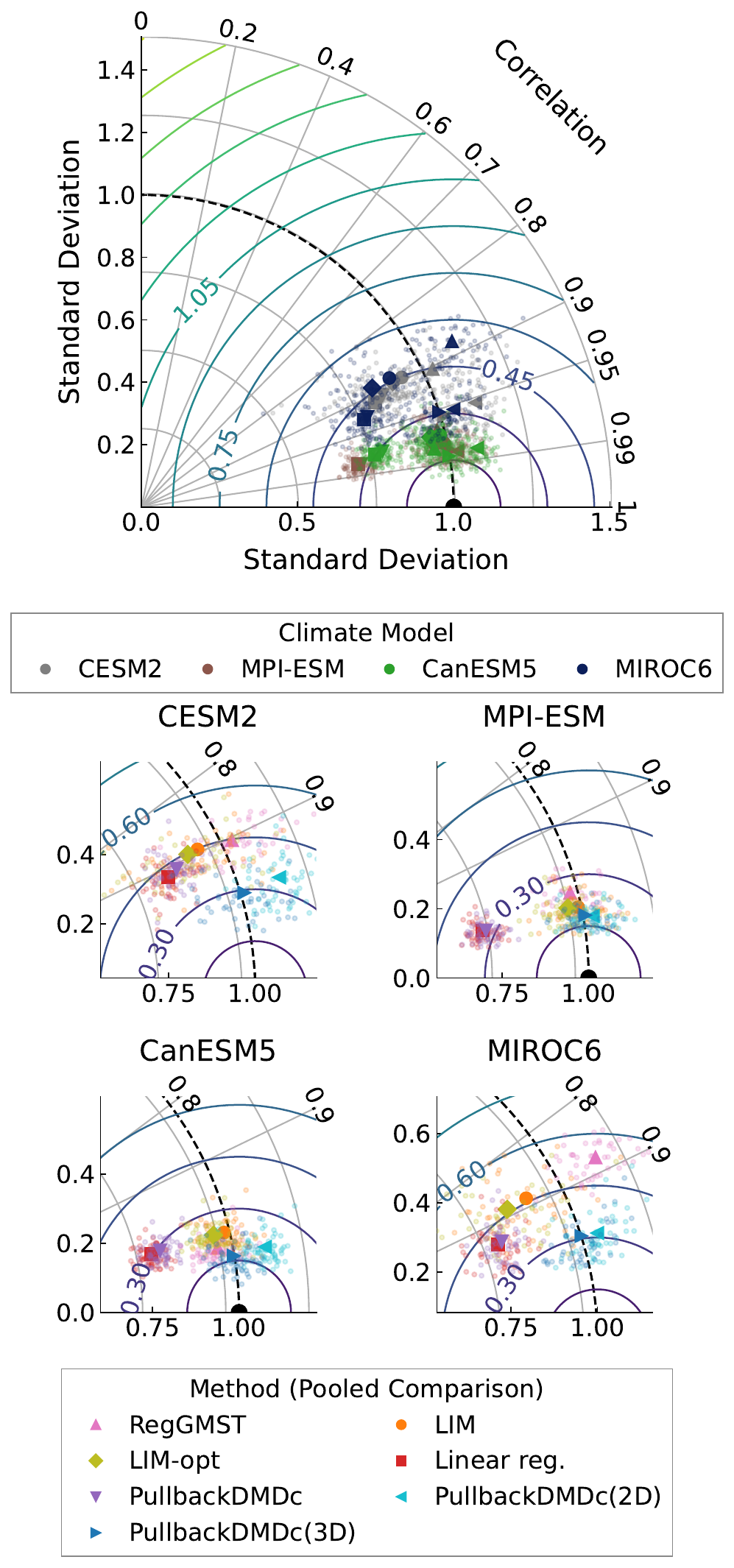}
    \end{subfigure}
    \caption{Same as in Fig. 2 (main text) but for comparing forced-response trend estimates on the $1850-2014$ time interval for SLP (left), SAT (middle), OSAT (right). \added{All statistics are uncentered and area-weighted, as in Fig.~4.}
    \methodname(3D) consistently achieves the lowest RMSD and highest correlation for SAT and OSAT, while \methodname(2D) and \methodname(3D) perform similarly for SLP. Across all methods and variables, the forced response is easiest to estimate in CanESM5 and most difficult in MIROC6, with OSAT being the most predictable variable and SLP the most challenging.}
    \label{fig:trend_taylors}
\end{figure}

\begin{figure}[H]
    \centering
    \includegraphics[width=\linewidth]{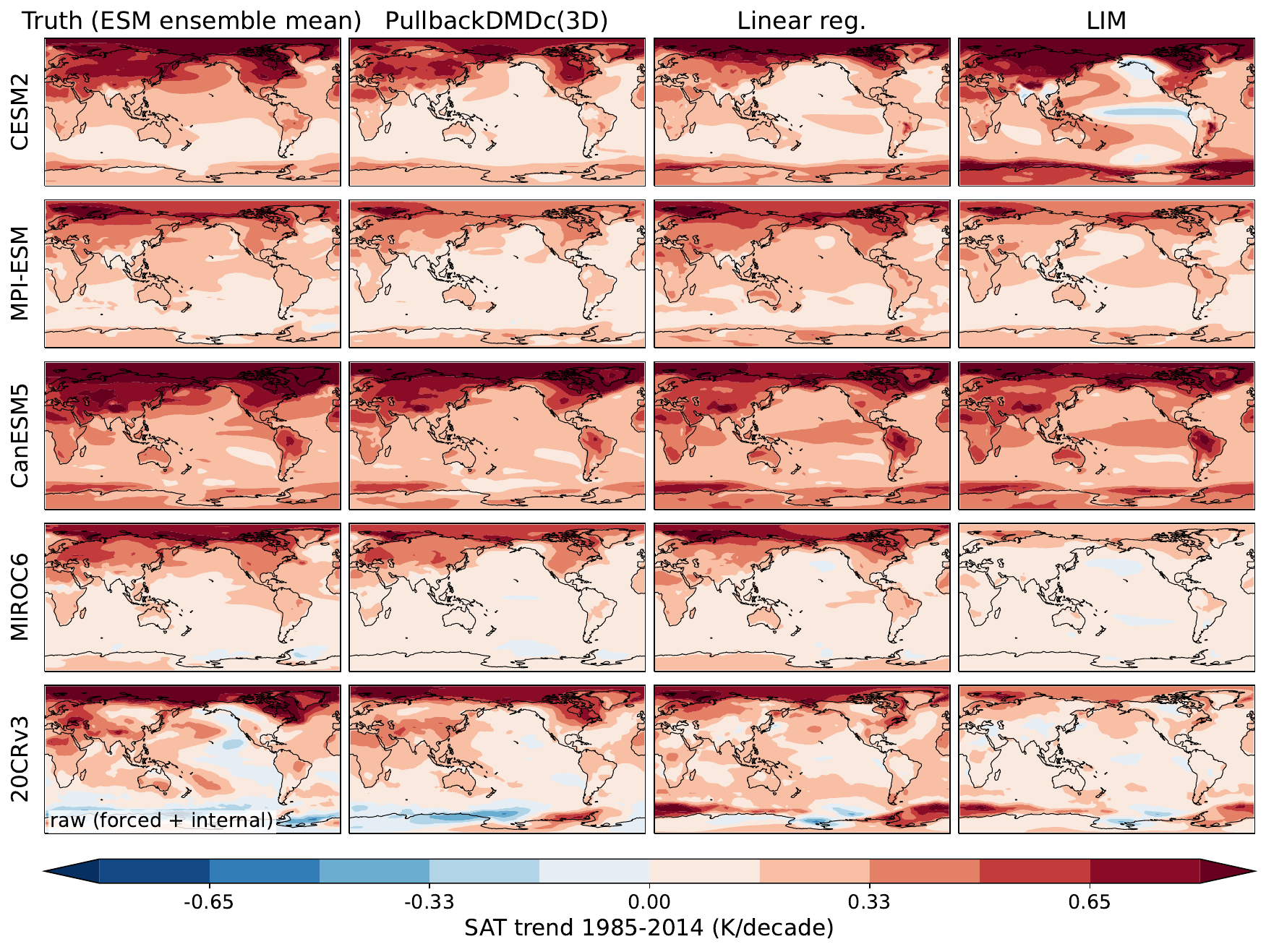}
    \caption{\added{As Fig.~4, but for SAT (K/decade). The forced trends of the ESMs show Arctic amplification and a land--ocean contrast, which \methodname(3D) and LR reproduce (dominated by the shared warming), while LIM fails for CESM2. For 20CRv3, the estimates differ near $60^\circ$S, where \methodname(3D) retains a cooling and LR and LIM give a warming.}}
    \label{fig:trend-map-tas}
\end{figure}

\begin{figure}[H]
    \centering
    \includegraphics[width=\linewidth]{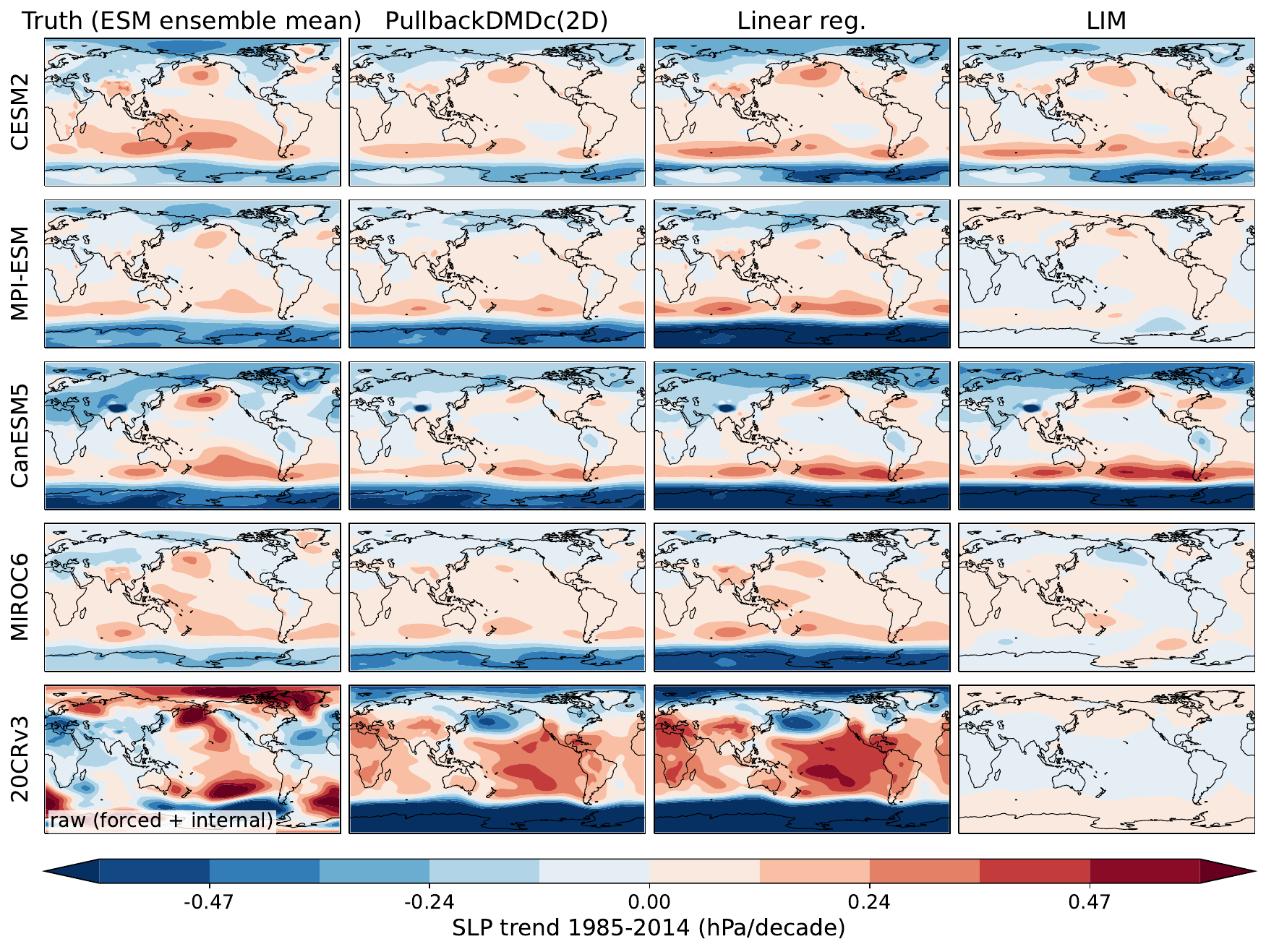}
    \caption{\added{As Fig.~4, but for SLP (hPa/decade), with \methodname(2D). The area-mean trend is close to zero, so the uncentered scores are close to centered pattern statistics. The forced trends of the ESMs project onto the positive phase of the Southern Annular Mode. LR overestimates their amplitude, especially over Antarctica, so \methodname(2D) has the smallest nRMSE in all ESMs (equal to LR for CanESM5), while LIM fails for MPI-ESM and MIROC6. For 20CRv3, \methodname(2D) and LR give nearly identical and much stronger trends than in the ESMs, possibly reflecting the SLP change in the early, weakly constrained part of the reanalysis.}}
    \label{fig:trend-map-psl}
\end{figure}

\begin{figure}[H]
    \centering
    \begin{subfigure}[b]{0.3\linewidth}
        \centering
        \includegraphics[width=\textwidth]{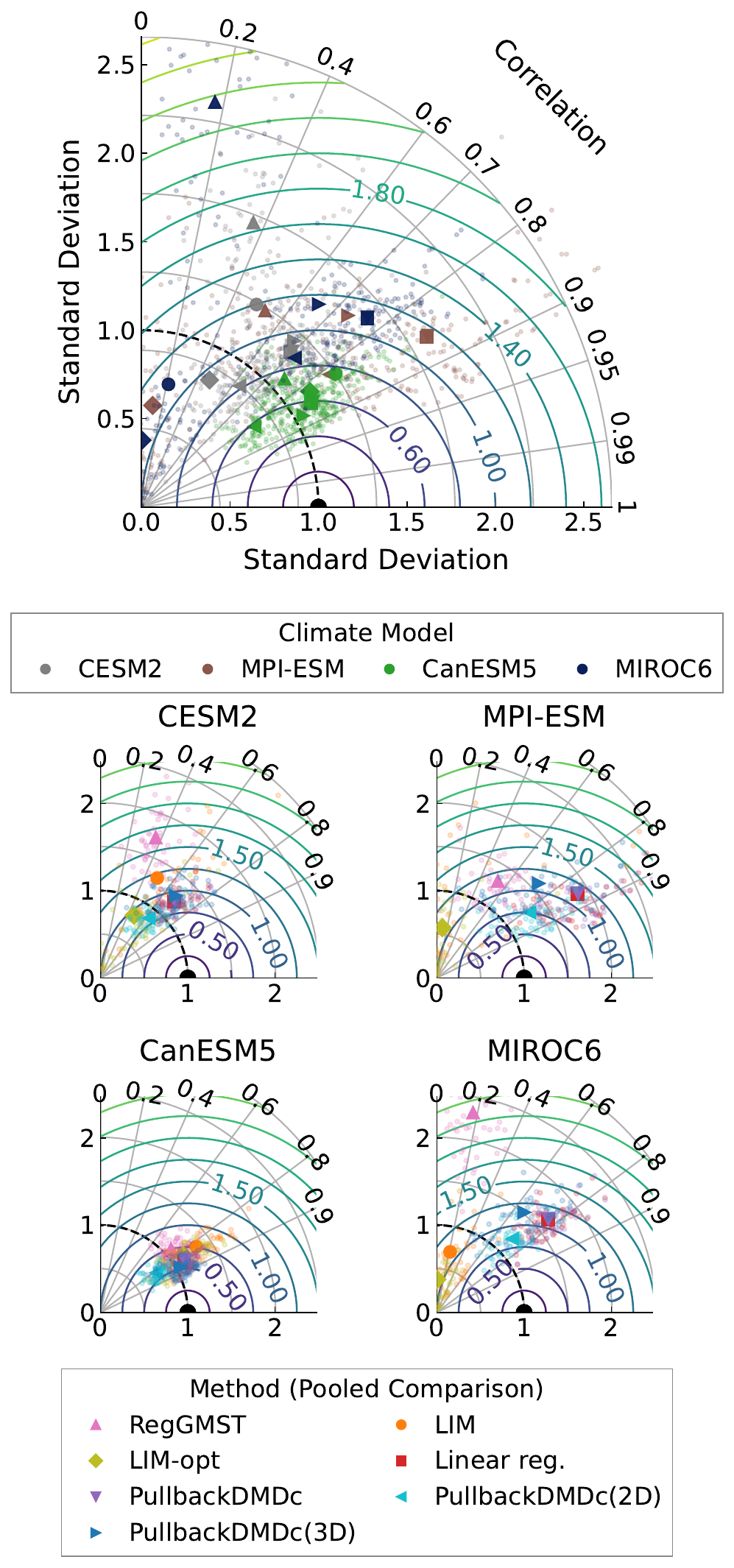}
    \end{subfigure}
    \hfill
    \begin{subfigure}[b]{0.3\linewidth}
        \centering
        \includegraphics[width=\textwidth]{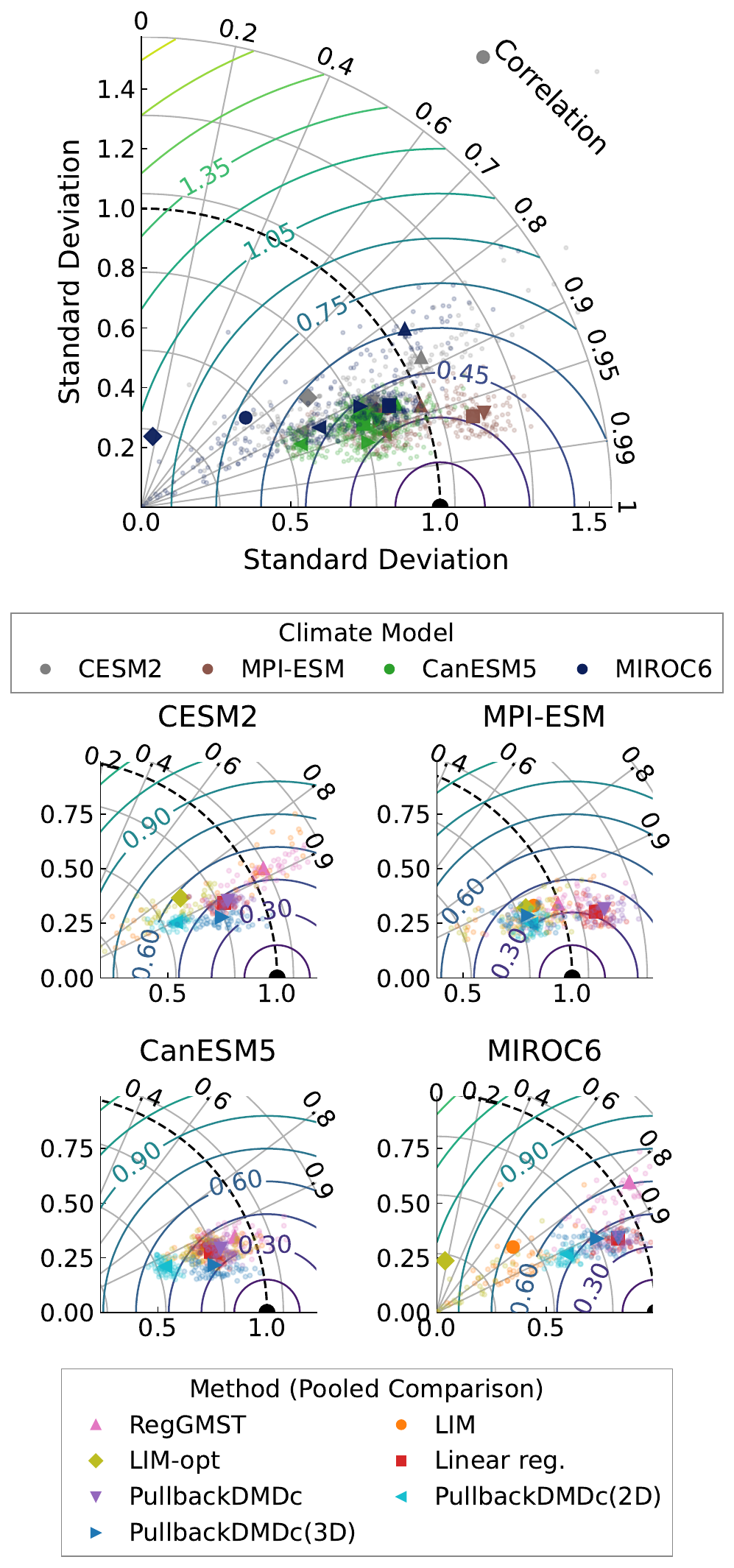}
    \end{subfigure}
    \hfill
    \begin{subfigure}[b]{0.3\linewidth}
        \centering
        \includegraphics[width=\textwidth]{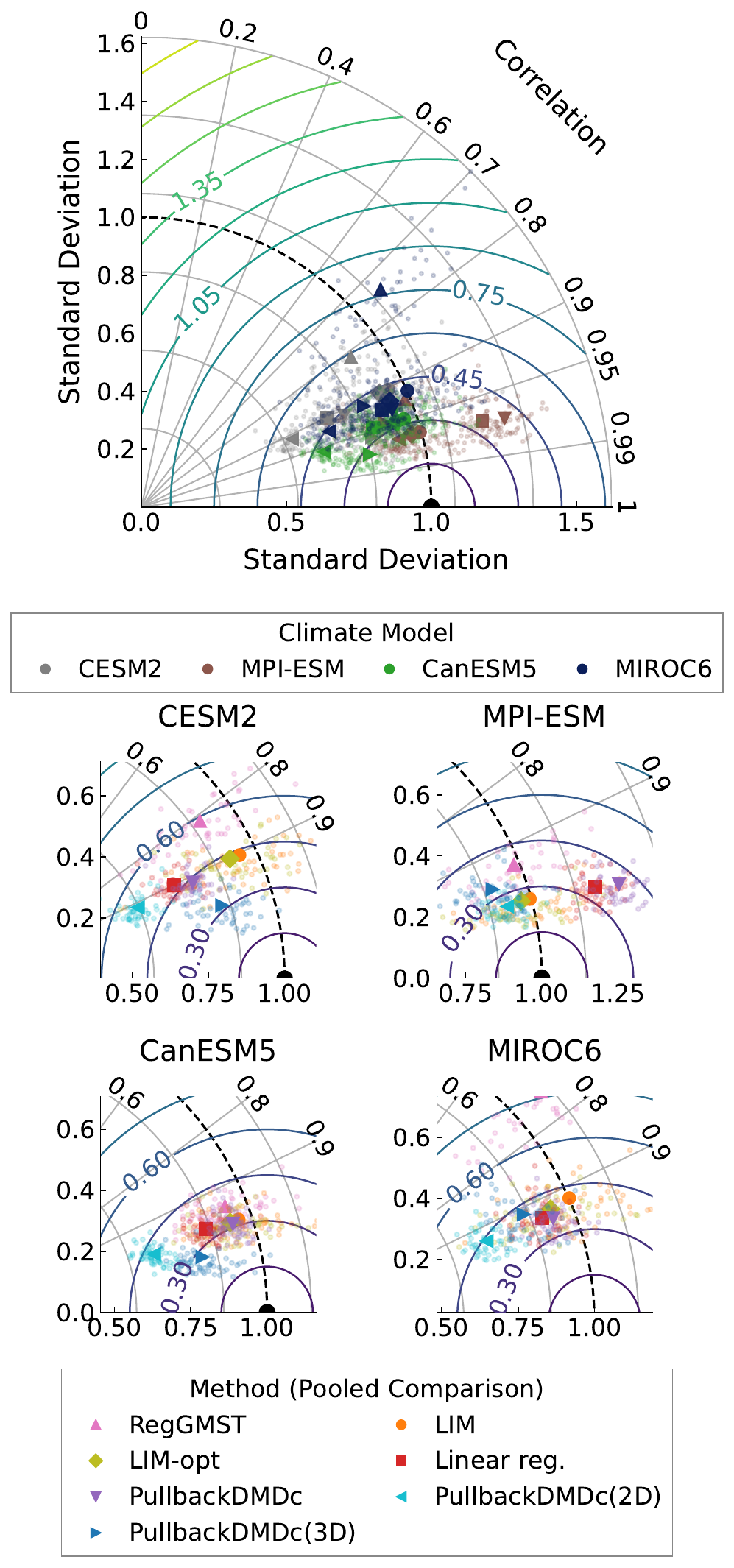}
    \end{subfigure}

    \caption{\added{As Fig.~\ref{fig:trend_taylors}, but for the linear trends of the forced-response estimates over $1985$--$2014$ for (left) SLP (middle) SAT and (right) OSAT. All statistics are uncentered, as in Fig.~4: the distance from the origin is the RMS amplitude of the trend map, and the distance from the reference point is the RMS error. Over this short window, where the forced response is close to linear, \methodname(3D) performs similarly to LR for SAT and OSAT and underestimates the trend amplitude for MPI-ESM and MIROC6, so its advantage over the full record (Fig.~\ref{fig:trend_taylors}) largely disappears. For SLP, LR overestimates the amplitude, and \methodname(2D) has the smallest or equal-smallest RMS error in all ESMs.}}
    \label{fig:classic_taylor_trend_window}
\end{figure}

\begin{figure}[H]
    \centering
    \begin{subfigure}[b]{0.3\linewidth}
        \centering
        \includegraphics[width=\textwidth]{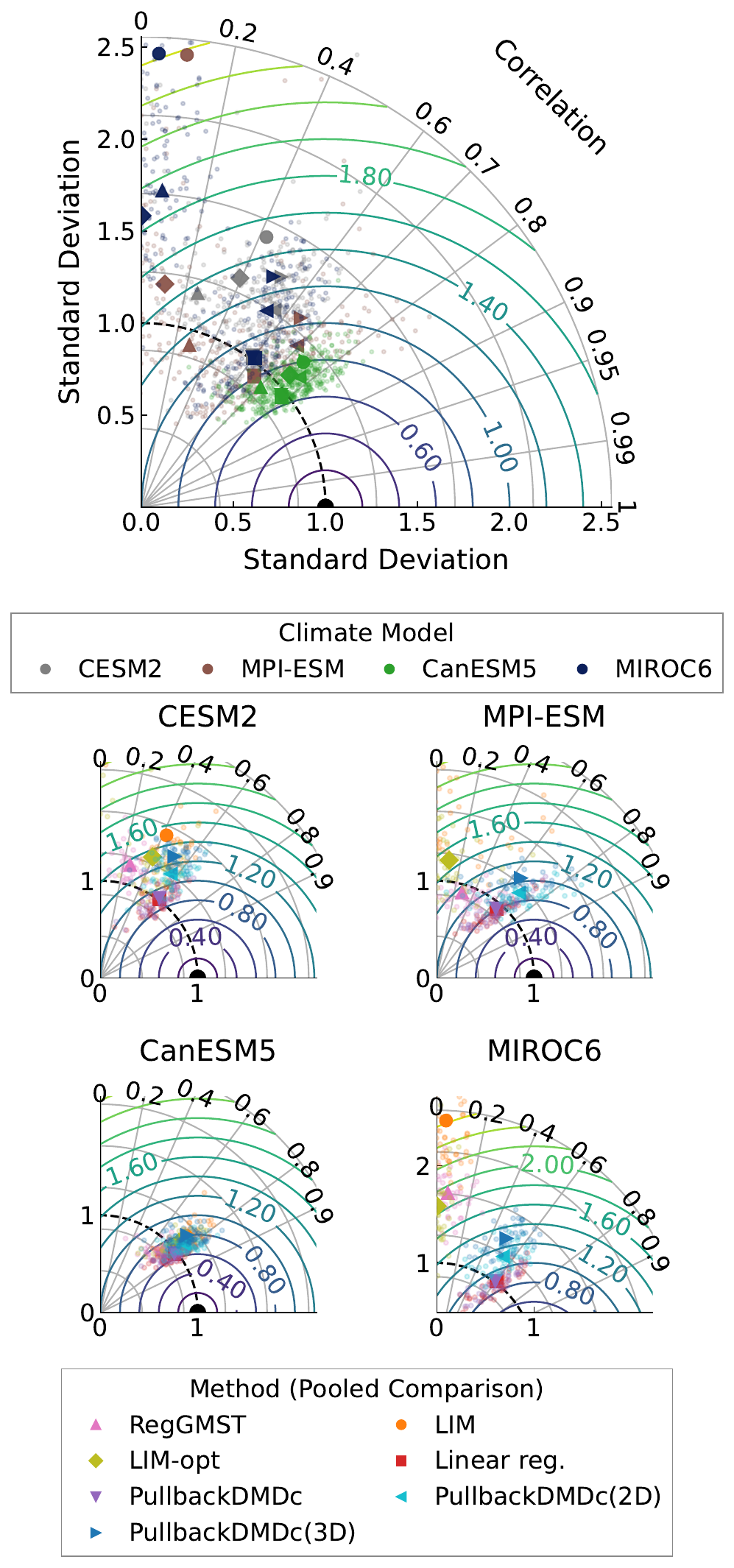}
    \end{subfigure}
    \hfill
    \begin{subfigure}[b]{0.3\linewidth}
        \centering
        \includegraphics[width=\textwidth]{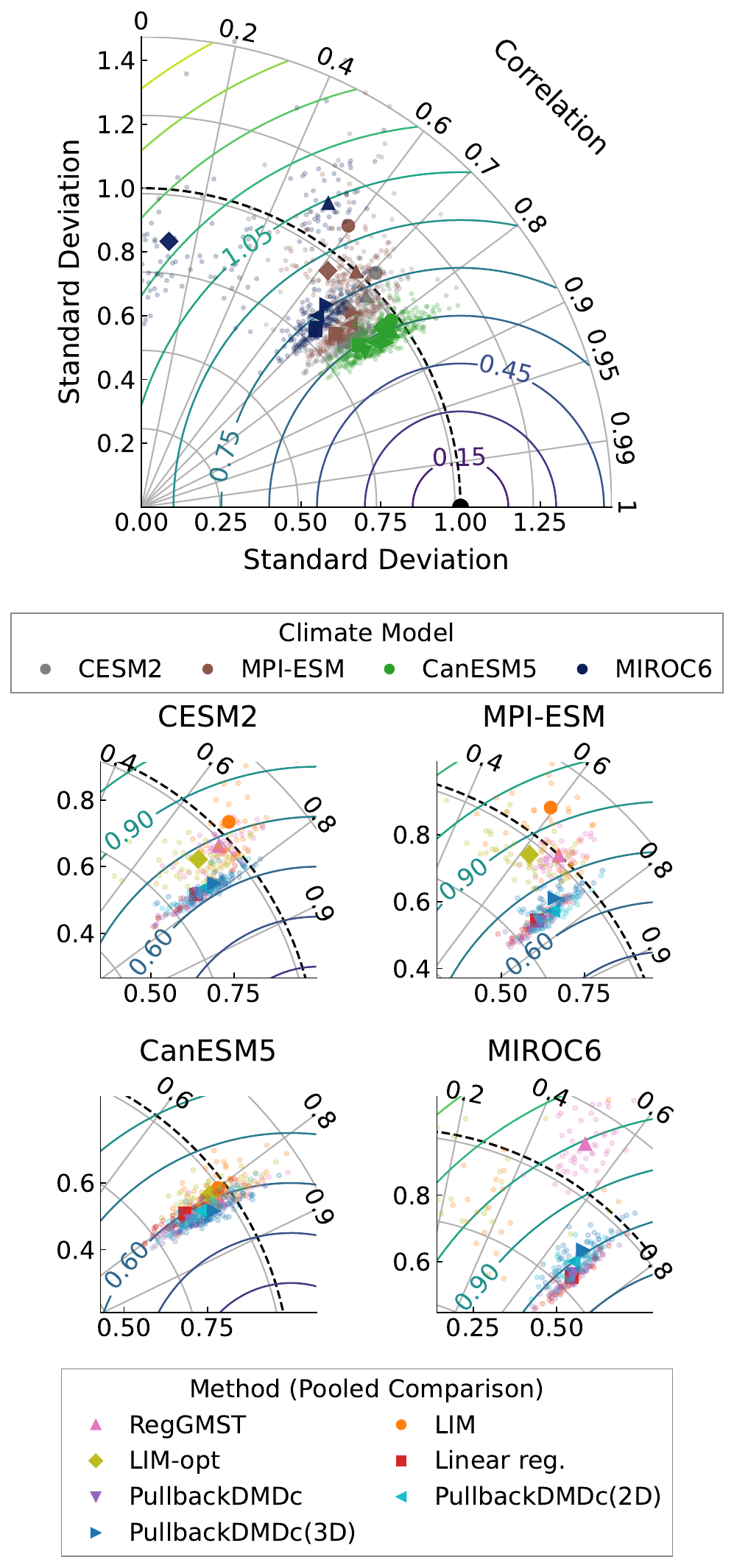}
    \end{subfigure}
    \hfill
    \begin{subfigure}[b]{0.3\linewidth}
        \centering
        \includegraphics[width=\textwidth]{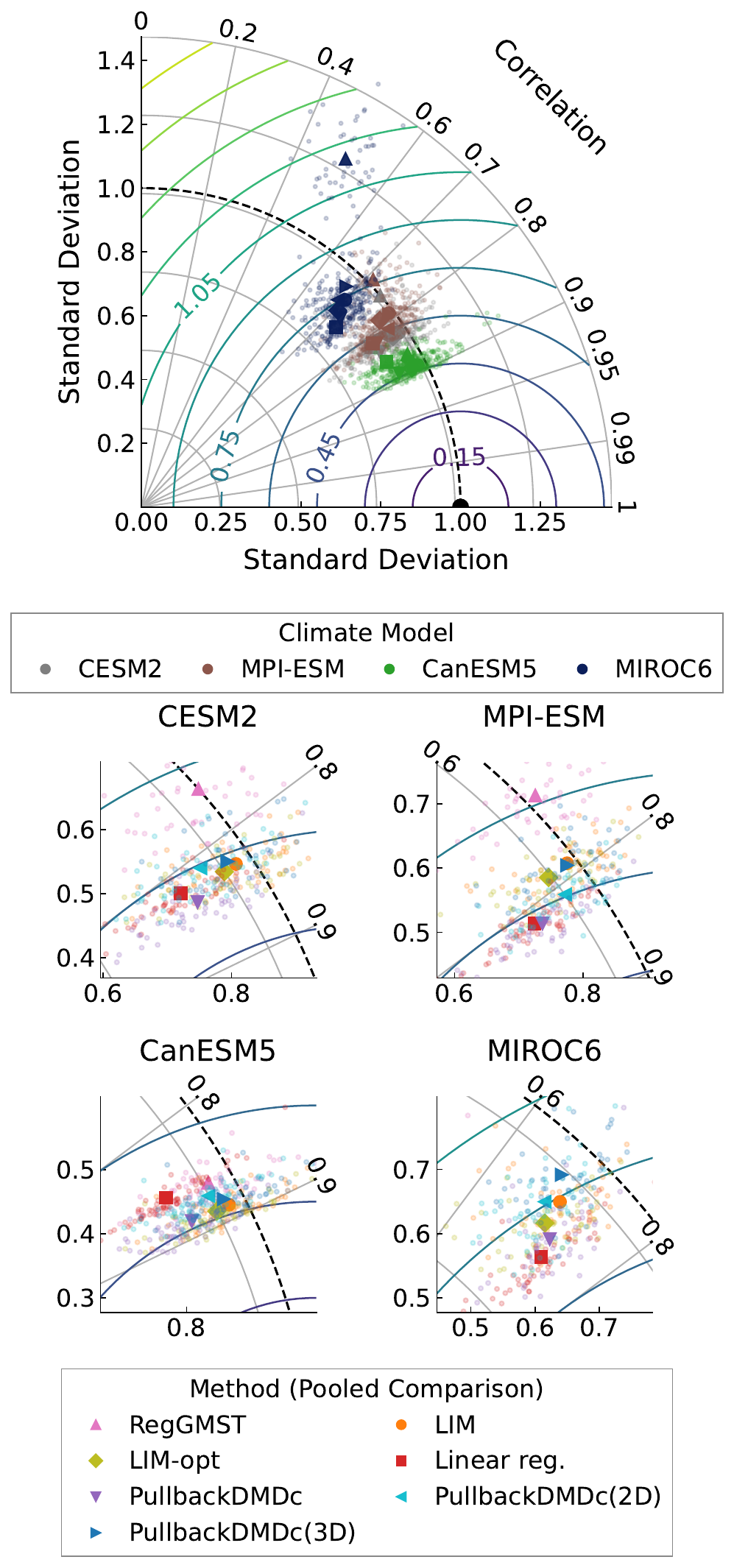}
    \end{subfigure}
    \caption{Same as in Fig. 2 (main text) but for SLP (left), SAT (middle), OSAT (right) on a short interval ($1950-2014$).
    For this period, relative model predictability mirrors the longer interval, with CanESM5 most and MIROC6 least predictable. Forced-response performance degrades overall, and baselines become more competitive with Linear Regression and \methodname\ variants. As in the longer interval, Linear Regression and at least one \methodname\ variant remain the most skillful methods, and, unlike the longer period, the $1$-dimensional forcing predictors (Linear Regression, \methodname) often achieve the lowest RMSD overall, though they underestimate the forced-response scale for OSAT and SAT while accurately capturing it for SLP.}
    \label{fig:classic_taylor_short}
\end{figure}

\begin{figure}[H]
    \centering
    \begin{subfigure}{.9\linewidth}
        \centering
        \includegraphics[width=\linewidth, trim={0 1.9cm 0 .26cm}, clip]{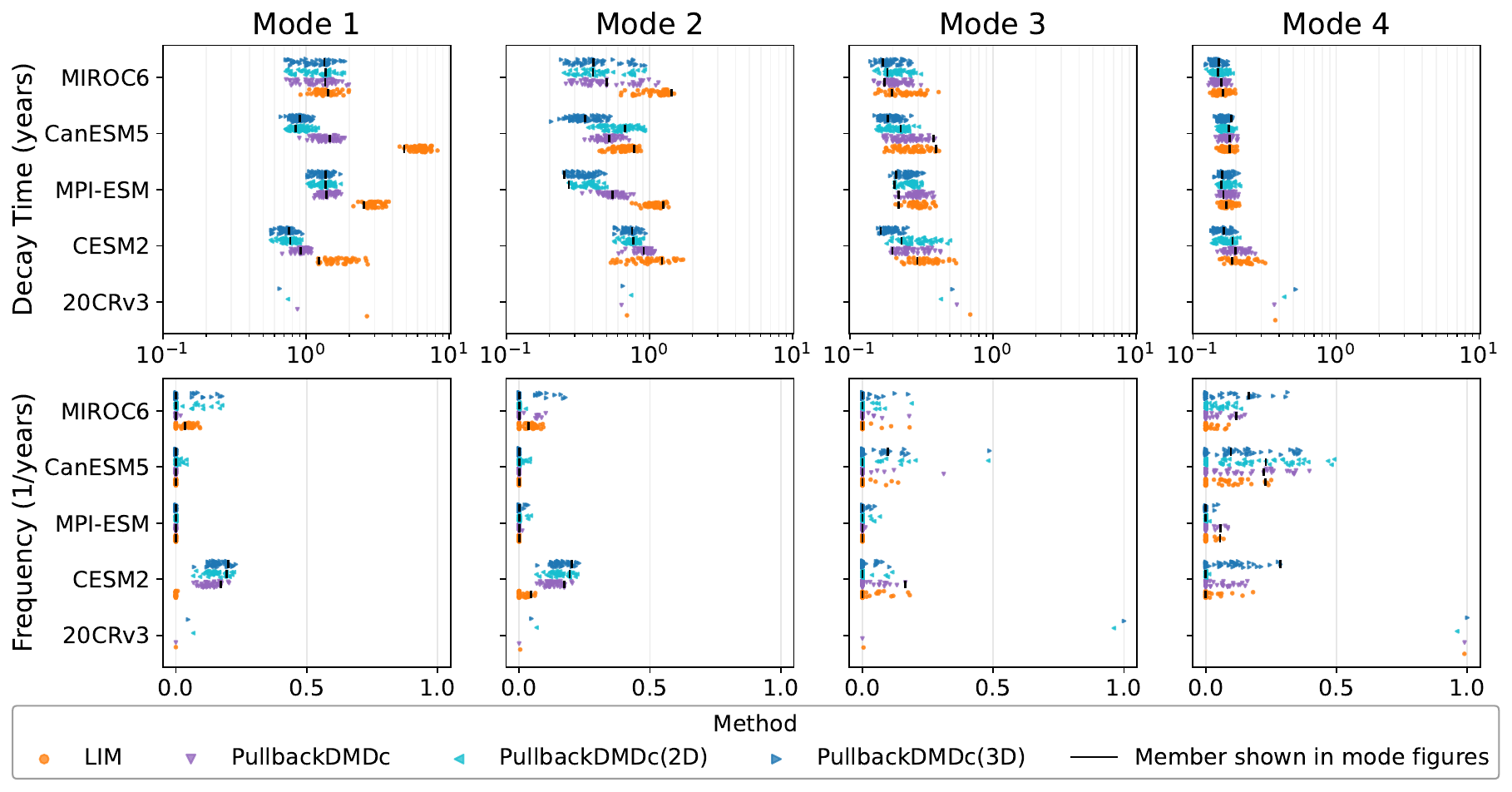}
        \label{fig:eigs_tas_ocean}
    \end{subfigure}

    \begin{subfigure}{.9\linewidth}
        \centering
        \includegraphics[width=\linewidth, trim={0 .2cm 0 .26cm}, clip]{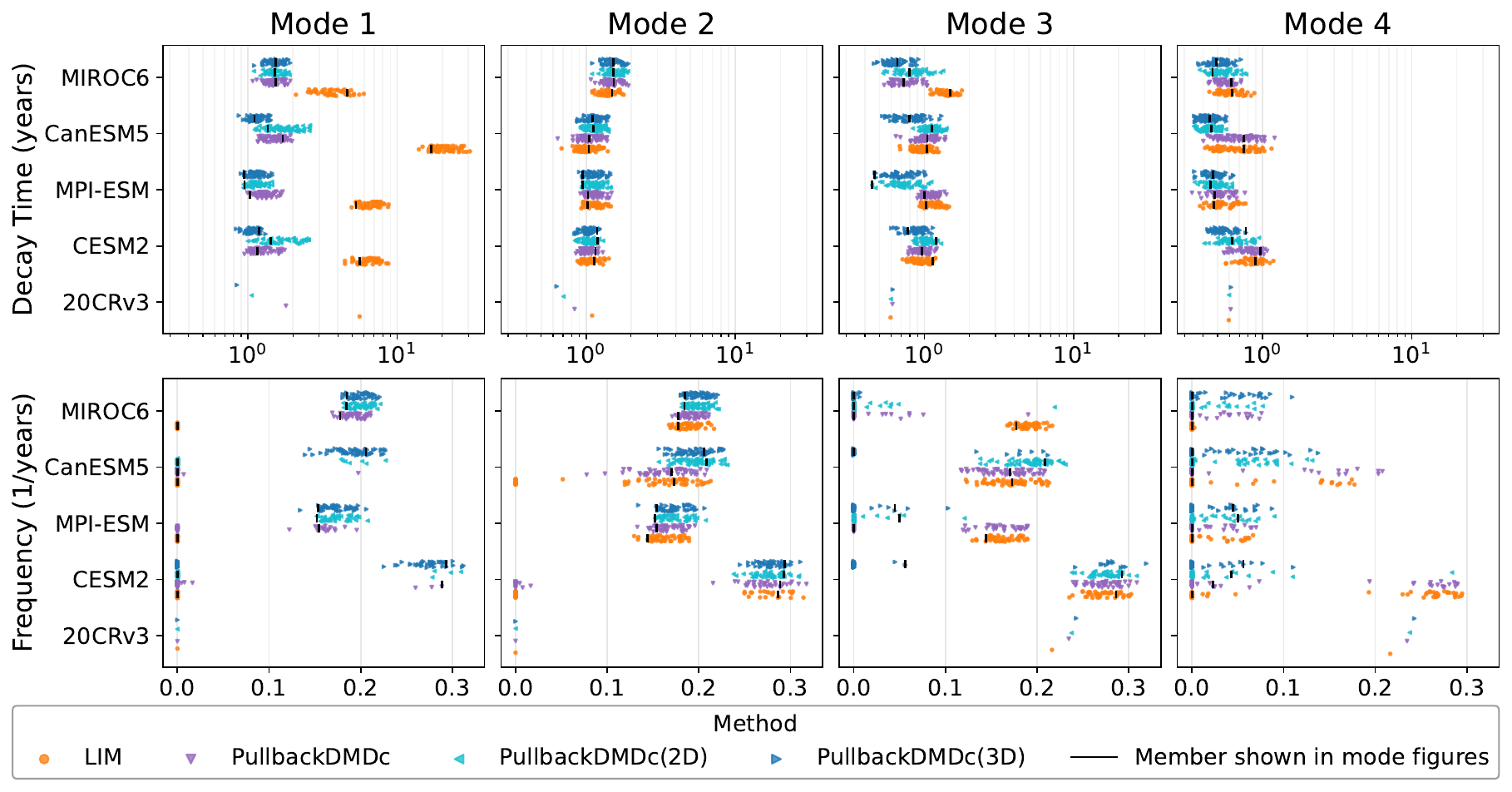}
        \label{fig:eigs_tas}
    \end{subfigure}
    \caption{\added{Comparison of LIM and~\methodname~decay time (top) and oscillation frequency (bottom) for the top four modes for SAT (top) and OSAT (bottom). Each column represents one mode (ordered by eigenvalue decay time).} For each mode and model, scatter points represent individual ensemble members. Outliers are removed using an interquartile-range filter ($\mathrm{IQR}\times1.5$ for decay time and $\mathrm{IQR}\times1.0$ for frequency). The decay time of the first LIM mode is longer than that of the~\methodname~variants in both variables. \methodname~exhibits the oscillatory leading modes 1-2 when LIM does not (except for some MIROC6 and CESM2 members in SAT). \added{LIM has no control input, so any slowly varying forced signal is represented by the least damped mode of the evolution operator, with a long decay time to accommodate the trend. In~\methodname, forcing enters through the control. This allows $\A$ to describe internal dynamics alone, leading to shorter leading decay times and the emergence of oscillatory leading modes. In other words, the LIM operator conflates internal and forced dynamics, whereas the~\methodname~operator $\A$ better approximates the internal evolution operator.}
    }
    \label{fig:eigs_supp}
\end{figure}

\begin{figure}[H]
    \centering
    \begin{subfigure}{0.7\linewidth}
        \centering
        \includegraphics[width=\linewidth]{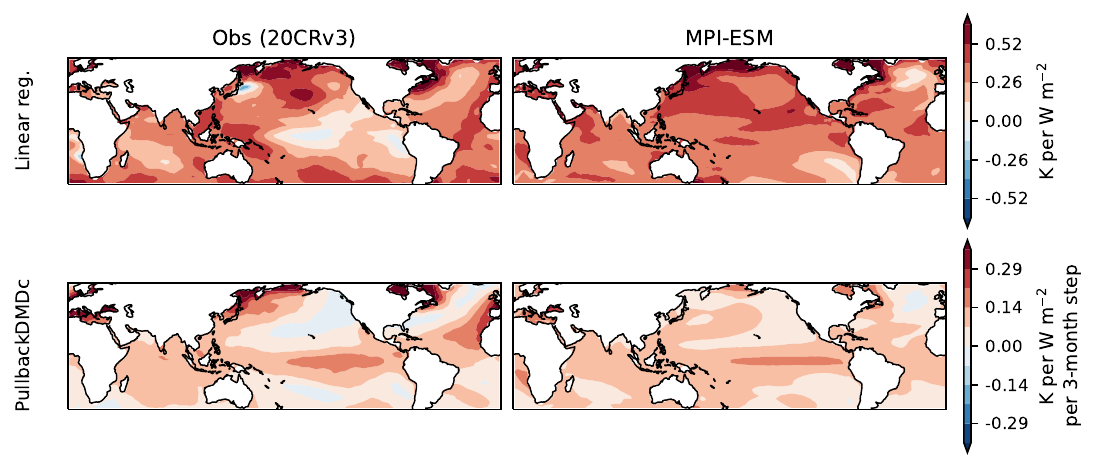}
    \end{subfigure}
    \begin{subfigure}{0.7\linewidth}
        \centering
        \includegraphics[width=\linewidth]{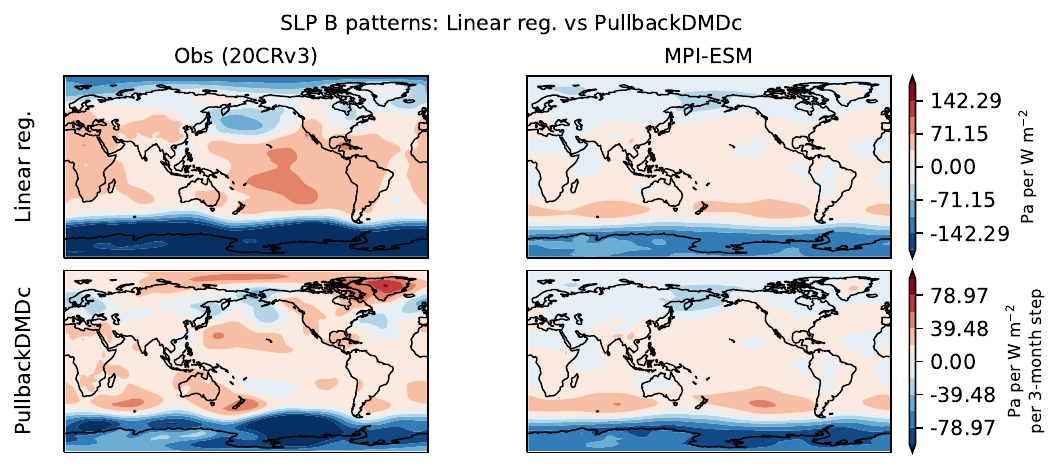}
    \end{subfigure}
    \caption{Spatial patterns of the LR and~\methodname~$\boldsymbol{B}$ matrix for (top) ocean surface air temperature (OSAT) and (bottom) sea level pressure (SLP). These patterns are the leading forcing footprint derived from observations (left) and the ensemble-mean footprint from MPI-ESM (right). Amplitudes are not directly comparable across panels. Compared to LR, \methodname~produces a weaker forcing signal with a distinct spatial structure, with differences that are more pronounced for SLP than OSAT. \added{Importantly, the two $\boldsymbol{B}$ matrices are different objects and should be read differently. In LR, $\boldsymbol{B}$ maps forcing directly onto the state, so it is the full forced-response fingerprint: it integrates all feedbacks and slow adjustments, and accordingly recovers the familiar greenhouse-gas signature (i.e., near-uniform warming and low pressure over both polar regions). In~\methodname, the full forced response is estimated by propagating this input through the internal dynamics $\A$ over a long spin-up interval. Thus, we do not expect the~\methodname~$\boldsymbol{B}$ to resemble the LR fingerprint, and its weaker amplitude reflects that it represents a single step of a cumulative response. We include this figure to verify that the two methods estimate distinct input operators, rather than to compare forced-response patterns directly}}
    \label{fig:B}
\end{figure}

\begin{figure}[H]
    \centering
    \includegraphics[width=\textwidth]{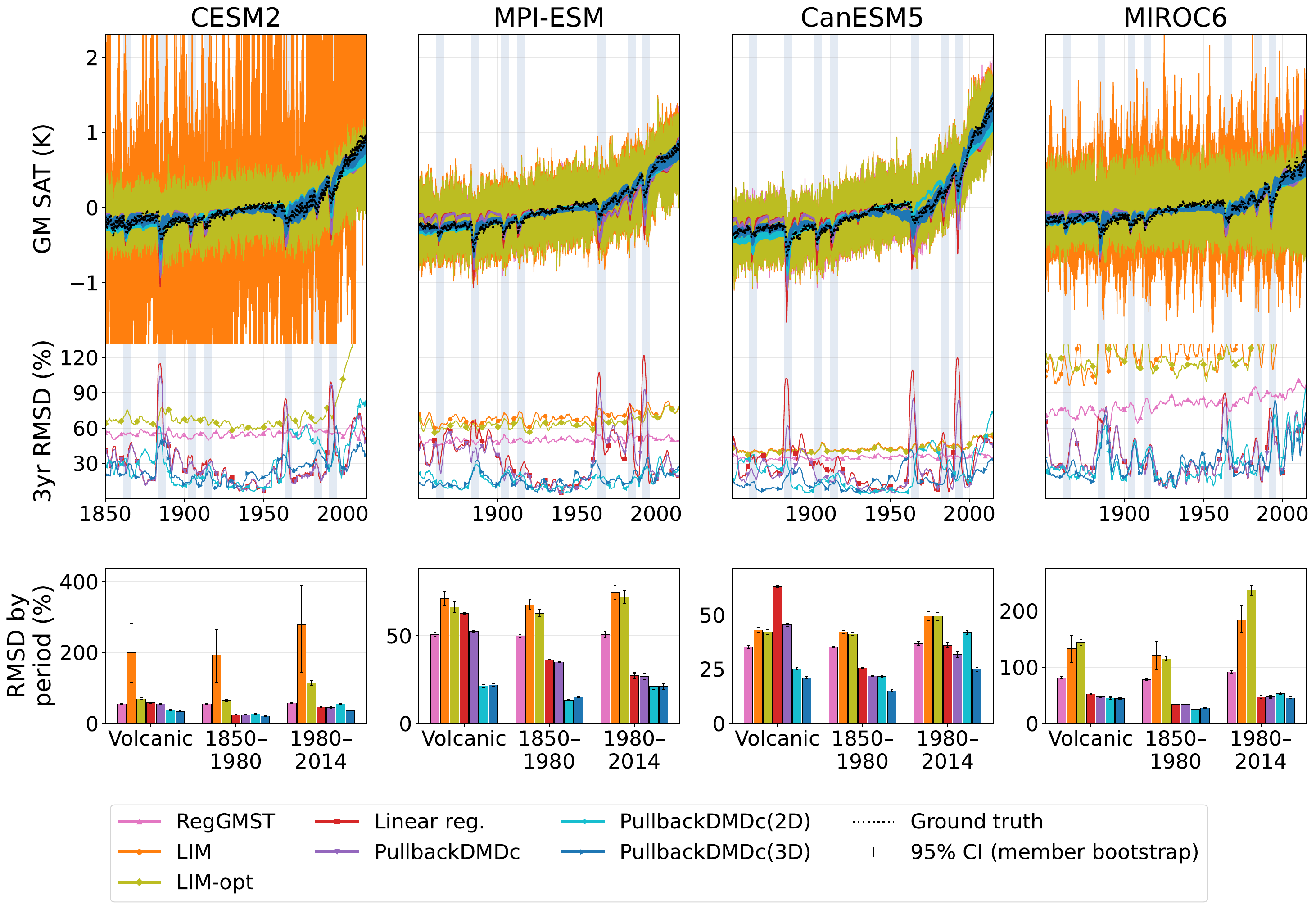}
    \caption{Same as in Fig. 3 (main text) but for SAT. 
    As with OSAT, we see that LIM baselines exhibit high variance in global mean time series across ensemble members. Additionally, these baselines, along with RegGMST, are less skillful than Linear Regression and~\methodname~variants. However, the bottom two rows reveal a substantial degradation in forced-response estimation skill for~\methodname~variants and Linear Regression during volcanic intervals. On these intervals, we find that methods with a higher-dimensional forcing signal like~\methodname(2/3D) have lower errors.  Overall,~\methodname(3D) maintains relatively consistent skill as the best method in all $3$ rows.}
    \label{fig:gm_tas}
\end{figure}

\begin{figure}[H]
    \centering
    \includegraphics[width=\textwidth]{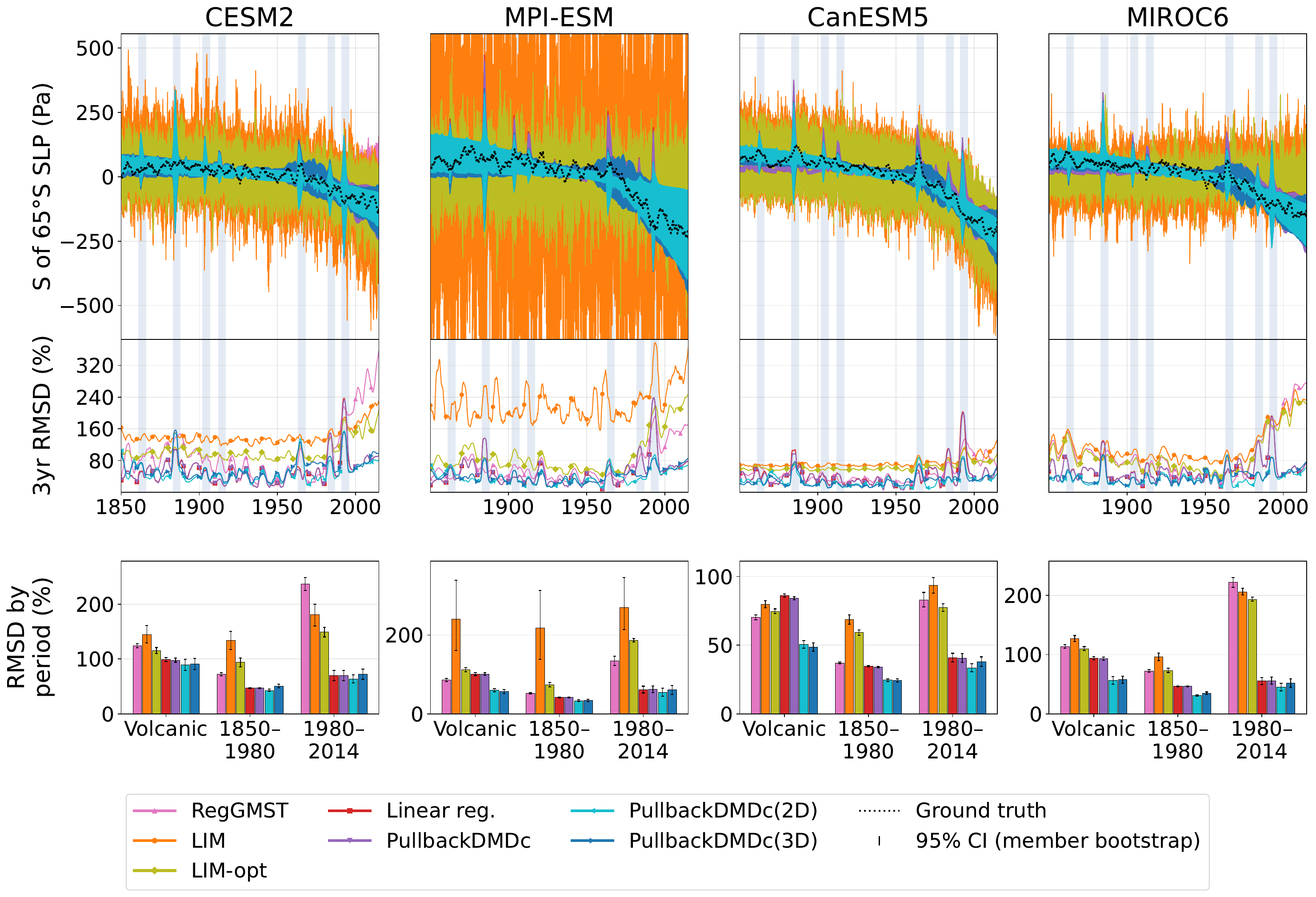}
    \caption{Same as Fig.~\ref{fig:gm_tas} but for sea level pressure (SLP) \added{averaged south of $65^\circ$S, used instead of the global mean, which is nearly constant because the mass of the atmosphere is conserved. The ensemble mean decreases mainly after the 1960s, a trend towards the positive phase of the Southern Annular Mode. \methodname(2D) and \methodname(3D) have the smallest errors in most ESMs and periods, especially in volcanic intervals, where LR and the one-dimensional \methodname~overshoot. RegGMST, LIM and LIM-opt have errors exceeding the forced signal after 1980 in CESM2, MPI-ESM and MIROC6.}}
    \label{fig:gm_psl}
\end{figure}

\begin{figure}[H]
    \centering
    \includegraphics[width=\textwidth]{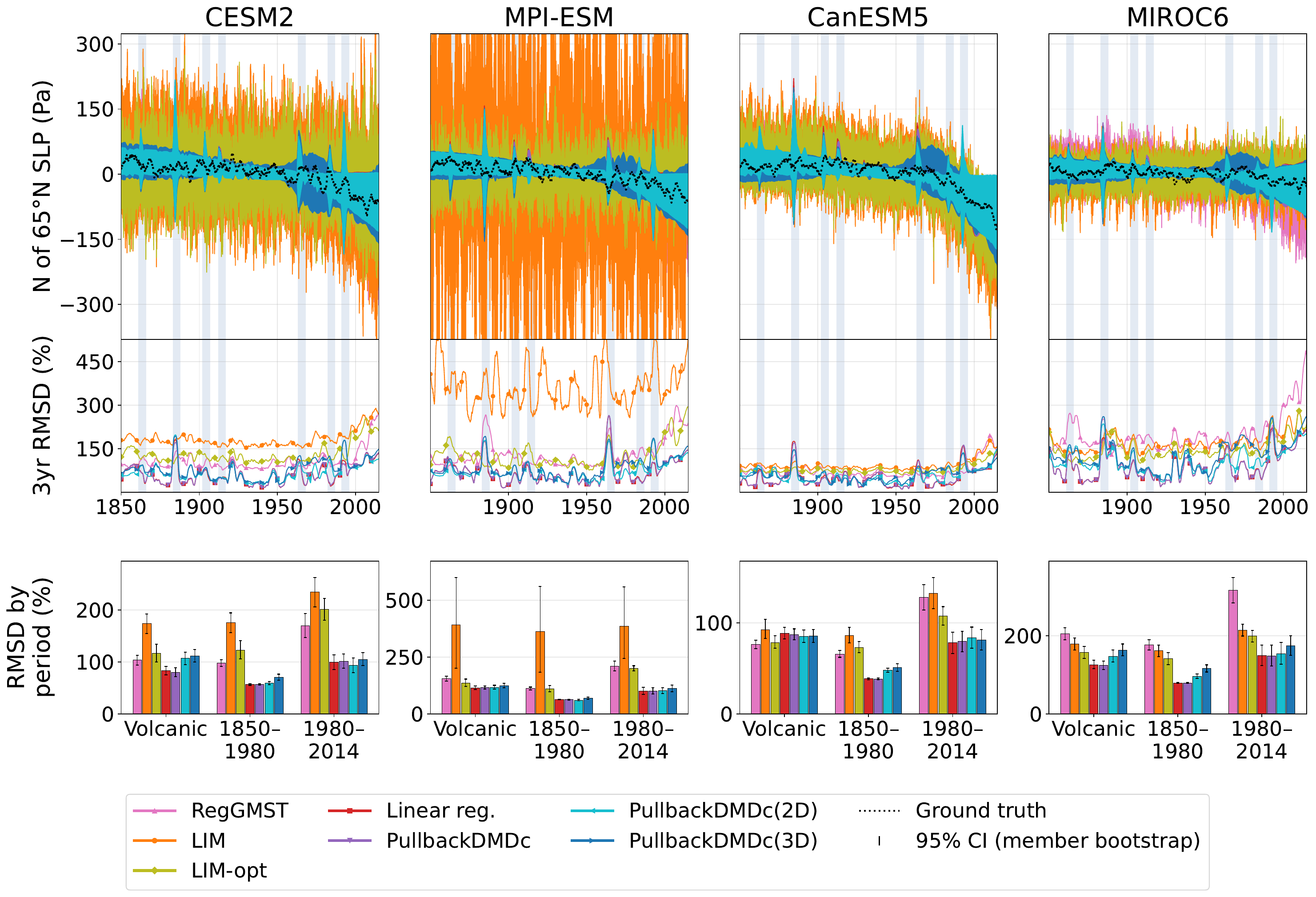}
    \caption{\added{Same as Fig.~\ref{fig:gm_tas} but for sea level pressure (SLP) averaged north of $65^\circ$N. The forced signal is weak compared with internal variability, with a slight decrease after about $1980$ in the ensemble means, and all methods have errors comparable to or larger than their standard deviation, especially after $1980$.}}
    \label{fig:gm_psl_n65}
\end{figure}

\begin{figure}[H]
    \centering
    \includegraphics[width=\textwidth]{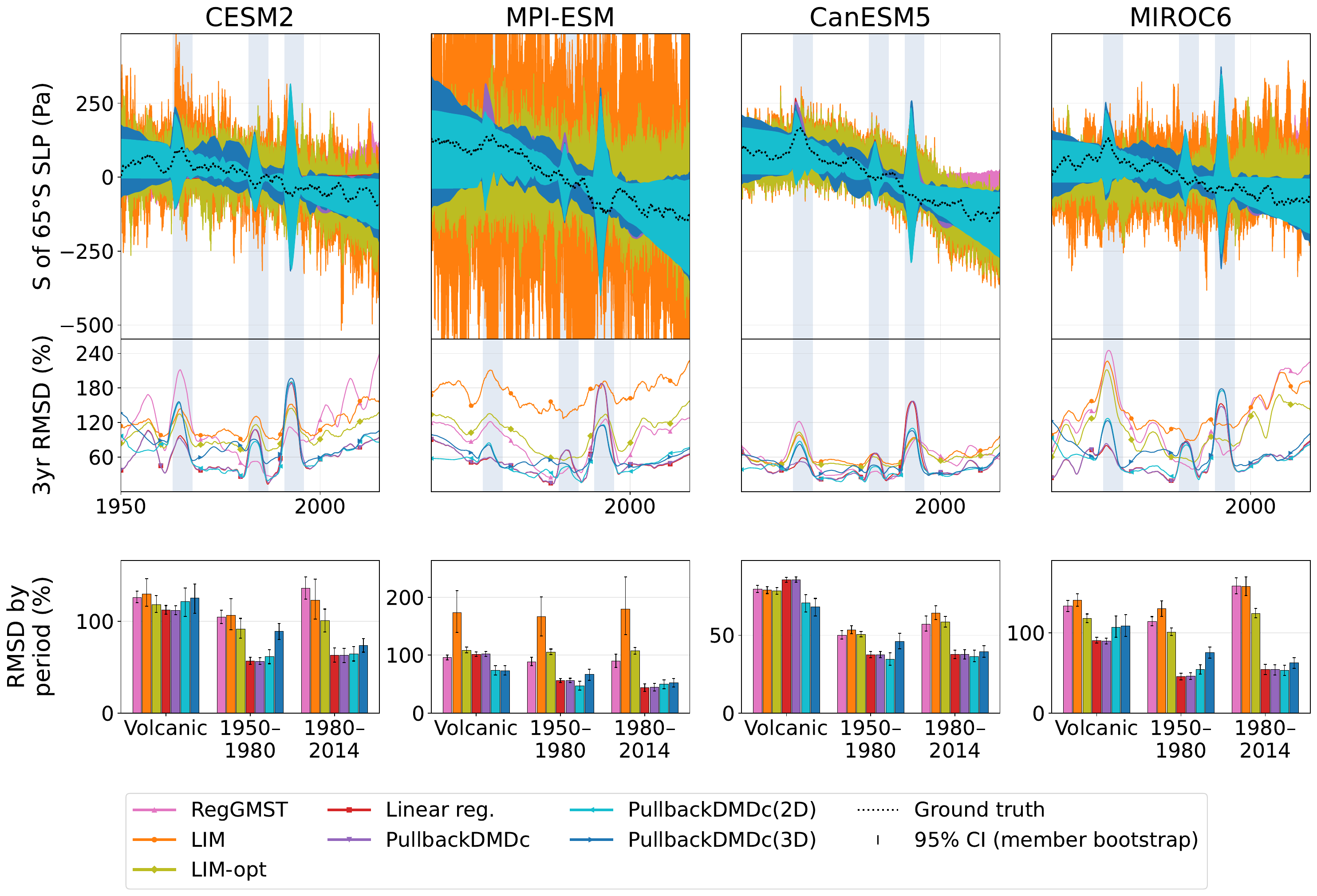}
    \caption{Same as in Fig.~\ref{fig:gm_tas}, but for sea level pressure (SLP) \added{averaged south of $65^\circ$S, as in Fig.~\ref{fig:gm_psl},} on a short interval ($1950-2014$).
    Since this analysis covers a shorter period, there are only three volcanic intervals. The overall qualitative behavior is similar to the longer record, with amplified inter-ensemble variability and improved baseline performance relative to~\methodname.
    }
    \label{fig:gm_short_psl}
\end{figure}

\begin{figure}[H]
    \centering
    \includegraphics[width=\textwidth]{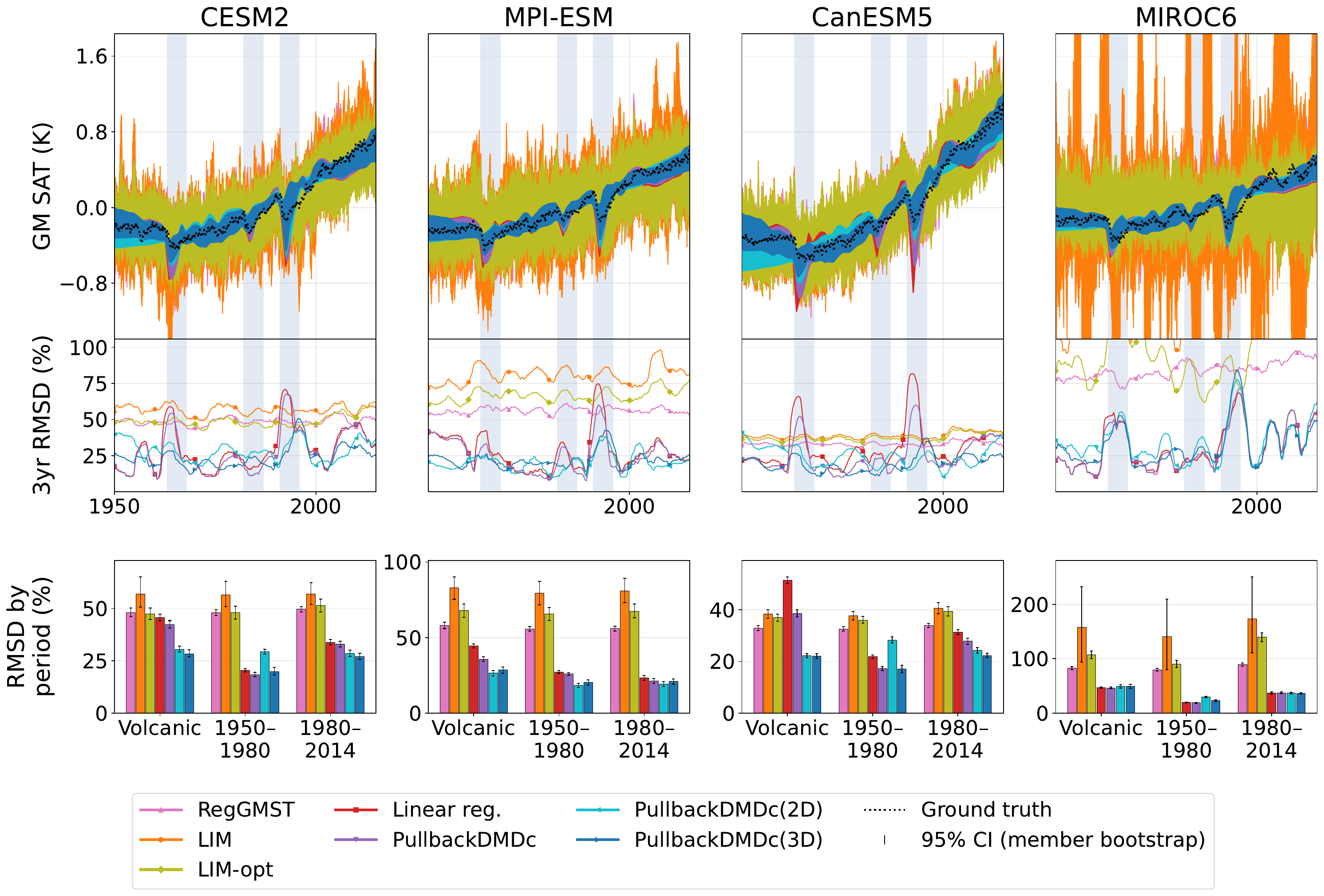}
    \caption{\added{Same as Fig.~\ref{fig:gm_short_psl}, but for surface air temperature (SAT).}}
    \label{fig:gm_short_tas}
\end{figure}

\begin{figure}[H]
    \centering
    \includegraphics[width=\textwidth]{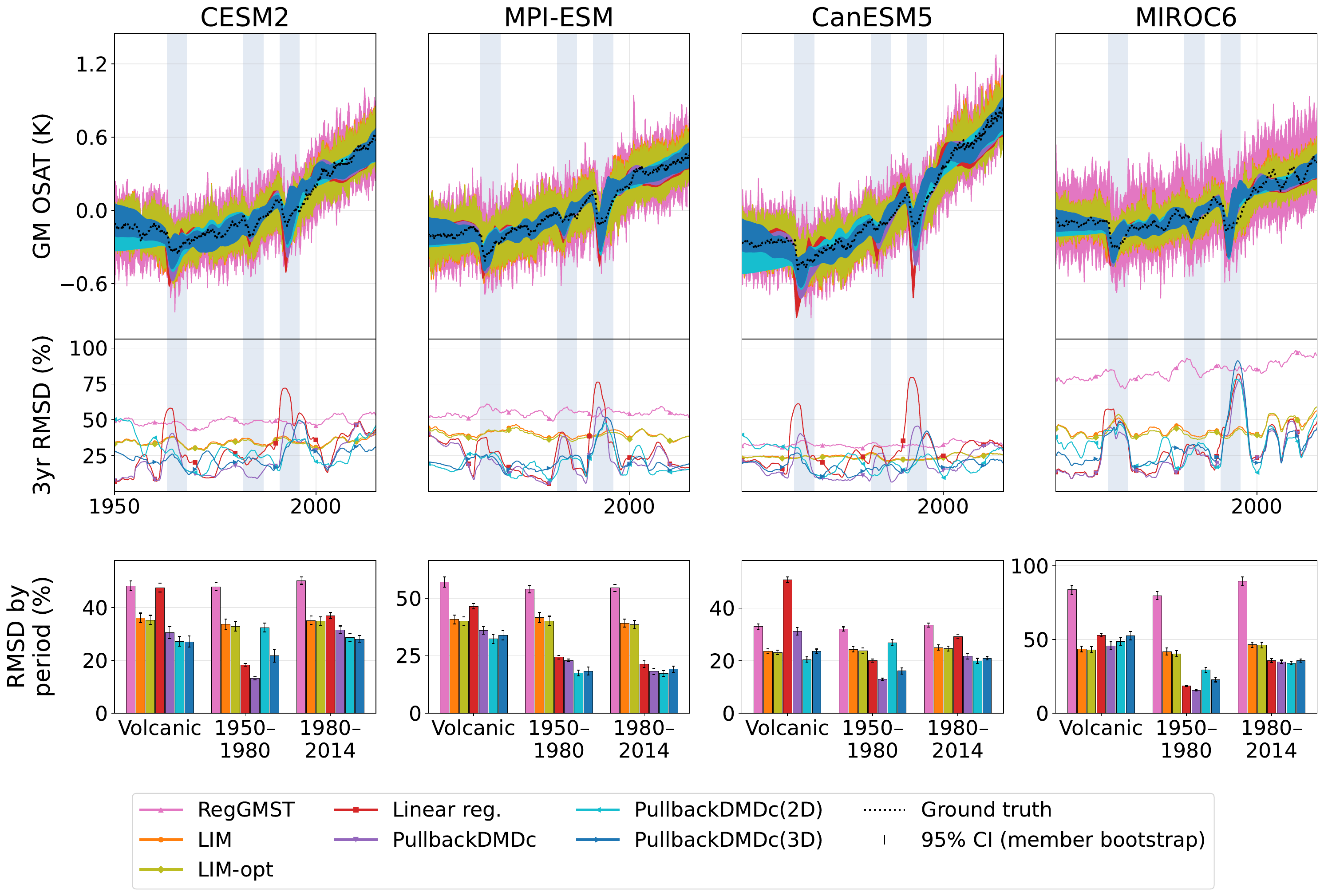}
    \caption{\added{Same as Fig.~\ref{fig:gm_short_psl}, but for ocean surface air temperature (OSAT).}}
    \label{fig:gm_short_tas_ocean}
\end{figure}

\begin{figure}[H]
    \centering
    \includegraphics[width=0.85\linewidth]{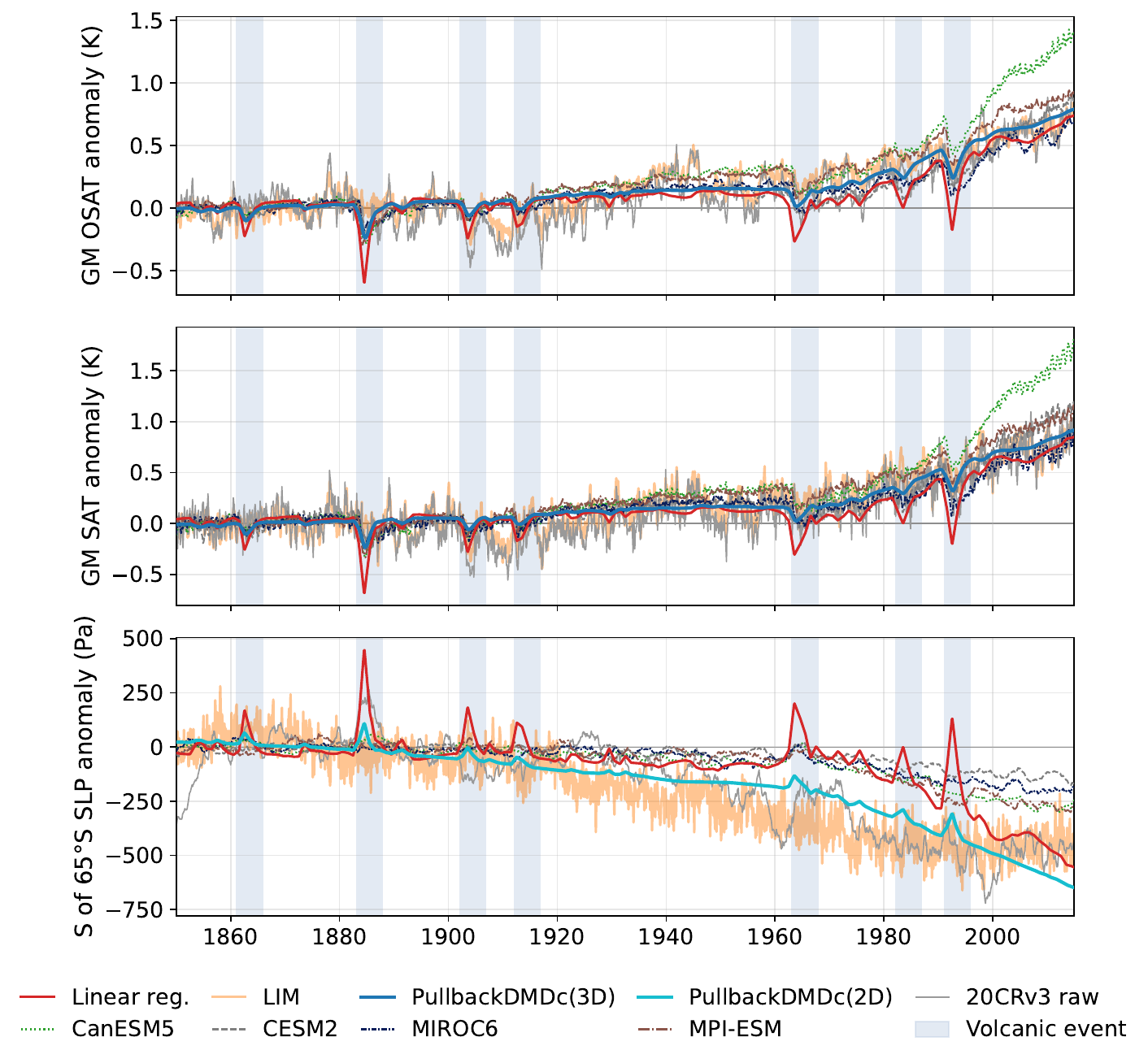}
    \caption{\added{Forced response estimated from 20CRv3: global mean of (top) OSAT and (middle) SAT, and (bottom) SLP averaged south of $65^\circ$S. Grey: raw 20CRv3 (forced + internal). Coloured: forced-response estimates of LR, LIM and \methodname(3D) (OSAT, SAT) or \methodname(2D) (SLP). Dashed and dotted: ensemble means of the four ESMs, shown as a reference rather than a ground truth. Shading marks the volcanic intervals as in Fig.~3. The forced OSAT and SAT estimates of \methodname(3D) are close to the ensemble means of CESM2, MIROC6 and MPI-ESM, while LR overestimates the cooling after volcanic eruptions, as in the ESMs. The strong decrease of the SLP estimates of \methodname(2D) and LIM since about $1900$ may reflect the weakly constrained early reanalysis at high southern latitudes; we do not interpret it further.}}
    \label{fig:gm_20crv3}
\end{figure}

\begin{figure}[H]
    \centering
    \includegraphics[width=\linewidth]{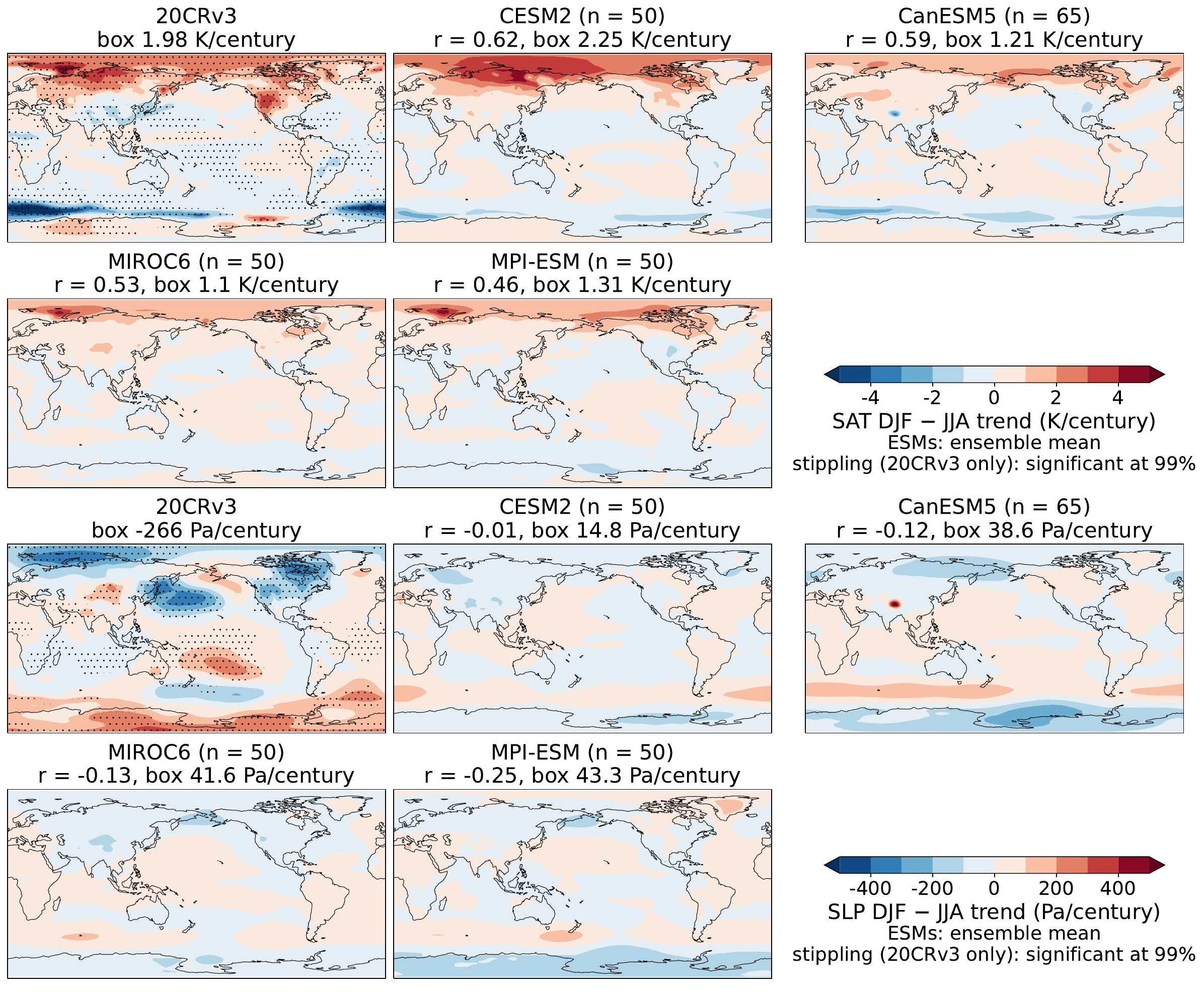}
    \caption{\added{Linear trend of DJF$-$JJA SAT (top) and SLP (bottom) over $1850$--$2014$ in 20CRv3 and in the ensemble mean of each ESM ($n$ members). Titles give the trend averaged north of $65^\circ$N (SAT) or over the North Pacific box ($25^\circ$--$45^\circ$N, $135^\circ$E--$165^\circ$W; SLP), and the area-weighted pattern correlation $r$ with 20CRv3. Stippling (20CRv3 only) as in Fig.~5. Single members match the DJF$-$JJA trend pattern of their own ensemble mean. Single members closely match their ensemble-mean trend patterns, whereas ensemble means show moderate pattern agreement for SAT and near-zero correlation for SLP relative to reanalysis.}}
    \label{fig:seasonal_drift_esm}
\end{figure}

\begin{figure}[H]
    \centering
    \includegraphics[width=\linewidth]{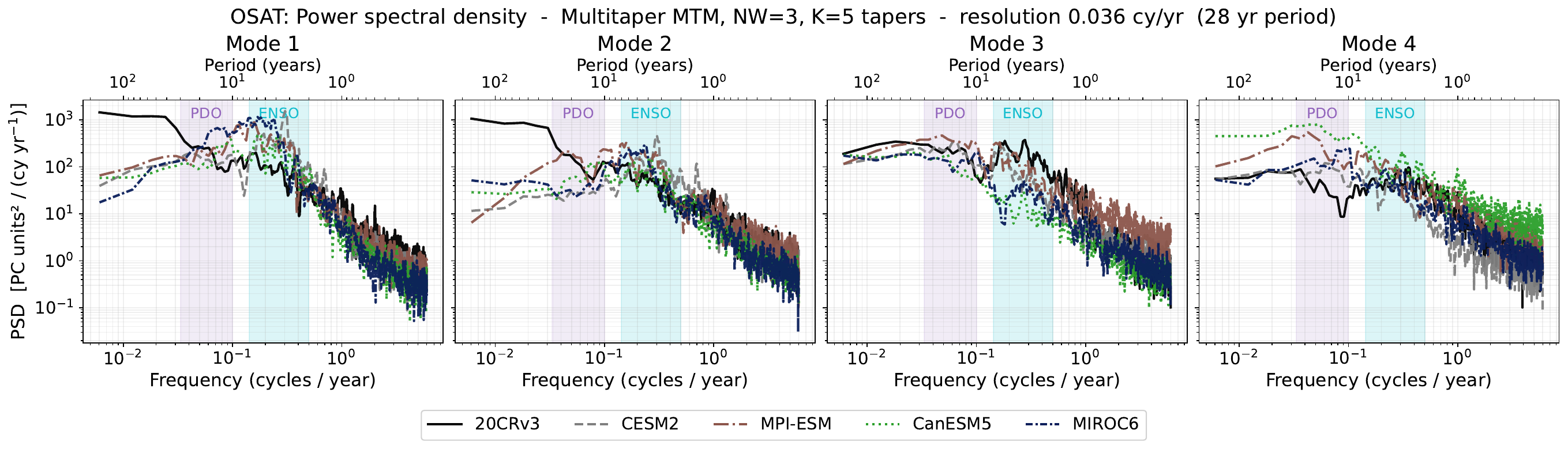}
    \caption{Power spectral density (PSD) of the top four DMD modes for OSAT obtained from 20CRv3 reanalysis (black solid line) and four ESMs [CESM2, MPI-ESM, CanESM5, and MIROC6]. PSD estimates are computed using the multitaper method ($NW = 3$, 5 tapers) with a frequency resolution of $0.036$ $cy /yr$ ($\approx28$ yr period) over the $1850$-$2014$ period. Frequency bands associated with ENSO ($2$–$7$ yr) and PDO ($10$–$30$ yr) as in Fig.~6. Each panel represents one mode, ordered by descending magnitude. The secondary x-axis provides equivalent periodicities. We generalize the lack of strong multidecadal variability in individual ensemble members to all ensemble members of all ESMs using the power spectral density (PSD) of the internal-variability time series. Specifically, there are high plateaus for reanalysis (20CRv3) for timescales larger than ENSO in modes $1$ and $2$. These maxima are not reached by any of the top $4$ modes of any ESM. The closest to reaching $10^3$ is mode $4$ of CanESM5.}
    \label{fig:psd_tas_ocean}
\end{figure}

\begin{figure}[H]
    \centering
    \includegraphics[width=.4\linewidth]{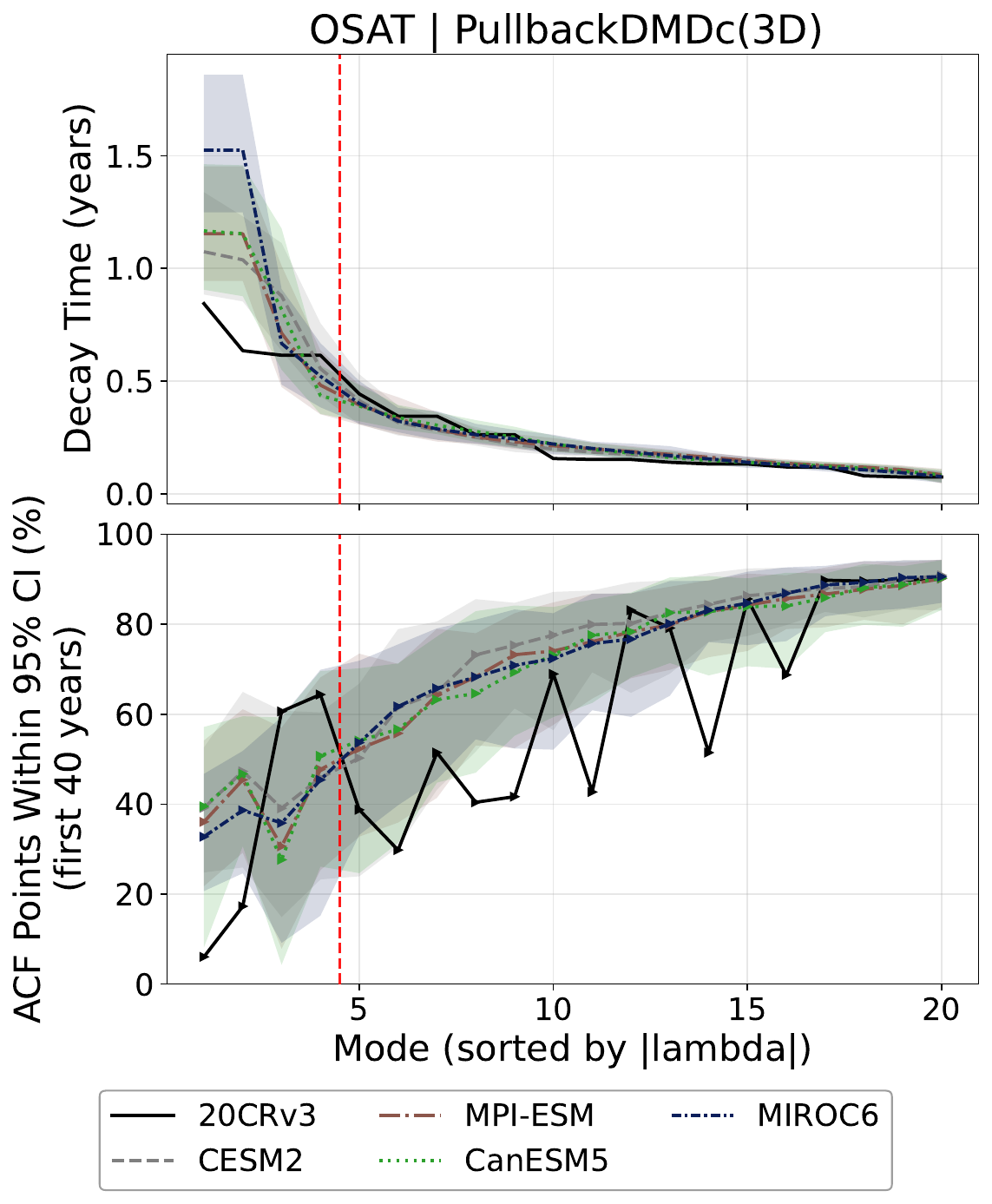}
    \caption{Two proxies for the signal in the leading $20$ \added{OSAT modes of \methodname(3D) for 20CRv3 and the four ESMs over $1850$--$2014$}, with modes sorted by decreasing $|\lambda|$. (Top) Decay time in months $-\tau/(12\ln|\lambda_k|)$, where $\lambda_k$ are the eigenvalues of $\A$ and $\tau = 3$~months. (Bottom) Percentage of ACF values of the internal component $\hat{z}^{(i)}(t)$ within the $95\%$ white-noise interval $\pm 1.96/\sqrt{N-k}$ ($N = 1980$) at monthly lags $k = 1,\dots,480$ (up to $40$~years). White noise gives about $95\%$, so lower values indicate more persistent structure. For the ESMs, lines show the mean across members and shading the $2.5$th--$97.5$th percentile range; 20CRv3 is a single realization. The dashed red line separates the four modes shown in Fig.~\ref{fig:dmd_modes_side_by_side_20crv3} from the rest.}
    \label{fig:mode_selection}
\end{figure}

\begin{figure}[H]
    \centering
    \includegraphics[width=\linewidth]{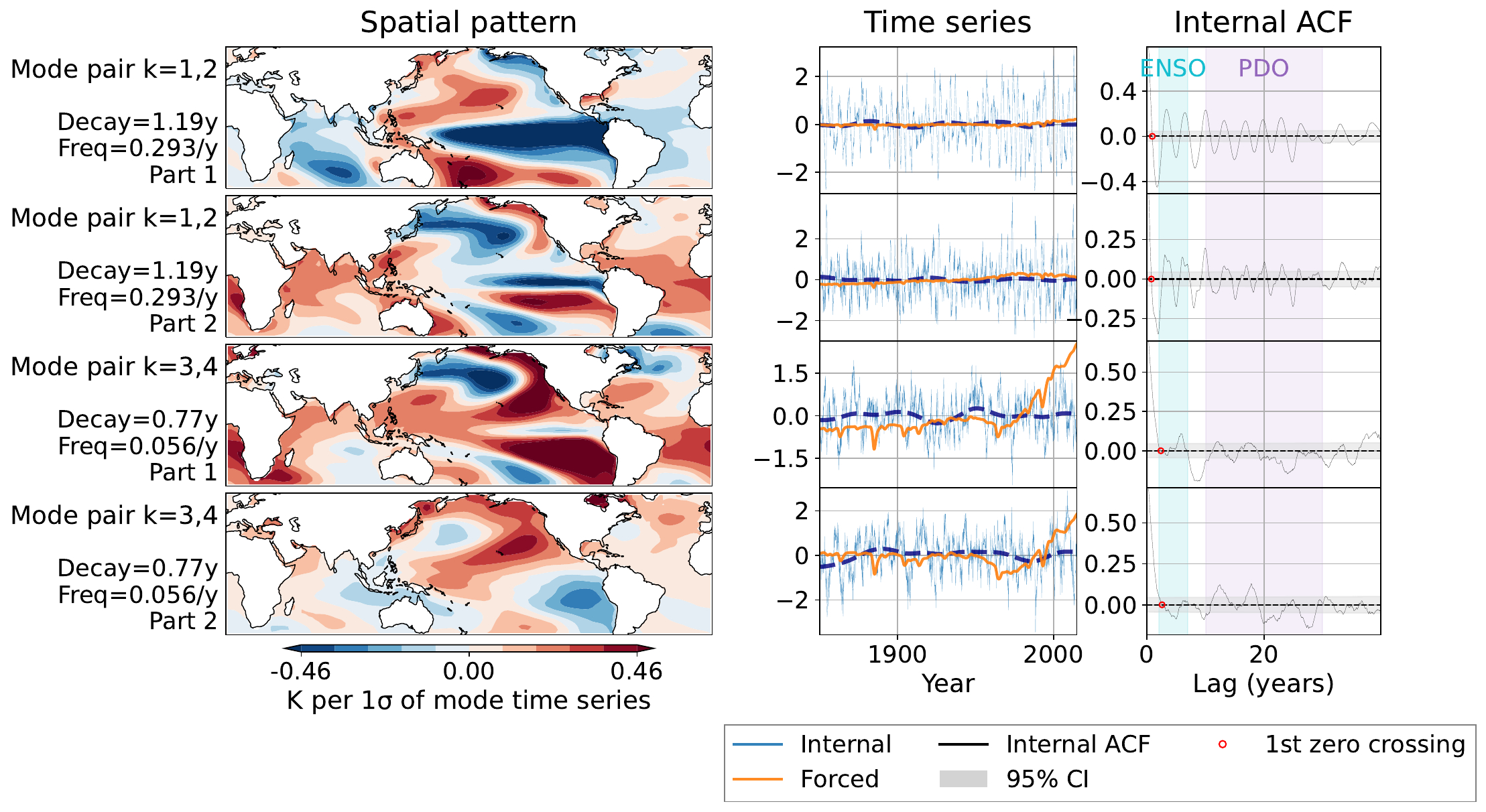}
    \caption{Same as in Fig. 6 (main text) but computed for an example member of CESM2 \added{(1281.005i1p1f1)}.
    }
    \label{fig:modes_tas_cesm2}
\end{figure}

\begin{figure}[H]
    \centering
    \includegraphics[width=\linewidth]{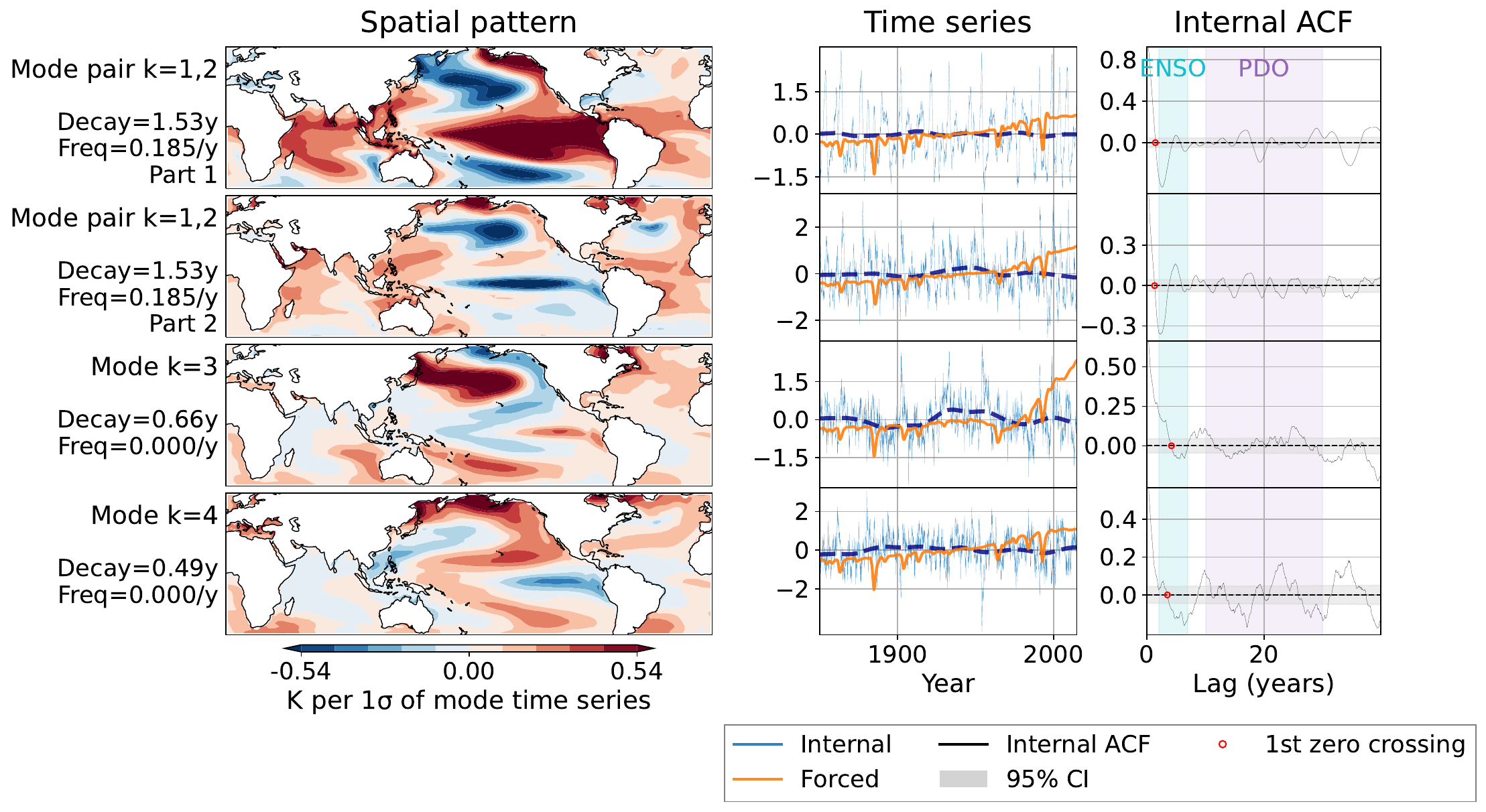}
    \caption{Same as in Fig. 6 (main text) but computed for an example member of MIROC6 \added{(r35i1p1f1)}.}
    \label{fig:modes_tas_miroc6}
\end{figure}

\begin{figure}[H]
    \centering
    \includegraphics[width=\linewidth]{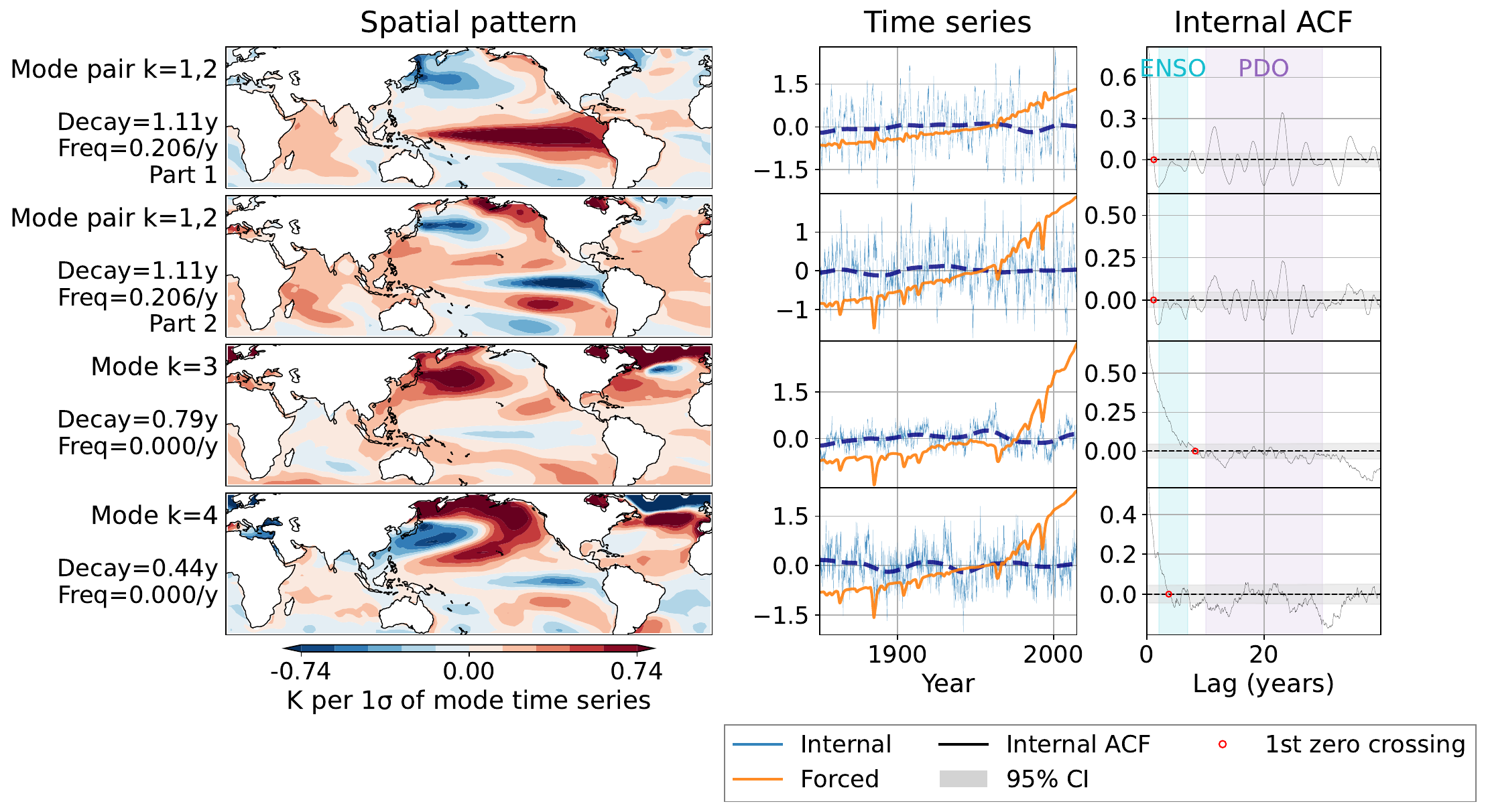}
    \caption{Same as in Fig. 6 (main text) but computed for an example member of CanESM5 \added{(r16i1p1f1)}. }
    \label{fig:modes_tas_cesm5}
\end{figure}

\begin{figure}[H]
    \centering
    \includegraphics[width=\linewidth]{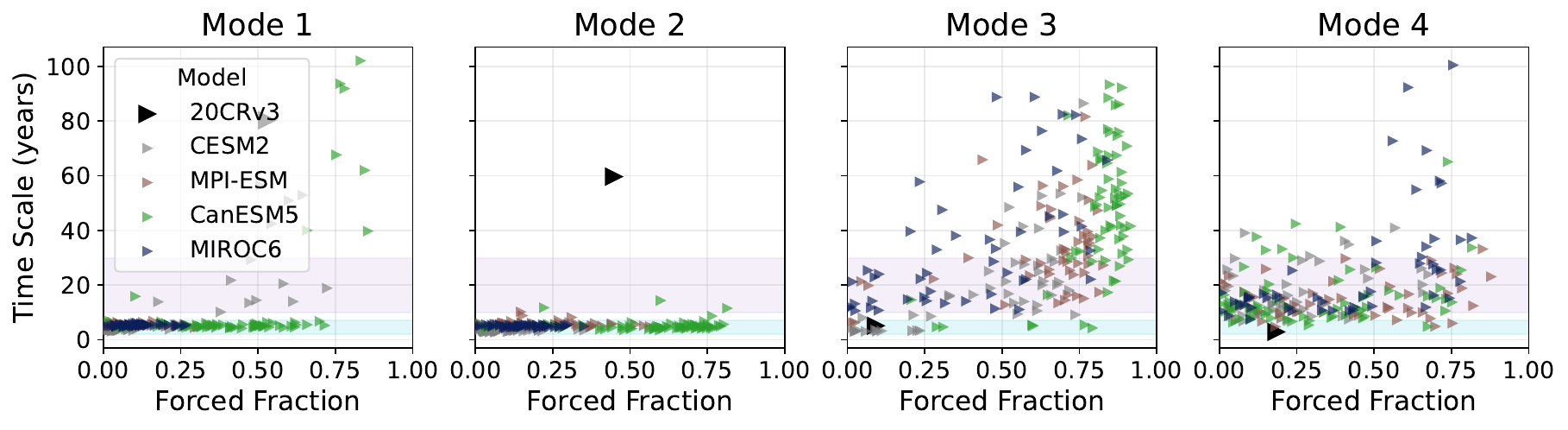}
    \caption{Relationship between forced-response fraction and intrinsic timescale for the leading four dynamical modes of ocean surface air temperature (OSAT) anomalies. Each panel represents one eigenmode of~\methodname(3D). The x-axis shows the fraction of total variance explained by the forced response, computed as the ratio of forced-response variance to the total variance of the mode, while the y-axis shows the dominant timescale of the corresponding internal variability estimated from its autocorrelation function. Horizontal shaded bands are explained in Fig.~6. Each point represents a single ensemble realization. 
    Consistent with reanalysis, ensemble members with slower internal-variability tend to show a stronger forced response. However, unlike in reanalysis, ENSO-like modes in the ESMs can still carry a non-negligible forced response.
    }
    \label{fig:forced_fraction_vs_timescale}
\end{figure}

\begin{figure}[H]
    \centering
    \includegraphics[width=\linewidth]{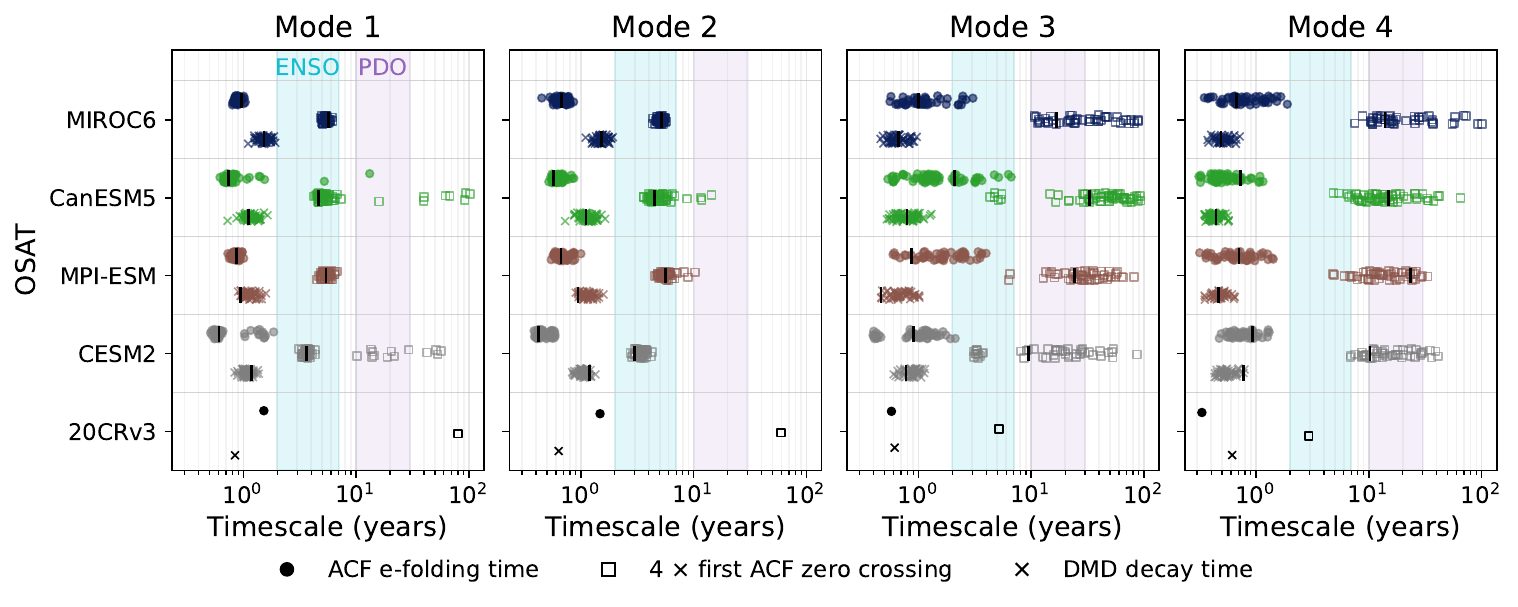}
    \caption{\added{Timescales of the leading \methodname(3D) OSAT modes: ACF e-folding time (circles), four times the first ACF zero crossing (squares, as in Fig.~8) and DMD decay time (crosses). Black bars mark the members in Figs.~7 and S23--S25; bands as in Fig.~6. The ACF e-folding time is set by the rapid initial drop of the ACF, dominated by fast, noise-like components, and is therefore of the same order as the decay time; the slow components appear only in the ACF tail, which the dominant timescale captures. Decay and e-folding times describe damping and stay below about $2$~years for all OSAT modes, whereas the dominant timescale (Section~3c of the main text) measures persistence, equals the period for oscillatory modes, and ranges from years to decades. We therefore separate ENSO-like from multidecadal variability by the dominant timescale, supported by the spectra (Fig.~S21).}}
    \label{fig:acf_efolding_tas_ocean}
\end{figure}

\begin{figure}[!ht]
    \centering
    \includegraphics[width=\linewidth]{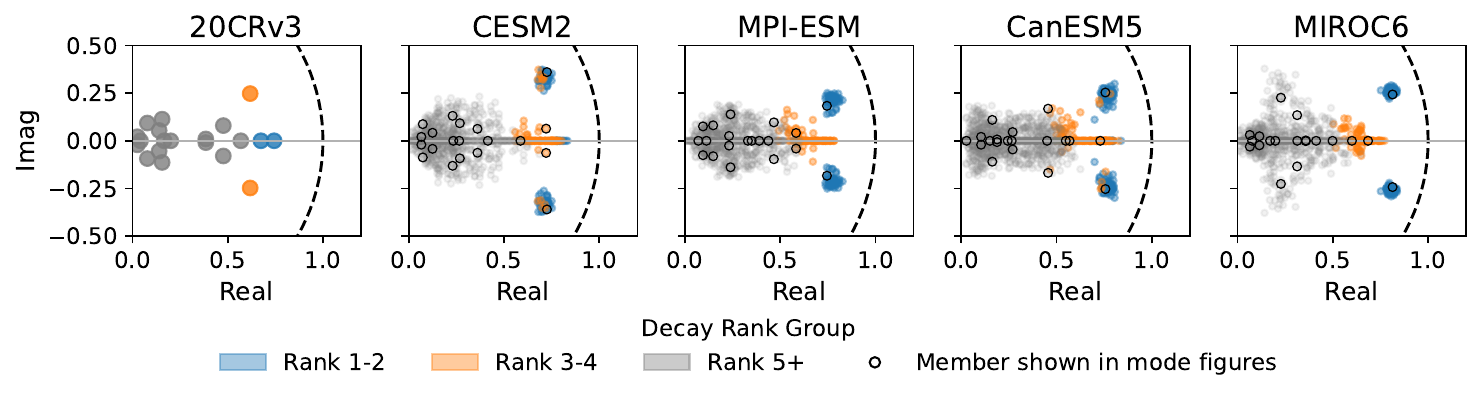}
    \caption{\added{Eigenvalues $\lambda_k$ of $\A$ from \methodname(3D) for OSAT in 20CRv3 and all ESM members, in the complex plane. Colors group the eigenvalues by decay rank (1--2, 3--4 and 5+), counted from the largest $|\lambda_k|$, i.e., the longest decay time (Eq.~(2.5) of the main text). Real eigenvalues lie on the horizontal axis, and complex-conjugate pairs lie symmetrically about it; the dashed arc is the unit circle. In 20CRv3, the two least-damped modes are real, followed by an oscillatory pair. In the ESMs, the order is reversed: the two least-damped modes are almost always an oscillatory pair, and real modes often follow at ranks 3--4. Black circles mark the members shown in Fig.~7 of the main text and Figs.~S23--S25. Fig.~S11 compares these eigenvalues with those of LIM.}}
    \label{fig:supp_eigs}
\end{figure}

\end{document}